\documentclass[10pt,journal,compsoc]{IEEEtran}
\usepackage{times}
\usepackage{epsfig}
\usepackage{graphicx}
\usepackage{amsmath}
\usepackage{amssymb}
\usepackage{multirow}
\usepackage{colortbl}
\usepackage[table]{xcolor}

\usepackage{enumitem}
\usepackage[normalem]{ulem}
\useunder{\uline}{\ul}{}
\newcommand\B[1]{\textcolor{blue}{#1}}
\usepackage[pagebackref=true,breaklinks=true,colorlinks,bookmarks=false]{hyperref}

\definecolor{gbygreen}{rgb}{0.9,0.9,0.85}
\definecolor{gbypink}{rgb}{0.99, 0.91, 0.95}
\definecolor{gbygray}{rgb}{0.9,0.8,0.9}

\ifCLASSOPTIONcompsoc
  \usepackage[nocompress]{cite}
\else
  \usepackage{cite}
\fi
\ifCLASSINFOpdf
\else
\fi
\begin{document}
%
\title{Digging Into Uncertainty-based Pseudo-label for Robust Stereo Matching}
%
%
%
%

\author{Zhelun~Shen,
        Xibin~Song,
        Yuchao~Dai,~\IEEEmembership{Member,~IEEE,}
        Dingfu~Zhou,
        Zhibo Rao
        and~Liangjun~Zhang 
\IEEEcompsocitemizethanks{\IEEEcompsocthanksitem Zhelun Shen, Xibin Song, Dingfu Zhou and Liangjun Zhang are with Robotics and Autonomous Driving Lab, Baidu Research, China. E-mail: \{shenzhelun, zhoudingfu, liangjunzhang\}@baidu.com, song.sducg@gmail.com.
\IEEEcompsocthanksitem Yuchao Dai, Zhibo Rao are with School of Electronics and Information, Northwestern Polytechnical University and Shaanxi Key Laboratory of Information Acquisition and Processing, Xi'an, 710072, China. E-mail: daiyuchao@gmail.com, raoxi36@foxmail.com.

\IEEEcompsocthanksitem Corresponding authors: Xibin Song and Yuchao Dai.}

}




\markboth{Journal of \LaTeX\ Class Files,~Vol.~14, No.~8, August~2015}%
{Shell \MakeLowercase{\textit{et al.}}: Bare Demo of IEEEtran.cls for Computer Society Journals}
%



\IEEEtitleabstractindextext{%
\begin{abstract}

Due to the domain differences and unbalanced disparity distribution across multiple datasets, current stereo matching approaches are commonly limited to a specific dataset and generalize poorly to others. Such domain shift issue is usually addressed by substantial adaptation on costly target-domain ground-truth data, which cannot be easily obtained in practical settings. In this paper, we propose to dig into uncertainty estimation for robust stereo matching. Specifically, to balance the disparity distribution, we employ a pixel-level uncertainty estimation to adaptively adjust the next stage disparity searching space, in this way driving the network progressively prune out the space of unlikely correspondences. Then, to solve the limited ground truth data, an uncertainty-based pseudo-label is proposed to adapt the pre-trained model to the new domain, where pixel-level and area-level uncertainty estimation are proposed to filter out the high-uncertainty pixels of predicted disparity maps and generate sparse while reliable pseudo-labels to align the domain gap. Experimentally, our method shows strong cross-domain, adapt, and joint generalization and obtains \textbf{1st} place on the stereo task of Robust Vision Challenge 2020. Additionally, our uncertainty-based pseudo-labels can be extended to train monocular depth estimation networks in an unsupervised way and even achieves comparable performance with the supervised methods. The code will be available at   \href{https://github.com/gallenszl/UCFNet}{https://github.com/gallenszl/UCFNet}.


\end{abstract}

}

\maketitle

\IEEEdisplaynontitleabstractindextext

%
\IEEEpeerreviewmaketitle

\section{Introduction}


Stereo matching is a classical research topic in computer vision, which aims to estimate a disparity/depth map from a pair of rectified stereo images. It is a key enabling technique for various applications, such as autonomous driving \cite{autonomousdriving}, robot navigation \cite{roboticsnavigation}, SLAM \cite{slam1,slam2}, etc. Currently, impressive performances have been achieved by many deep learning-based stereo methods on most of the standard benchmarks.



\begin{figure*}[!t]
	\centering
	\tabcolsep=0.05cm
	\begin{tabular}{c c c c}
    \includegraphics[width=0.2\linewidth]{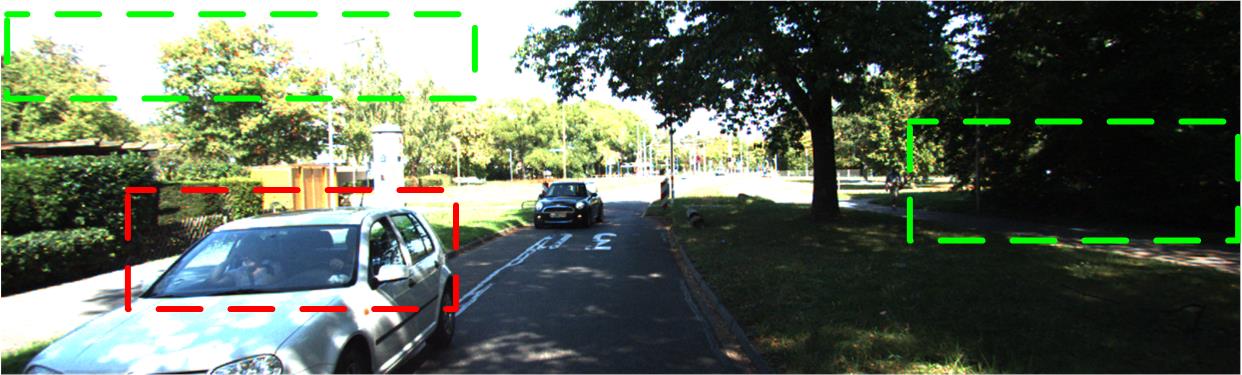}&
	\includegraphics[width=0.2\linewidth]{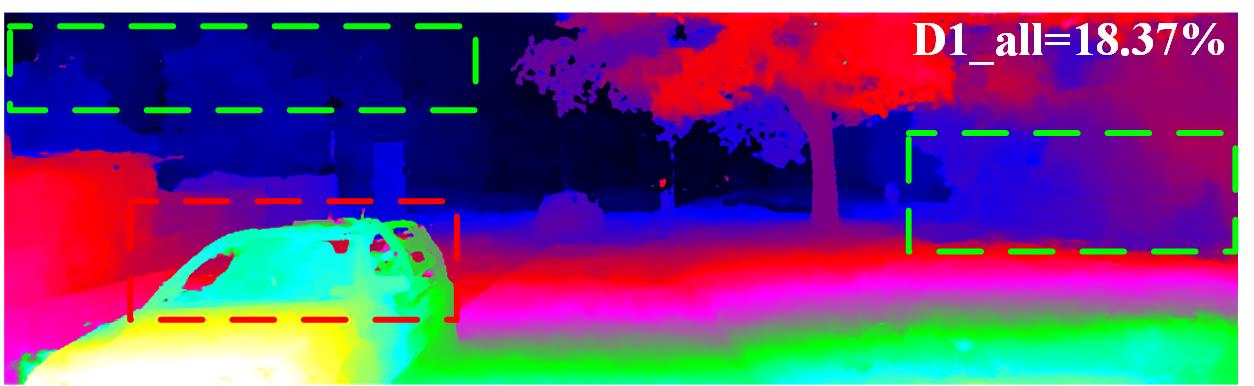}&
	\includegraphics[width=0.2\linewidth]{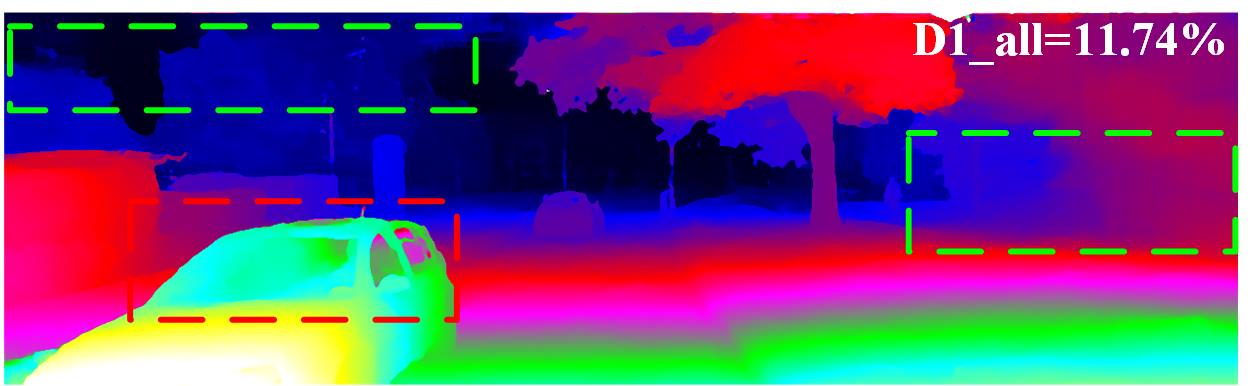}&
	\includegraphics[width=0.2\linewidth]{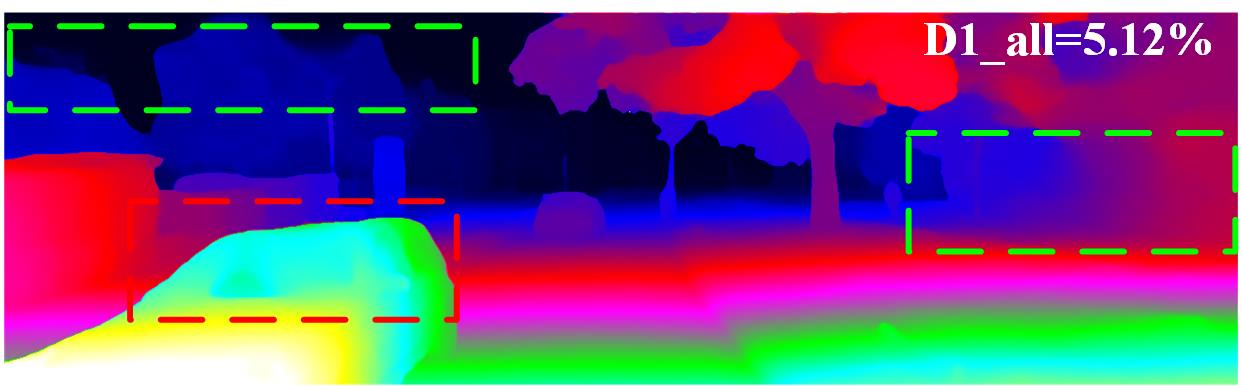}\\

	
	
	\includegraphics[width=0.2\linewidth]{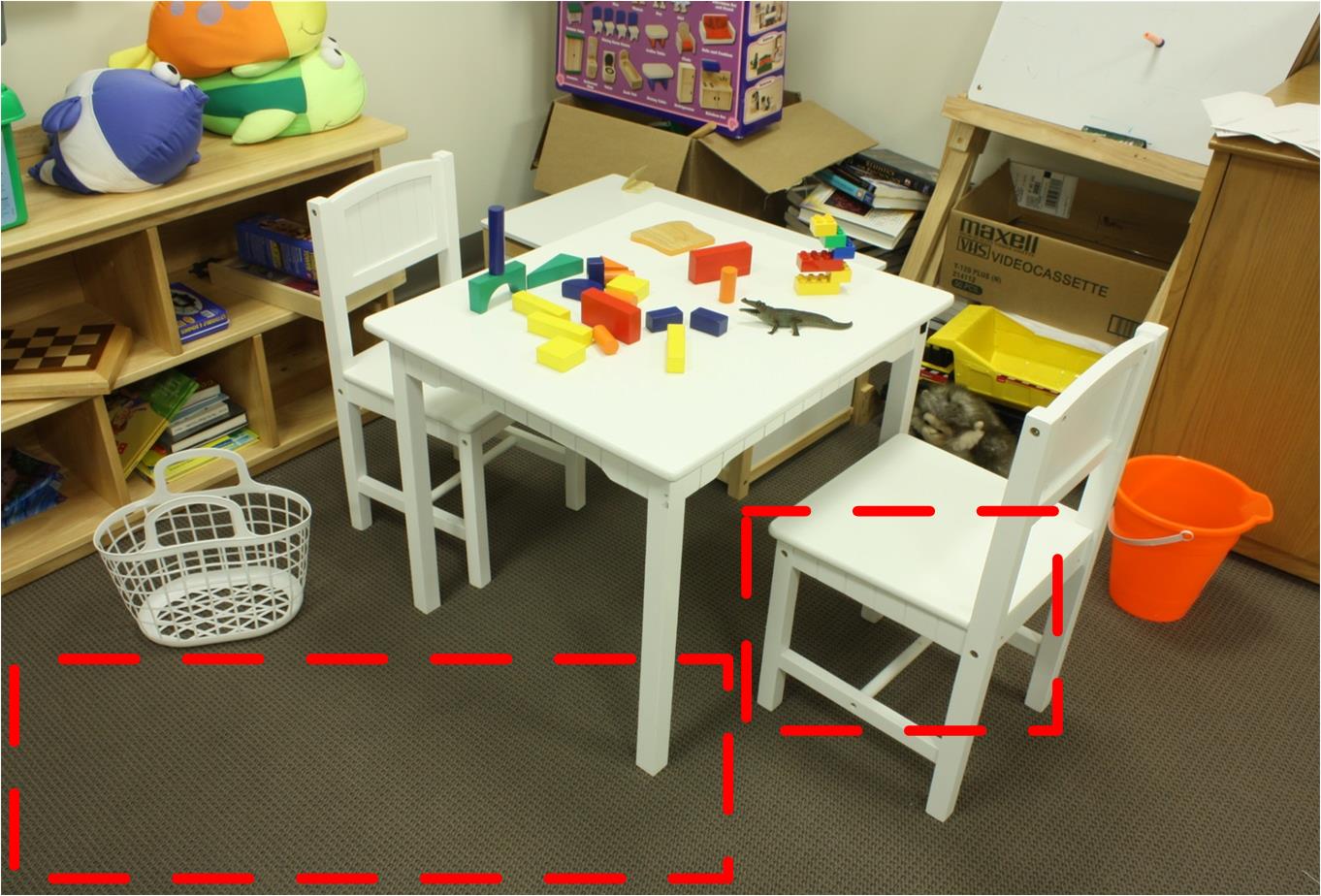}&
	\includegraphics[width=0.2\linewidth]{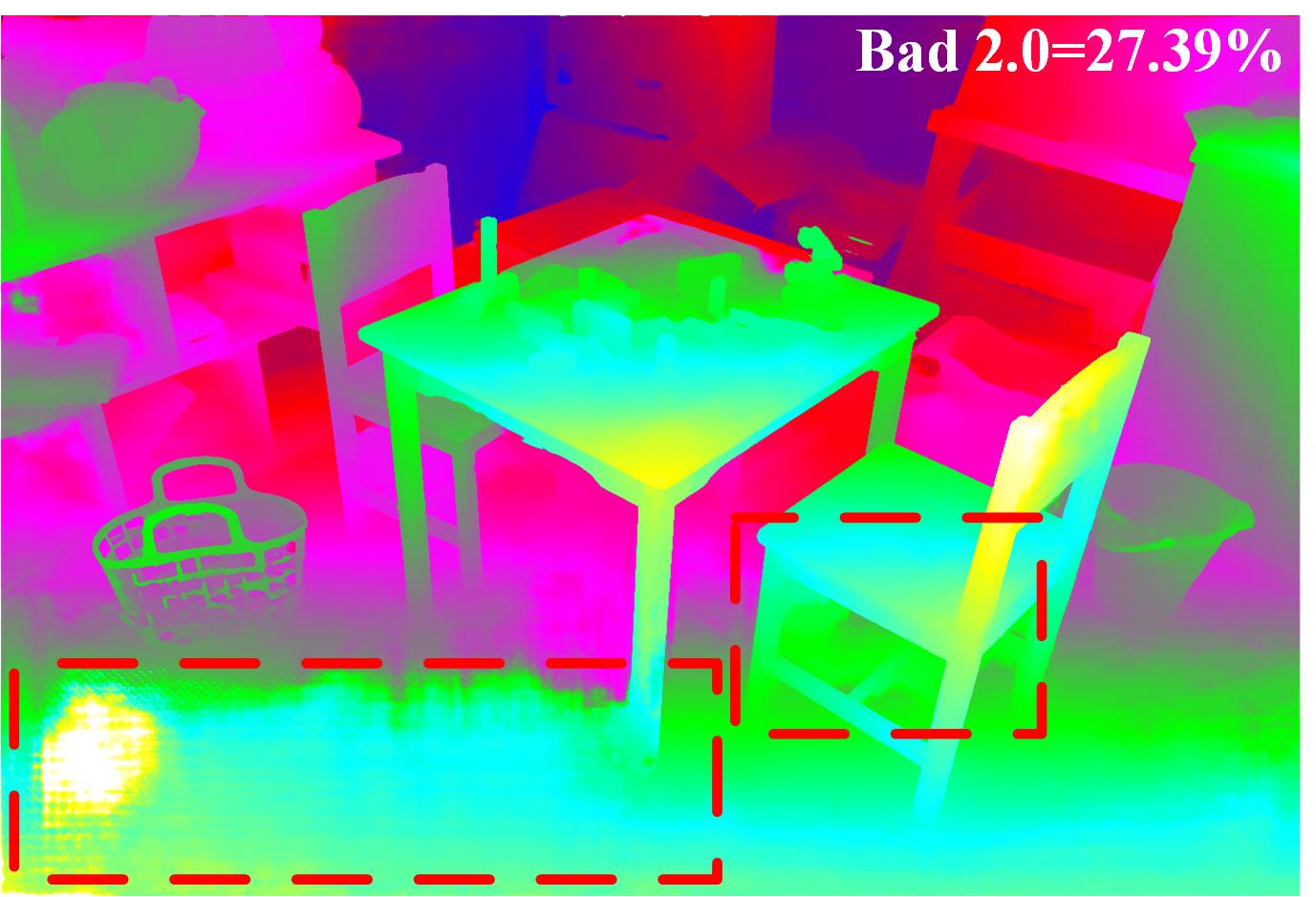}&
	\includegraphics[width=0.2\linewidth]{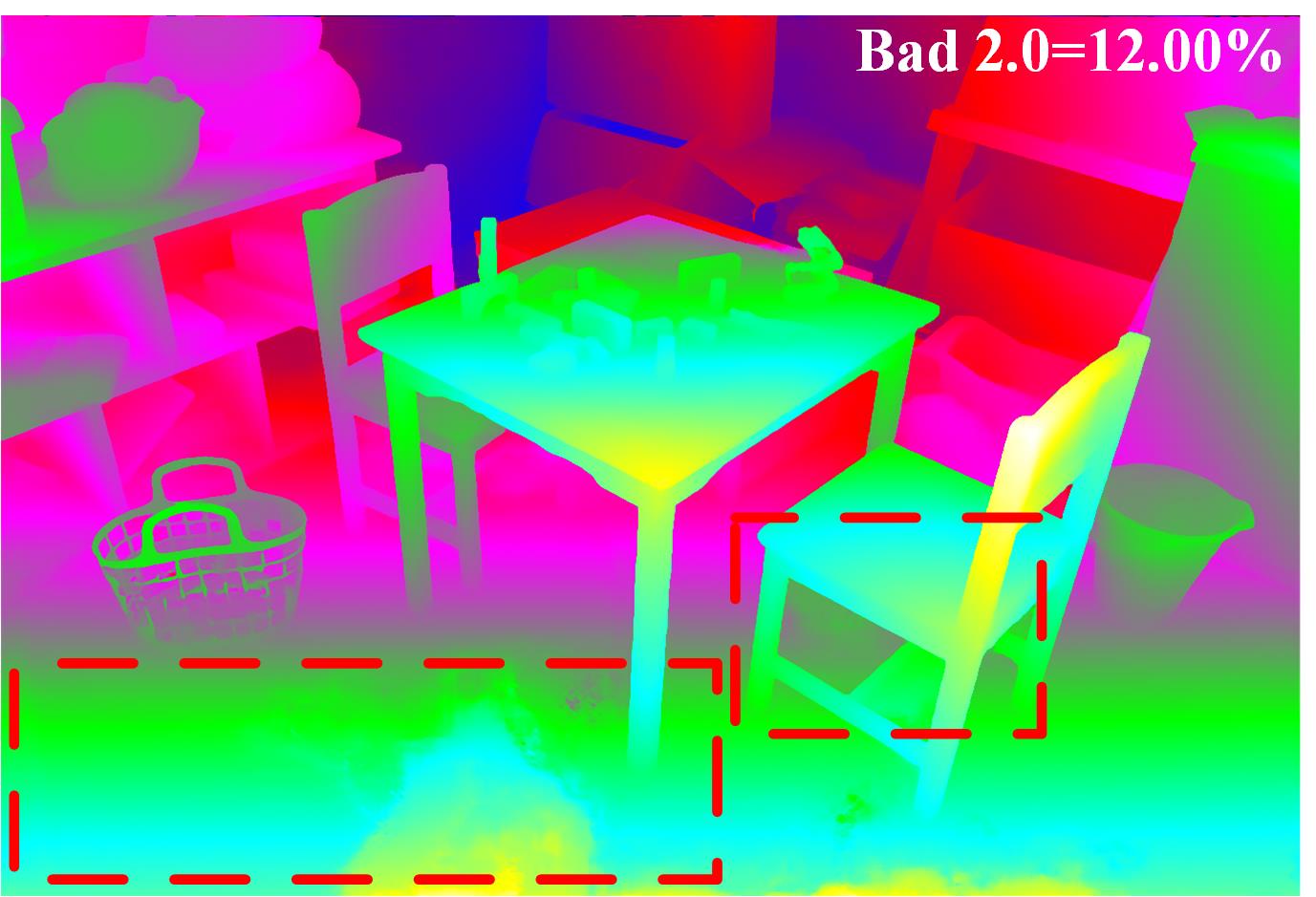}&
	\includegraphics[width=0.2\linewidth]{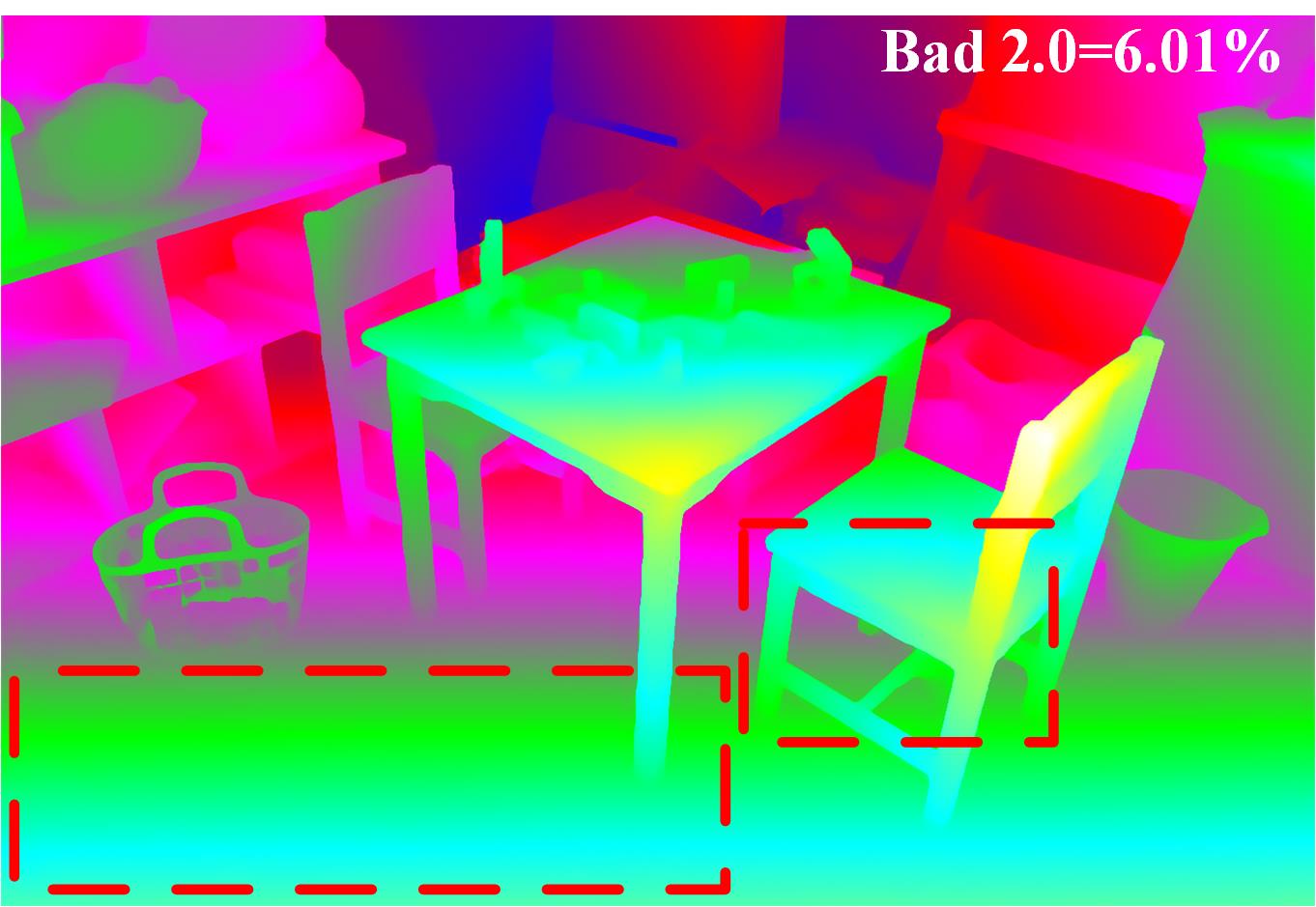}\\

	\includegraphics[width=0.2\linewidth]{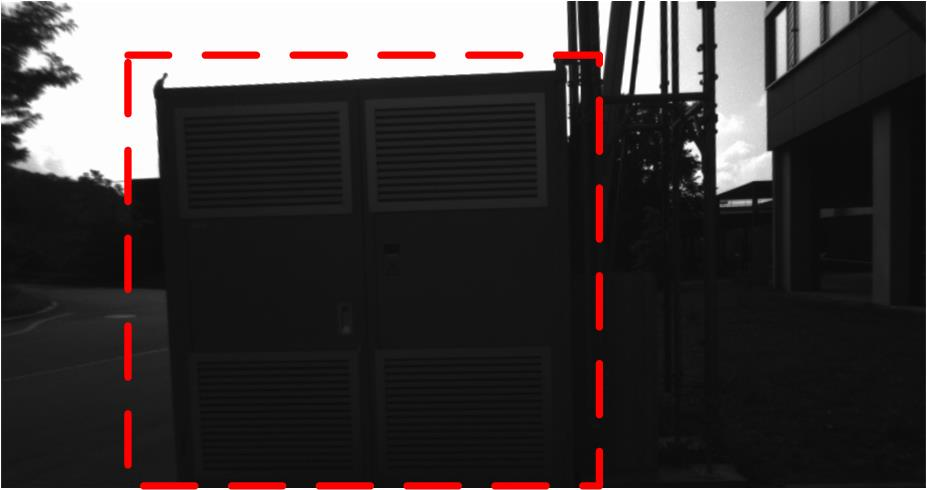}&
	\includegraphics[width=0.2\linewidth]{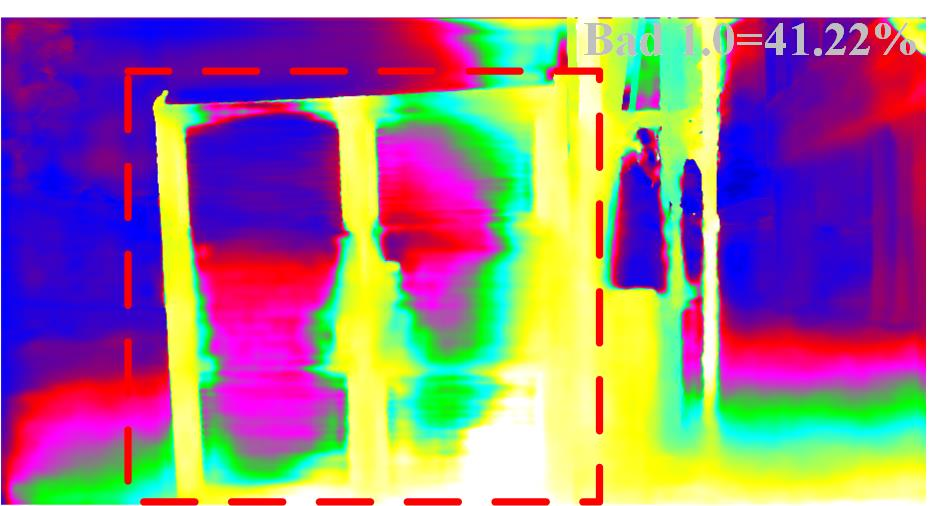}&
	\includegraphics[width=0.2\linewidth]{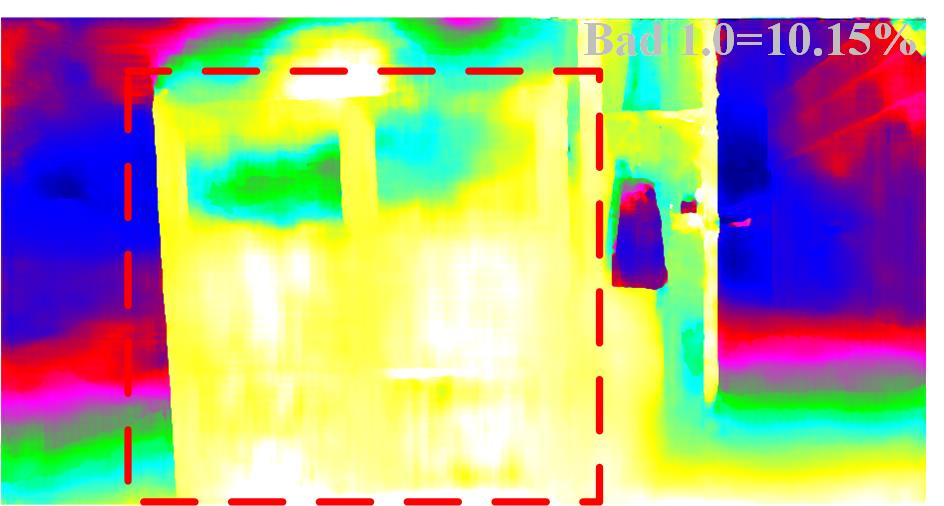}&
	\includegraphics[width=0.2\linewidth]{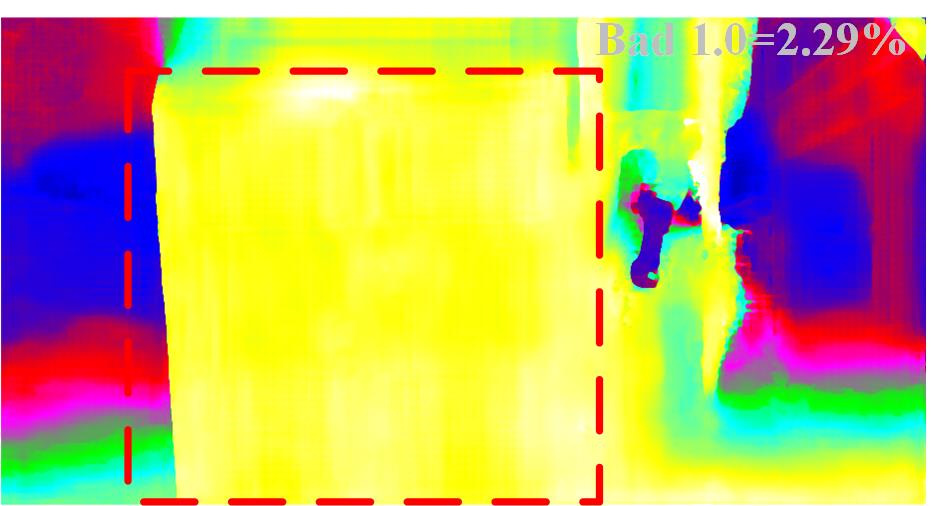}\\

	{\small (a) Left image} &  {\small (b) GANet \cite{ganet}} &	{\small (c) UCFNet\_pretrain }	&  {\small (d) UCFNet\_adapt }	  	\\
	\end{tabular}
 	\vspace{-0.05in}
	\caption{\footnotesize Generalization evaluation on three real-world datasets (from top to bottom: KITTI2015, Middlebury, and ETH3D). The left panel shows the left input image of the stereo image pair, and the others show the predicted colorized disparity map. GANet and UCFNet\_pretrain are trained on the Scene Flow dataset and tested on training images of real datasets. UCFNet\_adapt further uses the generated pseudo-labels to adapt the pre-trained model UCFNet\_pretrain to the target domain.}
	\label{fig:  adapt generalization intro}
	\vspace{-0.1in}
\end{figure*}

\begin{figure}[!htb]
	\centering
    \includegraphics[width=0.8 \linewidth]{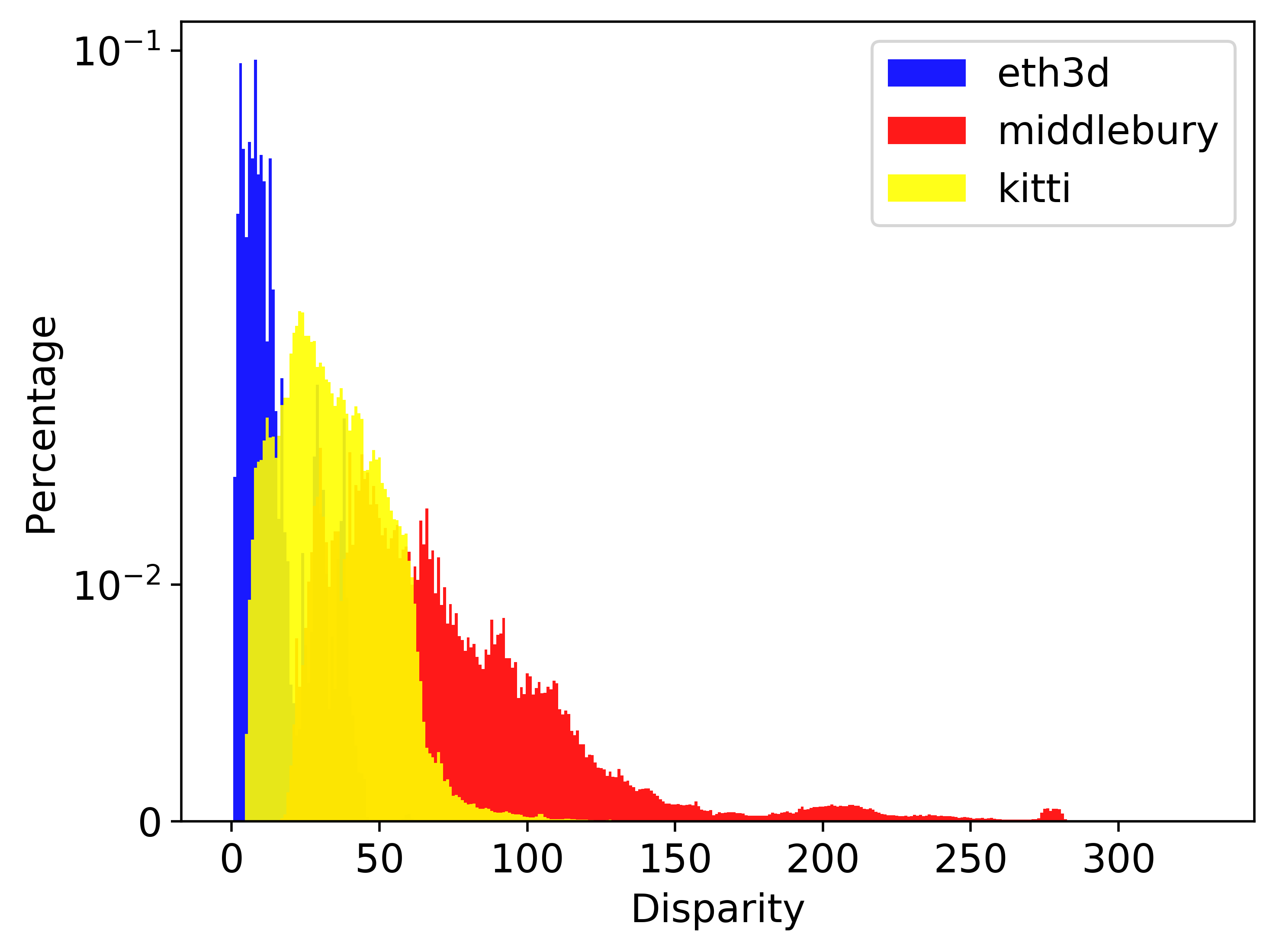} 
    	\vspace{-0.2in}
	\caption{\footnotesize Disparity distribution of KITTI 2015, Middlebury, and ETH3D training sets. We plot disparity distribution of half-resolution images for Middlebury while full-resolution images for other datasets. The disparity distribution across different datasets is unbalanced.}
	\label{fig: disparity distribution}
 	\vspace{-0.1in}
\end{figure}

However, significant domain shifts commonly exist among different datasets, which limits the generalization abilities of current state-of-the-art stereo matching methods. For example, the Middlebury \cite{mid} dataset mainly contains indoor high-resolution scenes while the KITTI dataset \cite{kitti1,kitti2} mainly consists of real-world urban driving scenarios. More specifically, as illustrated in Fig.~\ref{fig:  adapt generalization intro} (a), there are significant differences among various datasets, e.g., indoors vs outdoors, color vs gray, and real vs synthetic. In addition, the disparity ranges are different among various datasets. As illustrated in Fig.~\ref{fig: disparity distribution}, the disparity range of half-resolution images in Middlebury \cite{mid} is even more than 6 times larger than full-resolution images in ETH3D \cite{eth3d} (400 vs 64). Such unbalanced disparity distributions make the current approaches trained with a fixed disparity range difficult to cover the whole disparity range of another dataset without substantial finetuning.


\begin{figure}[!htb]
	\centering
	\tabcolsep=0.05cm
	\begin{tabular}{c c}
    \includegraphics[width=0.45\linewidth]{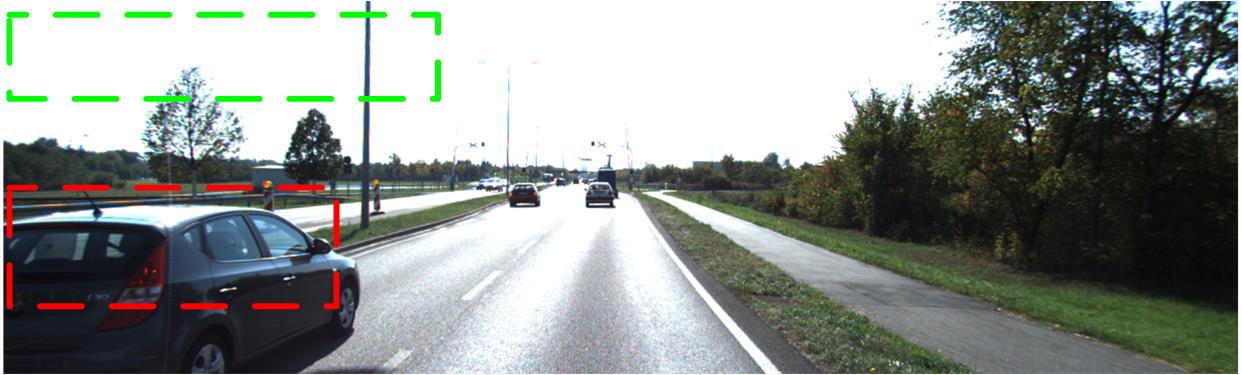}&
	\includegraphics[width=0.45\linewidth]{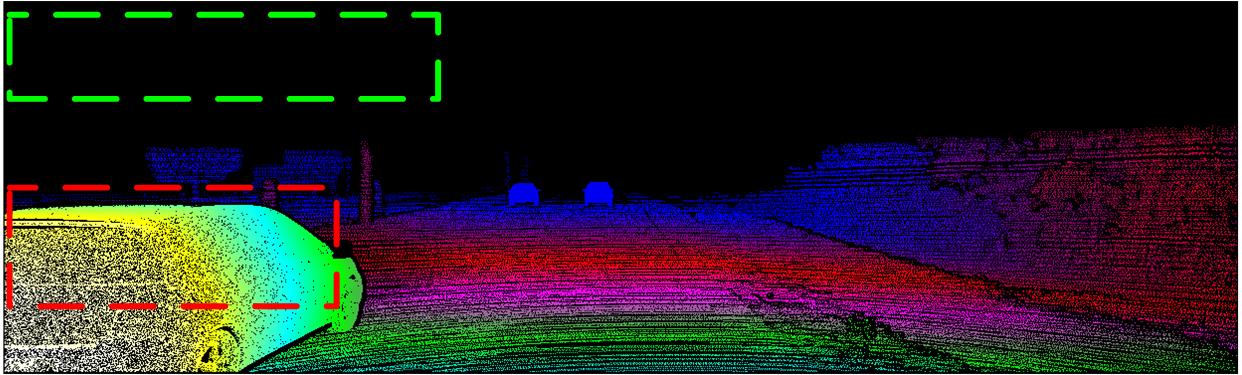} \\
	\multicolumn{2}{c}{\small (a) Left image and corresponding ground truth}		 \\

    \includegraphics[width=0.45\linewidth]{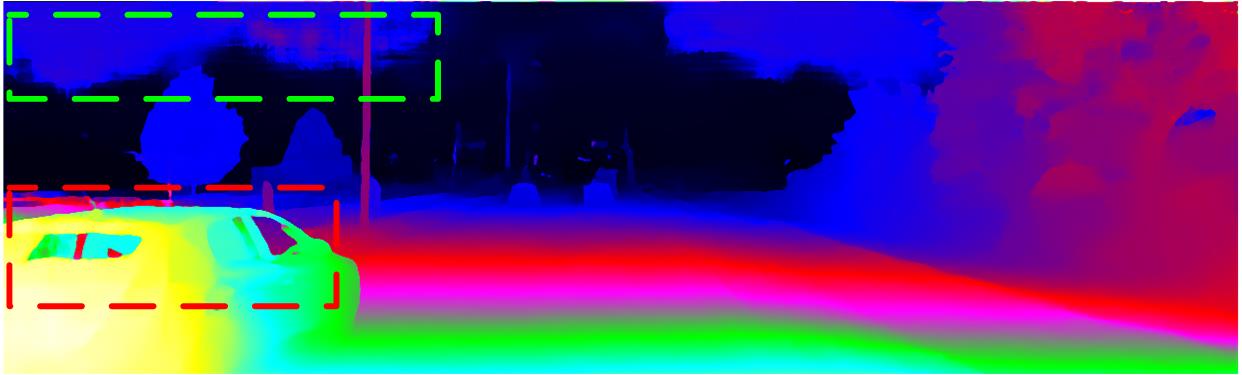}&
	\includegraphics[width=0.45\linewidth]{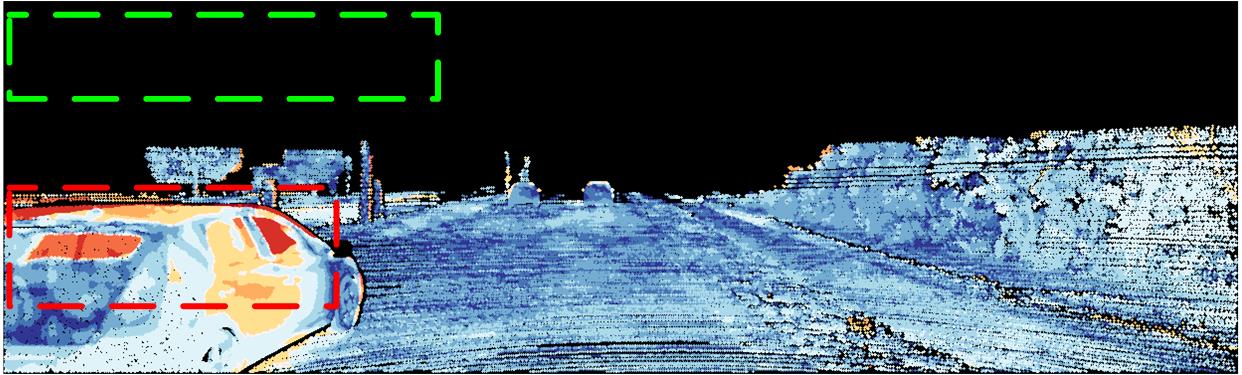} \\	
	\multicolumn{2}{c}{\small (b) Result of UCFNet\_pretrain}		 \\

    \includegraphics[width=0.45\linewidth]{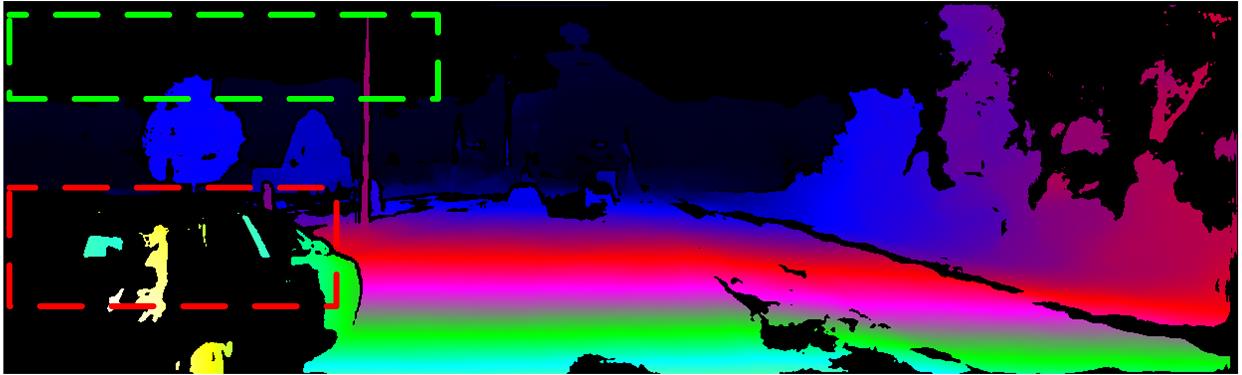}&
	\includegraphics[width=0.45\linewidth]{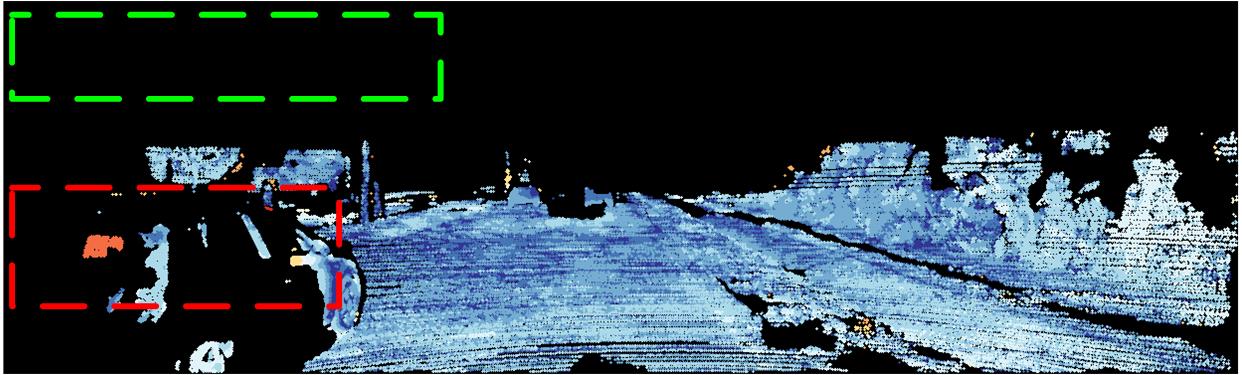} \\
	\multicolumn{2}{c}{\small (c) Result of generated pseudo-label}		 \\
    \includegraphics[width=0.45\linewidth]{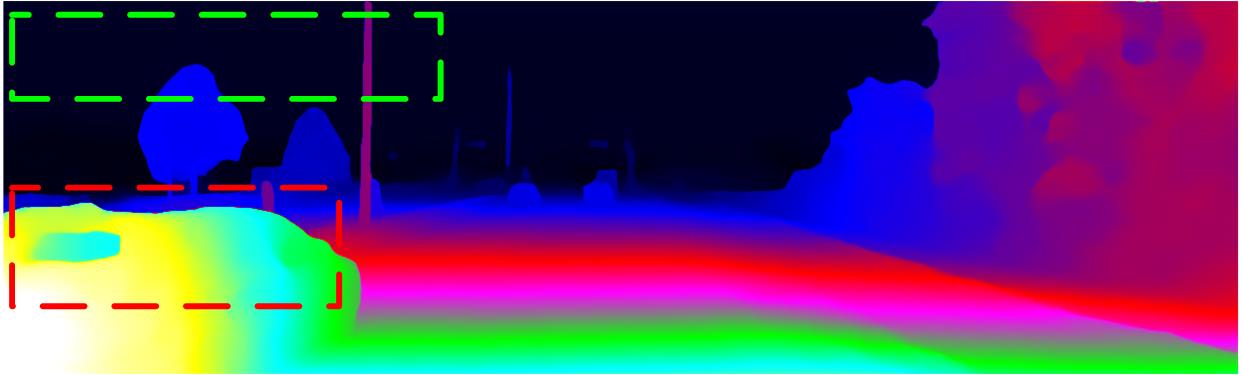}&
	\includegraphics[width=0.45\linewidth]{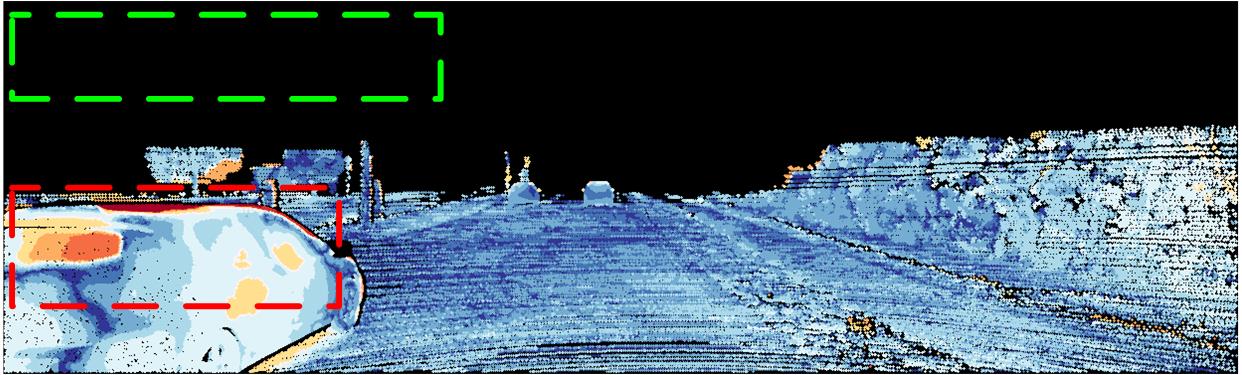} \\
	\multicolumn{2}{c}{\small (d) Result of UCFNet\_adapt}		 \\
	
	\end{tabular}
	\vspace{-0.05in}
	\caption{\footnotesize Generalization Visualization on KITTI2015 dataset. UCFNet\_pretrain are trained on the Scene Flow dataset and tested on training images of real datasets. UCFNet\_adapt further uses the generated pseudo-labels (sub-figure (c)) to adapt the pre-trained model UCFNet\_pretrain to the target domain. As shown, the self-generated proxy label can tremendously improve the performance of our pre-training model on both textureless areas of foreground (red dash boxes) and unlabeled areas of background (green dash boxes) of the target domain.} 
	\label{fig: intro all visual}
 	\vspace{-0.1in}
\end{figure}

Consequently, methods with state-of-the-art performance on one dataset often cannot achieve comparable results on other datasets without substantial adaptation. To relieve the problem, our conference paper CFNet\cite{cfnet} proposes a cascade and fused cost volume representation to narrow down the domain difference.
By employing the cascade cost volume representation to alleviate the unbalanced disparity distribution, the method can eliminate the need for adaptation and performs well across a variety of datasets with fixed model parameters and hyperparameters, i.e., joint generalization.
Unfortunately, such a method still needs suitable labeled target domain data, which cannot be easily obtained in most practical settings. Moreover, the labeled ground truth data is commonly obtained by expensive sensors (e.g. LiDAR) alongside careful calibration, which is cumbersome and costly, limiting the applicability in practical settings. Thus, we need to push methods to be robust and perform well across different datasets without using the groundtruth labels from the target domain.

In this paper, an Uncertainty-based Cascade and Fused cost volume representation (UCFNet) is proposed to dig into uncertainty estimation for robust stereo matching. Specifically, an uncertainty-based pseudo-labels generation method is proposed to adapt the pre-trained model to the new domain, i.e., domain adaptation. A key observation behind our method is that learning-based models can be successfully adapted to new domains even by deploying only a few sparse groundtruth annotations. For example, learning-based models can achieve state-of-the-art performance on KITTI datasets with limited sparse groundtruth (less than 1/3 pixels is annotated for totally 200 images). Thus, we can employ the proposed uncertainty estimation to filter out the high-uncertainty pixels of the pre-trained model and generate sparse while reliable disparity maps as pseudo-labels to adapt the pre-trained model. As shown in Fig.~\ref{fig: intro all visual}, the provided ground truth data is sparse and cannot provide valid annotation in the upper region of the scene. Instead, the proposed method can generate a denser disparity map as pseudo-labels, which can filter out most errors of UCFNet\_pretrain and cover all regions of the input picture. Consequently, the proposed method can tremendously improve the performance of our pre-training model on textureless area of foreground (red dash boxes) and unlabeled area of background (green dash boxes) by solely employing self-generated proxy labels as ground truth. More specifically, pixel-level and area-level uncertainty estimation are employed to generate reliable pseudo-labels. Given current disparity estimation results, we first employ pixel-level uncertainty estimation to quantify the degree to which the current disparity probability distribution tends to be multi-modal and employ it to evaluate the pixel-level confidence of current estimations. Then, area-level uncertainty estimation is proposed to leverage the multi-modal input and neighboring pixel information to further refine the initial uncertainty map. By the cooperation between pixel-level and area-level uncertainty estimation, we can obtain a denser and more robust pseudo label for domain adaptation without requiring cumbersome and expensive depth annotations.

Experimentally, we perform extensive experimental evaluations on various benchmarks to verify the generalization of the proposed method. When trained on synthetic datasets and generalized to unseen real-world datasets, our pre-trained model shows strong cross-domain generalization and can generate a good initial value for subsequent adaptation. Then, our model can further promote its performance by solely feeding the target domain synchronized stereo images and generated pseudo-labels, i.e., without the need for ground truth. In specific, the proposed method outperforms other domain generalization/adaptation methods by a noteworthy margin on various stereo matching benchmarks. The Qualitative comparison among GANet\_pretrain, UCFNet\_pretrain, and UCFNet\_adapt on three real datasets is shown in Fig. \ref{fig:  adapt generalization intro}. It can be seen from the figure that the generalization of current dataset-specific methods is limited to unseen real scenes, while our pre-training method can correct most errors and generate a more reasonable result. Moreover, compared with the pre-training model UCFNet\_pretraining, the proposed UCFNet\_adapt can achieve consistent improvement on multiple datasets with different characteristics, which further verifies the effectiveness of the generated pseudo-labels. More visualization results can be seen in the video demo of supplementary. Additionally, our uncertainty-based pseudo-labels can further be extended to replace the ground truth of monocular depth estimation networks and train these networks in an unsupervised way. Experiments show that the deep monocular depth estimation network trained by our pseudo-labels can outperform all self- supervised monocular depth estimation algorithms by a noteworthy margin and even achieves comparable performance with supervised methods. The code will be available at   \href{https://github.com/gallenszl/UCFNet}{https://github.com/gallenszl/UCFNet}.


In summary, our main contributions are:
\begin{itemize}[leftmargin=*]
\item We propose an uncertainty-based cascade and fused cost volume representation to reduce the domain differences and balance different disparity distributions across a variety of datasets. Thus, a robust pre-trained model with strong cross-domain generalization can be obtained.
\item We propose an uncertainty-based pseudo-labels generation method to further narrow down the domain gap. By employing the generated pseudo-labels to adapt our pre-trained model to the new domain, we can greatly promote the performance of our method.
\item Our method shows strong cross-domain and adapt generalization and outperforms other domain generalization/domain adaptation methods by a noteworthy margin on various stereo matching benchmarks.
\item Our method can perform well on multiple datasets with fixed model parameters and hyperparameters and obtains 1st place on the stereo task of Robust Vision Challenge 2020 \footnote{\href{http://www.robustvision.net/rvc2020.php}{http://www.robustvision.net/rvc2020.php}}.
\item Our uncertainty-based pseudo-labels can further be extended to train monocular depth estimation networks in an unsupervised way and even achieves comparable performance with supervised methods.
\end{itemize}


\textbf{Differences with conference version \cite{cfnet}:} This paper extends the early ideas and findings presented in CFNet\cite{cfnet}. The differences with our conference paper can be summarized as follows:
\begin{itemize}[leftmargin=*]
\item In our previous work, we only focus on cross-domain generalization and joint generalization of stereo matching tasks. Here, we provide a general solution for cross-domain, joint, and adaptation generalization jointly by digging into uncertainty estimation in stereo matching. Hence, a more complete and standard solution is presented for robust stereo matching.  
\item In our previous work, the final disparity estimation is just half-resolution of the input image and needs to be upsampled to the original image size. Thus, we propose a simple, yet effective attention-based refinement module to recover the details loss caused by the bilinear sampling.
\item We extend our uncertainty-based pseudo-labels to train the monocular depth estimation network in an unsupervised way. Experiments show that the deep monocular depth estimation network trained by our pseudo-labels can outperform all self-supervised monocular depth estimation algorithms by a noteworthy margin and even achieves comparable performance with the supervised methods.
\end{itemize}

\section{Related Work}

\subsection{Multi-scale Cost Volume based Stereo Matching}
Cost volume construction is an indispensable step in the well-known four-step pipeline for stereo matching \cite{scharstein2002taxonomy, pamisurvey1, pamisurvey2}. Typically, current state-of-the-art stereo matching methods can be categorized into two types of cost volume-based methods, where the cost volume is a 4D tensor of height, width, disparity, and features. The first category usually uses the single-feature 3D cost volume generated by full correlation, which is efficient while losing much information due to the decimation of feature channels. Many real-time methods, such as Dispnet \cite{dispnet}, MADNet \cite{madnet, madnet_pami} and AANet \cite{aanet}, belongs to the category. Moreover, two-stage refinement \cite{mcvmfc} and pyramidal towers \cite{madnet} are commonly applied in the single-feature cost volume based network to construct multi-scale cost volume. The second category usually uses the multi-feature 4D cost volume generated by concatenation \cite{gcnet} or group-wise correlation \cite{gwcnet}, which can achieve better performance with higher computational complexity and memory consumption. Most top-performing networks, including GANet \cite{ganet}, CSPN \cite{cspn} and ACFNet \cite{acfnet} belong to this category. 
Recently, to alleviate the high computational complexity and memory consumption when employing multi-feature 4D cost volumes, \cite{cvpmvsnet, cascade, uscnet} propose to use cascade cost volume representation in multi-view stereo. These methods usually first predict an initial disparity at the coarsest resolution of the image and then gradually refine the disparity by narrowing down the disparity search space. More closely related to our approach is Casstereo \cite{cascade}, which first extended such representation to stereo matching. It selected to uniform sample a pre-defined range to generate the next stage’s disparity search range. Instead, we employ pixel-level uncertainty estimation to adaptively adjust the next stage disparity searching range and generate pseudo-labels for subsequent domain adaptation. Our method also shares similarities with UCSNet \cite{uscnet}, which constructs uncertainty-aware cost volume in multi-view stereo while it doesn’t employ uncertainty estimation to generate pseudo-labels.

\begin{figure*}[!t]
 \centering
 \includegraphics[width=0.8 \linewidth]{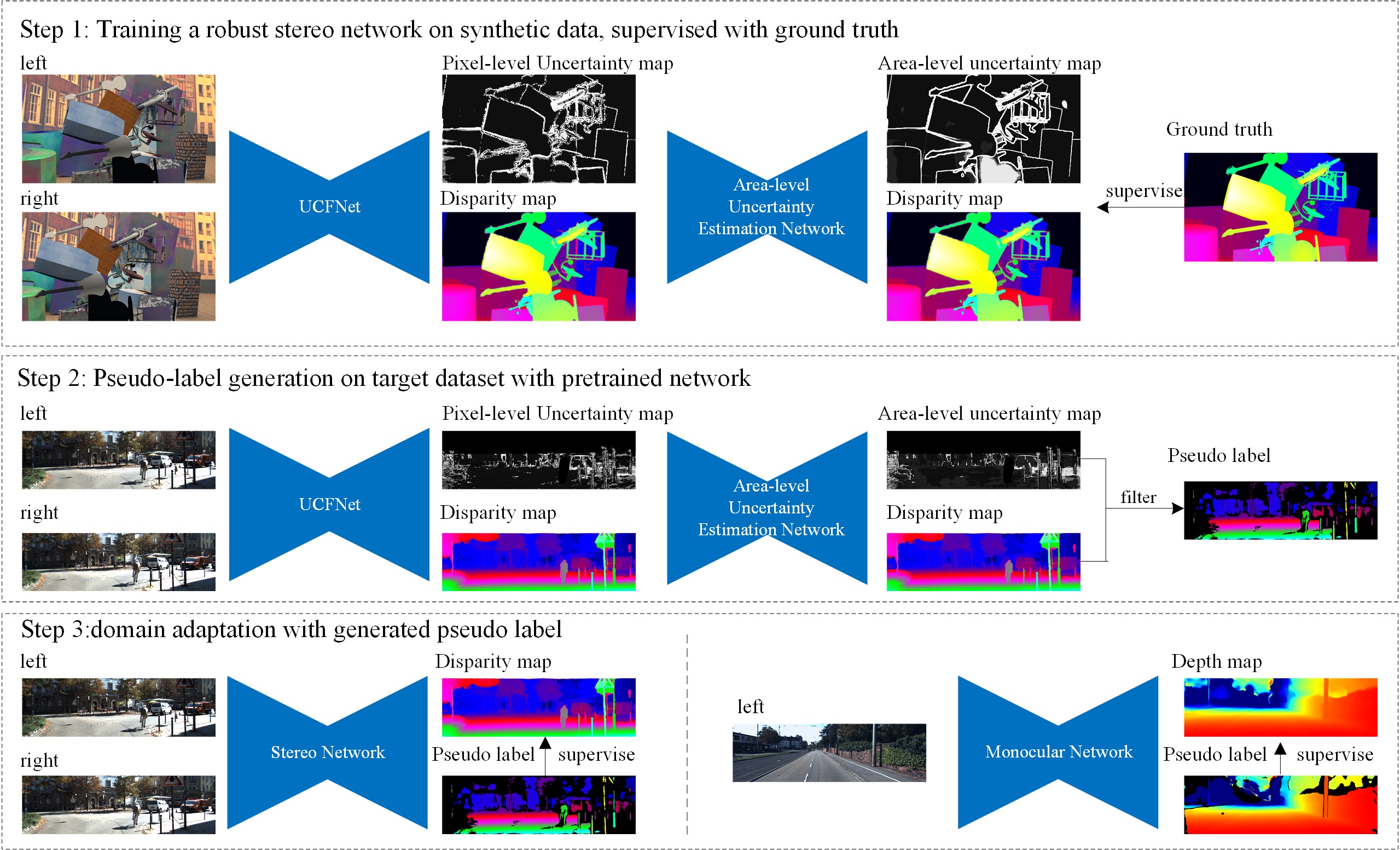} 
 \vspace{-0.1in}
 \caption{ \footnotesize The overall framework of our proposed network, which consists of 3 steps: Training a robust stereo matching network on the source domain, Pseudo-label generation on the target domain, and Domain adaptation with generated Pseudo-label. UCFNet denotes the proposed \textbf{U}ncertainty-based \textbf{C}ascade and \textbf{F}used cost volume representation.}
 \label{fig: UCFNet}
 \vspace{-0.1in}
\end{figure*}

\subsection{Robust Stereo Matching} 
There exist three categories of generalization definitions for robust stereo matching. 1) Cross-domain Generalization: the network’s ability to perform well on unseen scenes (cannot see the image pairs of the target domain in advance). Towards this end, Jia et al \cite{sungeneralizaiton} propose to incorporate scene geometry priors into an end-to-end network. Zhang et al \cite{dsmnet} introduce a domain normalization and a trainable non-local graph-based filter to construct a domain-invariant stereo matching network. 2) Adapt Generalization: the network’s ability to adapt pre-trained models to the new domain with unlabeled target data. Previous work usually pre-trains the models on synthetic data and then adapts it to new target domains with Graph Laplacian regularization \cite{zoom}, non-adversarial progressive color transfer \cite{adastereo}, and Knowledge Reverse Distillation \cite{aohnet}. More closely related to our approach are \cite{aohnet, unsuperviseddomainadaptation} in stereo matching and Monoresmatch \cite{monoresmatch} in monocular depth estimation, which also proposes to generate a pseudo-label for domain adaptation. However, these methods all select to employ classical stereo matching methods \cite{sgm} alongside with confidence estimators, e.g., left-right consistency check to generate pseudo-labels. That is all these methods need an independent method to generate corresponding pseudo-labels. Instead, the proposed method is an end-to-end network that can generate the predicted disparity map, corresponding uncertainty map and pseudo-labels jointly, which is a more simple, yet efficient way. 
3) Joint Generalization: the network’s ability to perform well on a variety of datasets with the same model parameters. MCV-MFC \cite{mcvmfc} introduces a two-stage finetuning scheme to achieve a good trade-off between generalization and fitting capability on multiple datasets. However, it doesn’t touch the inner difference between diverse datasets, e.g, the unbalanced disparity distribution. To further address this problem, we propose a cascade cost volume to adaptively the next stage disparity searching space, where the pixel-level uncertainty estimation is at the core.


\section{Our Approach}

\subsection{Framework Overview}

In this paper, we provide a general solution for cross-domain generalization, joint generalization, and adaptation generalization jointly by digging into uncertainty estimation in stereo matching. The overall architecture of our method is shown in Fig. \ref{fig: UCFNet}, which can divide into three steps:

\textbf{1) Training a robust stereo matching network on source domain:} Given stereo image pairs and corresponding ground-truth disparity of source domain (synthetic dataset), we first propose to employ the synthetic data to train a robust pre-train model with strong cross-domain generalization. Specifically, an \textbf{U}ncertainty-based \textbf{C}ascade and \textbf{F}used cost volume representation (UCFNet) is proposed to alleviate the unbalanced disparity distribution and large domain shifts across different datasets. 

\textbf{2) Pseudo-label generation on target domain:} After getting the pre-training model, we propose an uncertainty-based pseudo-label generation method to generate reliable pseudo-labels for domain adaptation. Specifically, the proposed method can be divided into two steps: (a) Given stereo image pairs of the target domain (real dataset), the pre-trained model is employed to predict the corresponding disparity estimation. (b) Two terms of uncertainty estimation, i.e., pixel-level and area-level are proposed to filter out the high-uncertainty pixels of the disparity estimation and generate sparse while reliable disparity maps as pseudo-label.

\textbf{3) Domain adaptation with generated Pseudo-label:} After getting the generated pseudo-label, we can employ it as supervision to adapt the pre-train model to the target domain. In addition, the generated pseudo-label can also be employed as supervision to train the Monocular depth estimation network in an unsupervised way. Experiments show the superiority of the proposed method on both monocular and binocular depth estimation tasks.

\begin{figure*}[!t]
 \centering
 \includegraphics[width=0.8 \linewidth]{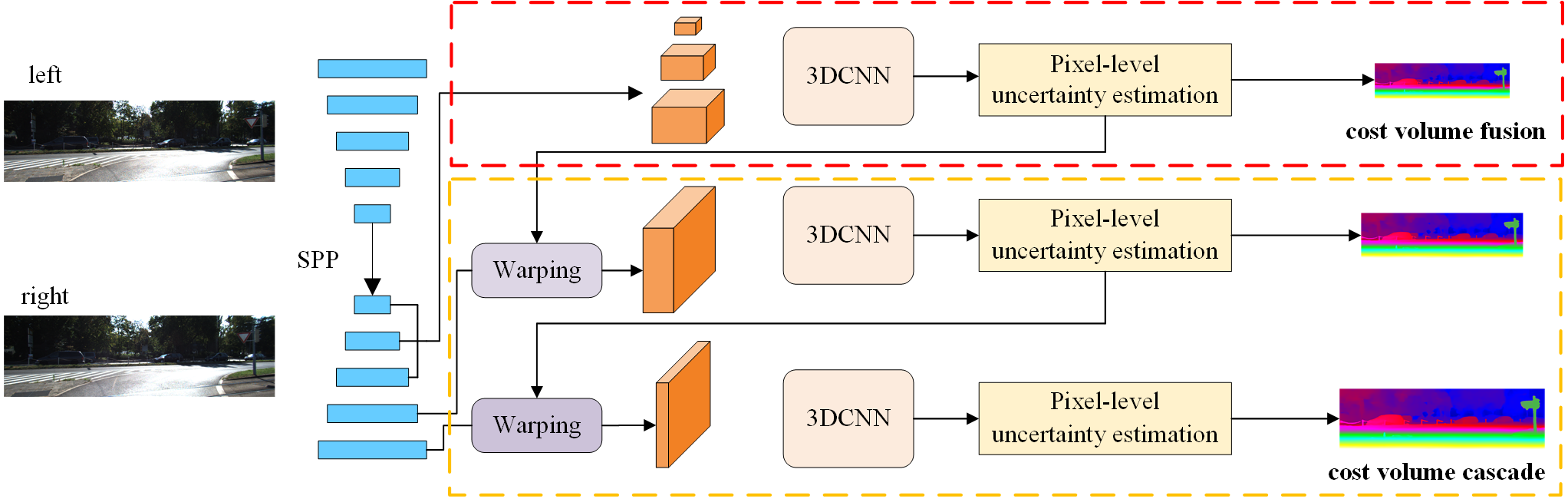} 
 \vspace{-0.1in}
 \caption{\footnotesize The architecture of the proposed \textbf{U}ncertainty-based \textbf{C}ascade and \textbf{F}used cost volume representation. Our network consists of 3 parts: pyramid feature extraction, fused cost volume, and cascade cost volume.}
 \vspace{-0.15in}
 \label{fig: CFNet}
\end{figure*}

The structure of this paper is organized as follows. In Sec.~\ref{sec:stereo_network}, we present the details about how to employ the \textbf{U}ncertainty-based \textbf{C}ascade and \textbf{F}used cost volume representation (UCFNet) for robust disparity estimation. Sec. \ref{sec:uncertainty} introduces the design of uncertainty estimation, which can filter out unreliable points of current estimations and generate reliable and sparse pseudo-label for subsequent domain adaptation. Sec. \ref{sec:domain_adaptation} introduces the mechanism of domain adaptation, i.e., how to employ the generated pseudo-label adapting the binocular/monocular depth estimation network to the new domain. Finally, we evaluate the results of our algorithms on both stereo matching and monocular depth estimation tasks in Sec. \ref{sec:stereo_experinment} and Sec. \ref{sec:monocular}, respectively.

\subsection{Uncertainty based cascade and fused cost volume for disparity estimation}
\label{sec:stereo_network}

To achieve robust stereo matching, we propose an \textbf{U}ncertainty-based \textbf{C}ascade and \textbf{F}used cost volume representation (UCFNet) to alleviate the unbalanced disparity distribution and large domain shifts across different datasets. As shown in Fig. \ref{fig: CFNet} and \ref{fig:channel_attention}, the proposed UCFNet consists of four parts, including pyramid feature extraction, fused cost volume, cascade cost volume, and attention-based disparity refinement. 

\subsubsection{Pyramid feature extraction}


Given an image pair, an unet-like \cite{hsm,unet} encoder-decoder architecture is first proposed to extract multi-scale image features. Specifically, the encoder consists of five residual blocks, followed by an SPP \cite{psmnet} module to better extract hierarchical context information. Compared with the widely used Resnet-like network \cite{cascade,gwcnet}, experiments show that the proposed unet-like feature extraction can preserve sufficient information with lower computational complexity. Then, the extracted multi-scale features can be divided into fused and cascade cost volumes and predict corresponding resolution disparity, respectively.

\begin{figure}[!t]
 \centering
 \includegraphics[width=0.85 \linewidth]{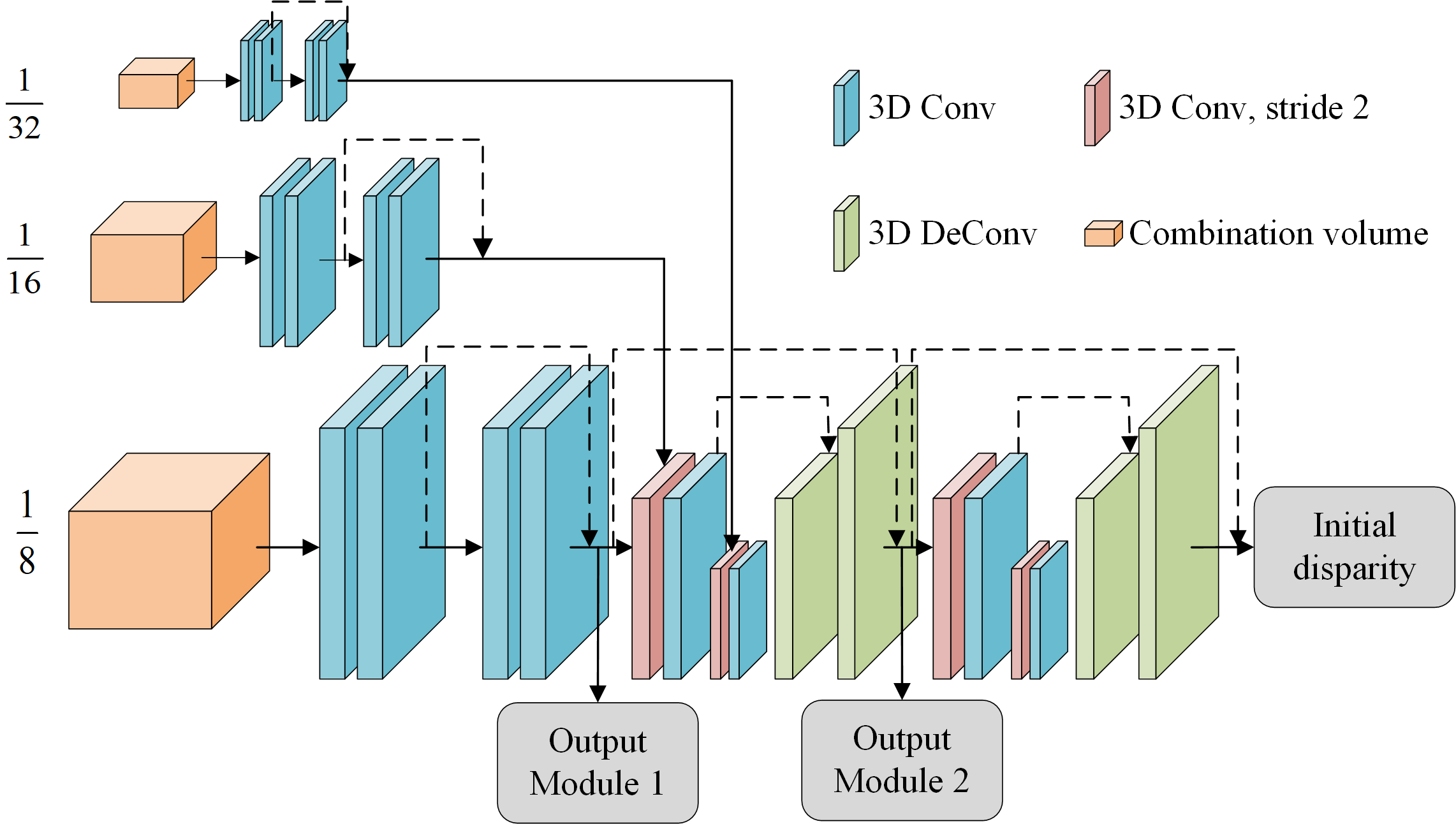} 
 \vspace{-0.1in}
 \caption{\footnotesize The architecture of our cost volume fusion module. Three low-resolution cost volumes ($i \in (3,4,5)$) are fused to generate the initial disparity map. }
 \label{fig: fusion}
\vspace{-0.1in}
\end{figure}

\subsubsection{Fused Cost Volume}


In this section, multiple low-resolution dense cost volumes are fused together to reduce the domain shifts across different datasets for initial disparity estimation. Our method is motivated by a simple observation that multi-scale cost volume can cover multi-scale receptive fields and drive the network to extract multi-level information, e.g., edges and areas are easier to be captured by low-resolution cost volume. Moreover, edges and areas are no-local information, which is less sensitive to domain changes. Hence, we can fuse multiple low-resolution dense cost volumes to incorporate hierarchical structural representations and generate a more accurate initial disparity estimation. Specifically, we first employ the input multi-scale features (smaller than 1/4 of the original input image resolution) to construct each scale cost volume respectively and then design a cost volume fusion module to integrate them. Details of the two steps will be provided below.


\textbf{Cost volume construction:} Inspired by \cite{msmdnet,gwcnet}, feature concatenation and group-wise correlation are employed to generate corresponding combination volume as follows:
\begin{eqnarray}
\begin{array}{c}
V_{concat}^i({d^i},x,y,f) = f_L^i(x,y)||f_R^i(x - {d^i},y) \\
\\
V_{gwc}^i({d^i},x,y,g) = \frac{1}{{N_c^i/{N_g}}}\left\langle {f_l^{ig}(x,y),f_r^{ig}(x - {d^i},y)} \right\rangle \\
\\
V_{combine}^i = V_{concat}^i||V_{gwc}^i
\end{array}
\label{eq:volume construction}
\end{eqnarray}
where $||$ denotes the vector concatenation operation. ${N_c}$ represents the channels of extracted features. ${N_g}$ is the amount of group. 
$\left\langle {{\rm{ }},{\rm{ }}} \right\rangle$ represents the inner product. ${f^i}$ denotes the extracted feature at scale (stage) $i$ and  $i = 0$ represents the original input image resolution.
 
Note that the disparity searching index ${d^i}$ is defined as 
${d^i} \in \{ 0,1,2 \ldots \frac{{{D_{\max }}}}{{{2^i}}} - 1\}$ and the hypothesis plane interval equals to 1 in the fused cost volume representation. That is, these cost volumes are all dense cost volumes with the size of  $\frac{H}{{{2^i}}} \times \frac{W}{{{2^i}}} \times \frac{{{D_{\max }}}}{{{2^i}}} \times F$. By densely sampling the whole disparity range in small resolution, we can efficiently generate the coarsest disparity map. Then pixel-level uncertainty estimation is employed to narrow down the disparity searching space at higher resolution and refine the disparity estimation in a coarse-to-fine manner. Please refer to Section 3.2.3 for more detail.


\textbf{Cost Volume fusion:} The architecture of cost volume fusion is shown in Fig. \ref{fig: fusion}. Specifically, we first employ four 3D convolution layers with skip connections to regularize each cost volume. Then, a 3D convolution layer (stride of two) is employed to downsample the scale 3 cost volume from 1/8 to 1/16 of the original input image size. Next, we concatenate the down-sampled cost volume and the next scale combination volume at the feature dimension and use one additional 3D convolution layer to decrease the feature channel to a fixed size. Similar operations are progressively employed until we downsample the cost volume to 1/32 of the original input image size. Finally, a 3D transposed convolution is adopted to up-sample the volume in the decoder and one 3-D hourglass network is further employed to aggregate the cost volume. Moreover, an output module is applied to predict the disparity from the fused cost volume. Specifically, we first employ two more 3D convolution layers to obtain a 1-channel 4D volume. Then, soft argmin \cite{gcnet} operation is applied to transform volume into probability and generate the initial disparity map ${D^3}$. The soft argmin operation is defined as:
\begin{eqnarray}
\widehat {{d^i}}{\rm{ = }}\sum\limits_{d = 0}^{\frac{{{D_{\max }}}}{{{2^i}}} - 1} {d \times \sigma ( - c_d^i)},
\label{softmax}
\end{eqnarray}
where $\sigma$ denotes the softmax operation and $c$ represents the predicted 1-channel 4D volume. ${\sigma ( - c_d)}$ denotes the discrete disparity probability distribution and the estimated disparity map is susceptible to all disparity indexes.

%
%
%
%

\subsubsection{Cascade Cost Volume}
\label{cascade_cost}

Given the initial disparity estimation, the next step is to construct a fine-grained cost volume and refine disparity maps in a coarse-to-fine manner. One naive way to construct the next stage disparity searching range is uniform sampling a pre-defined searching range \cite{cascade}. However, such a method treats all pixels equally and cannot make pixel-level adjustments. Furthermore, the unbalanced disparity distribution across different datasets requests networks to adjust the disparity searching range according to the input image adaptively. Hence, a question arises, can we drive the network to filter out invalid disparity indexes in a large disparity searching range and capture more possible pixel-level disparity searching space with prior knowledge of the last stage’s disparity estimation?

\begin{figure}[!t]
	\centering
	\tabcolsep=0.05cm
	\begin{tabular}{c c c}

	\includegraphics[width=0.31\linewidth]{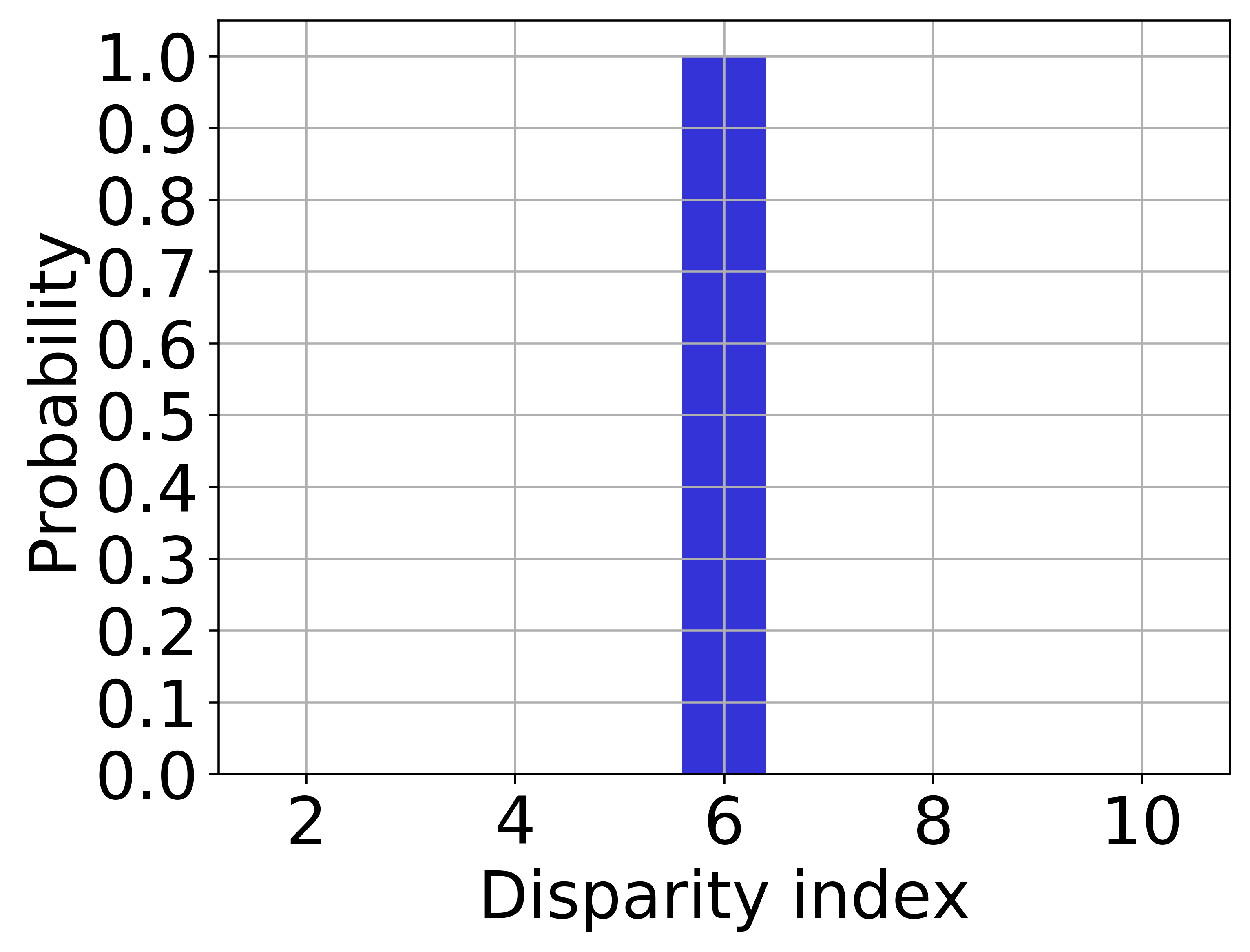}&
    \includegraphics[width=0.31\linewidth]{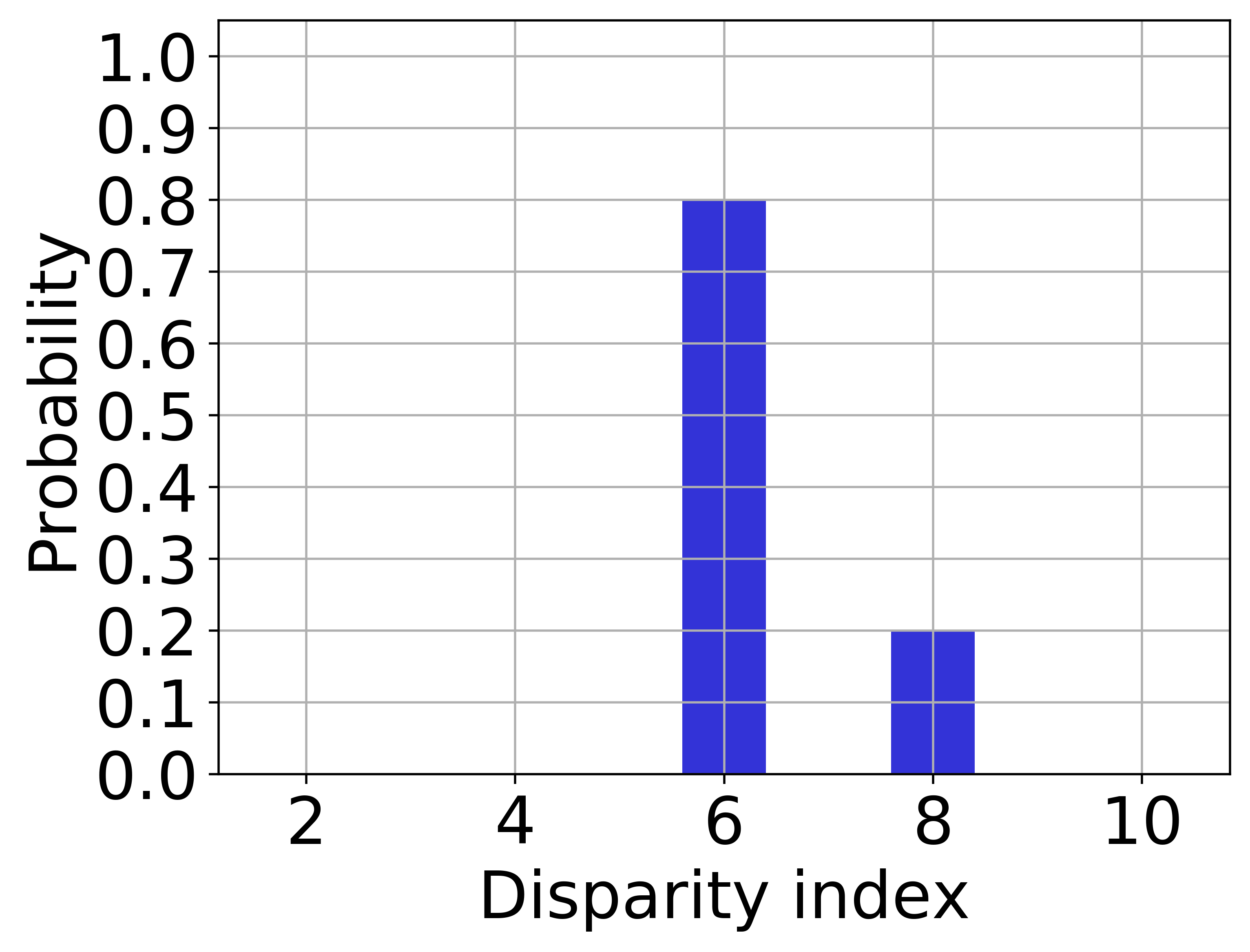}&
     \includegraphics[width=0.31\linewidth]{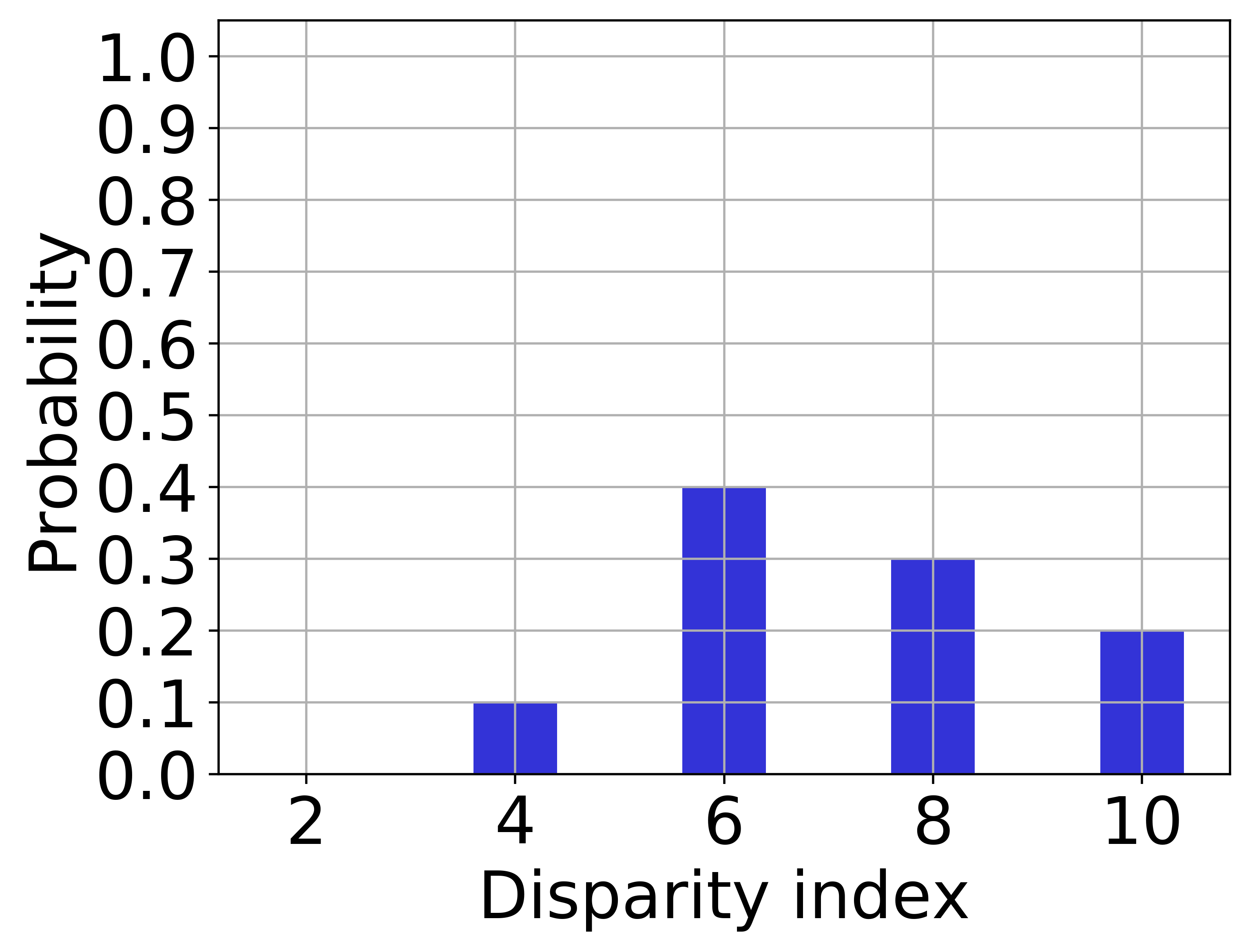}\\

    {     \scriptsize$\hat d = 6.0,U = 0.0$} &  {     \scriptsize$\hat d = 6.4,U = 0.64$} &  {     \scriptsize$\hat d = 7.2,U = 3.36$}\\
    {\scriptsize(a)Unimodal} &  {\scriptsize(b)Predominantly unimodal} & {\scriptsize(c)Multi-modal}


	\end{tabular}
 	\vspace{-0.1in}
	\caption{\footnotesize Some samples of pixel-level uncertainty estimation. Expected value (ground truth) is 6px. The disparity searching range is from 2 to 10 with 5 hypothesis planes.}
 	\vspace{0.1in}
	\label{fig: uncertainty estimation sample}

	\centering
	\tabcolsep=0.05cm
	\begin{tabular}{c c}

	\includegraphics[width=0.48\linewidth]{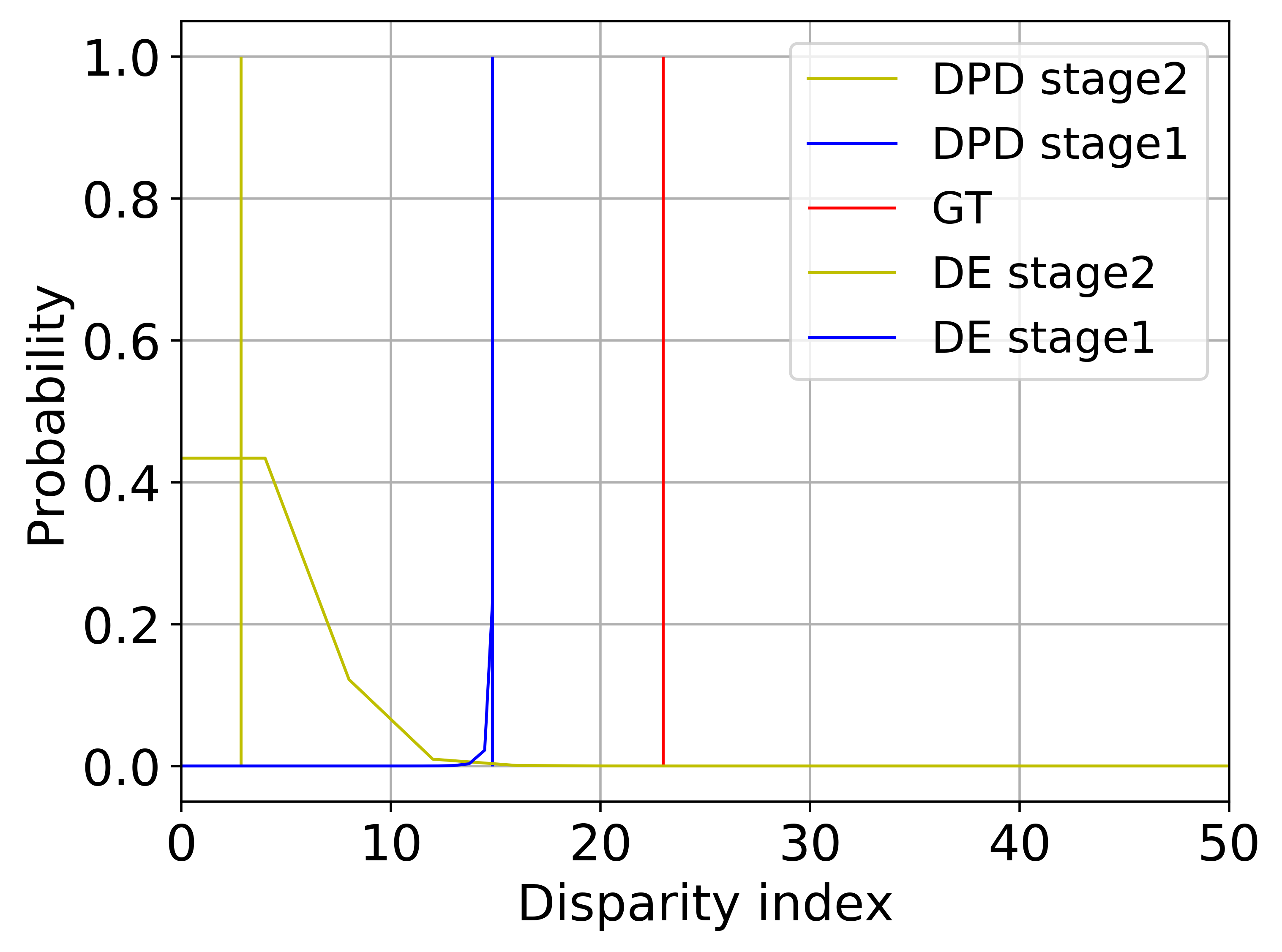}&
	\includegraphics[width=0.48\linewidth]{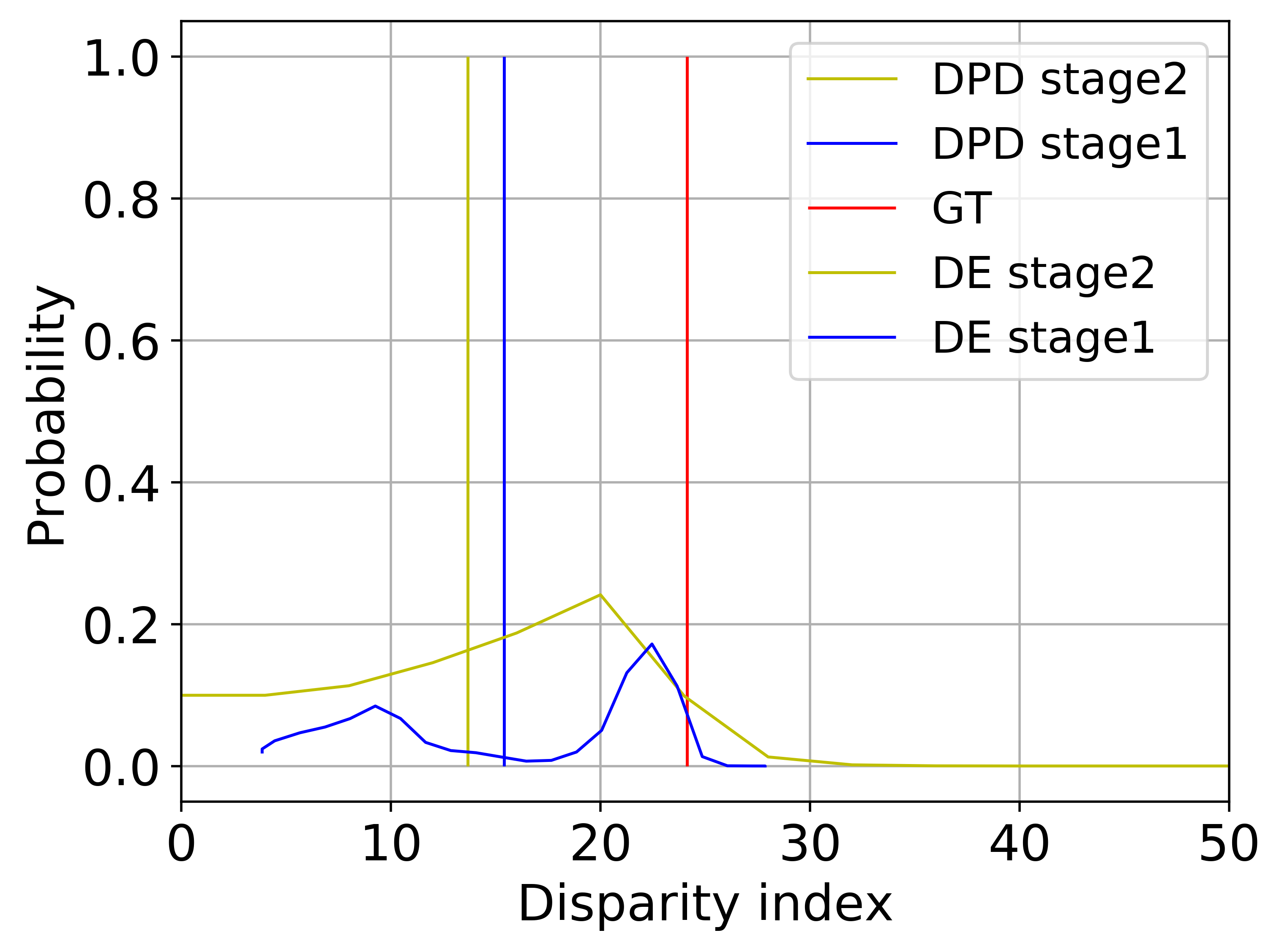}\\

	\includegraphics[width=0.48\linewidth]{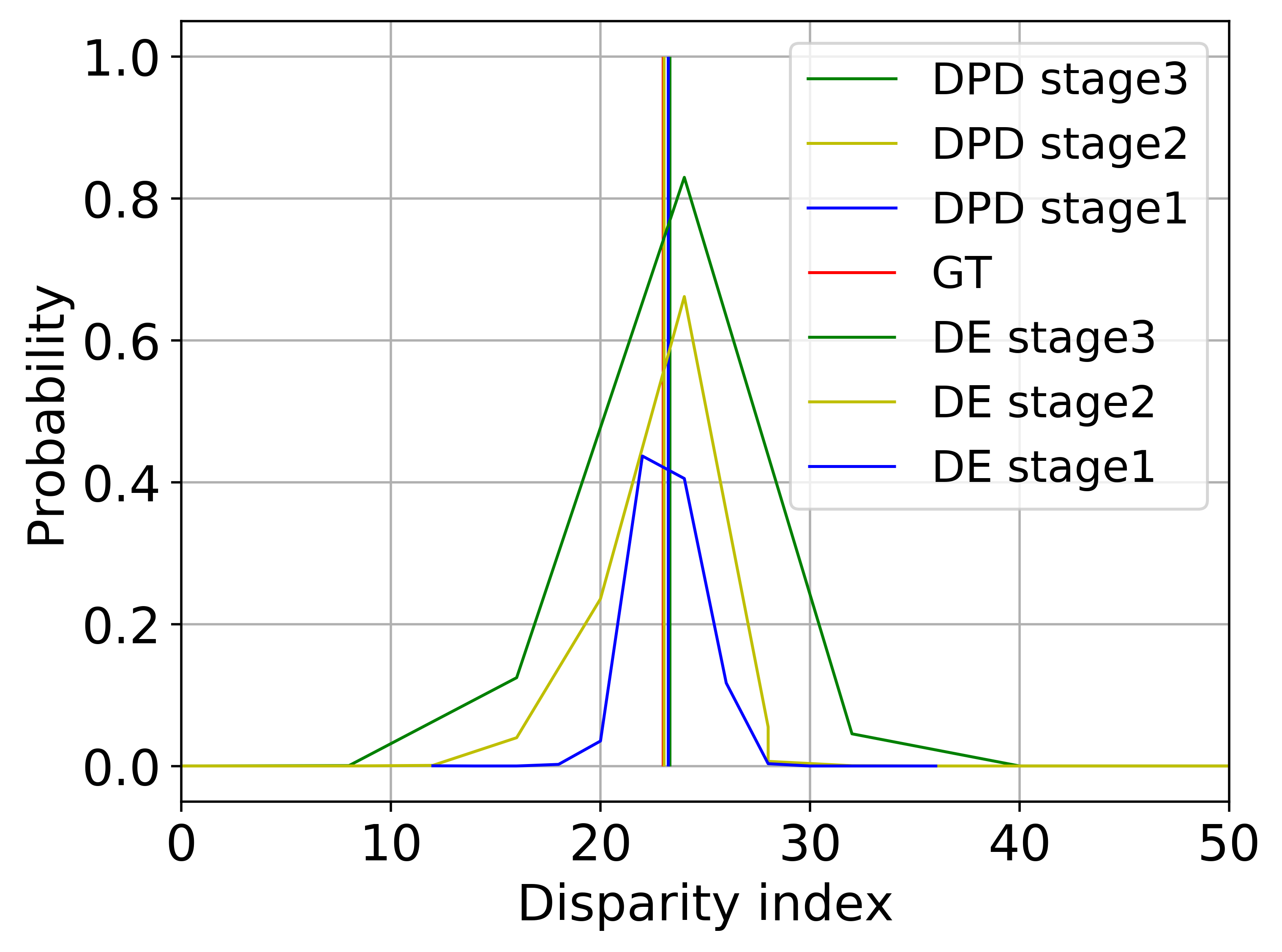}&
	\includegraphics[width=0.48\linewidth]{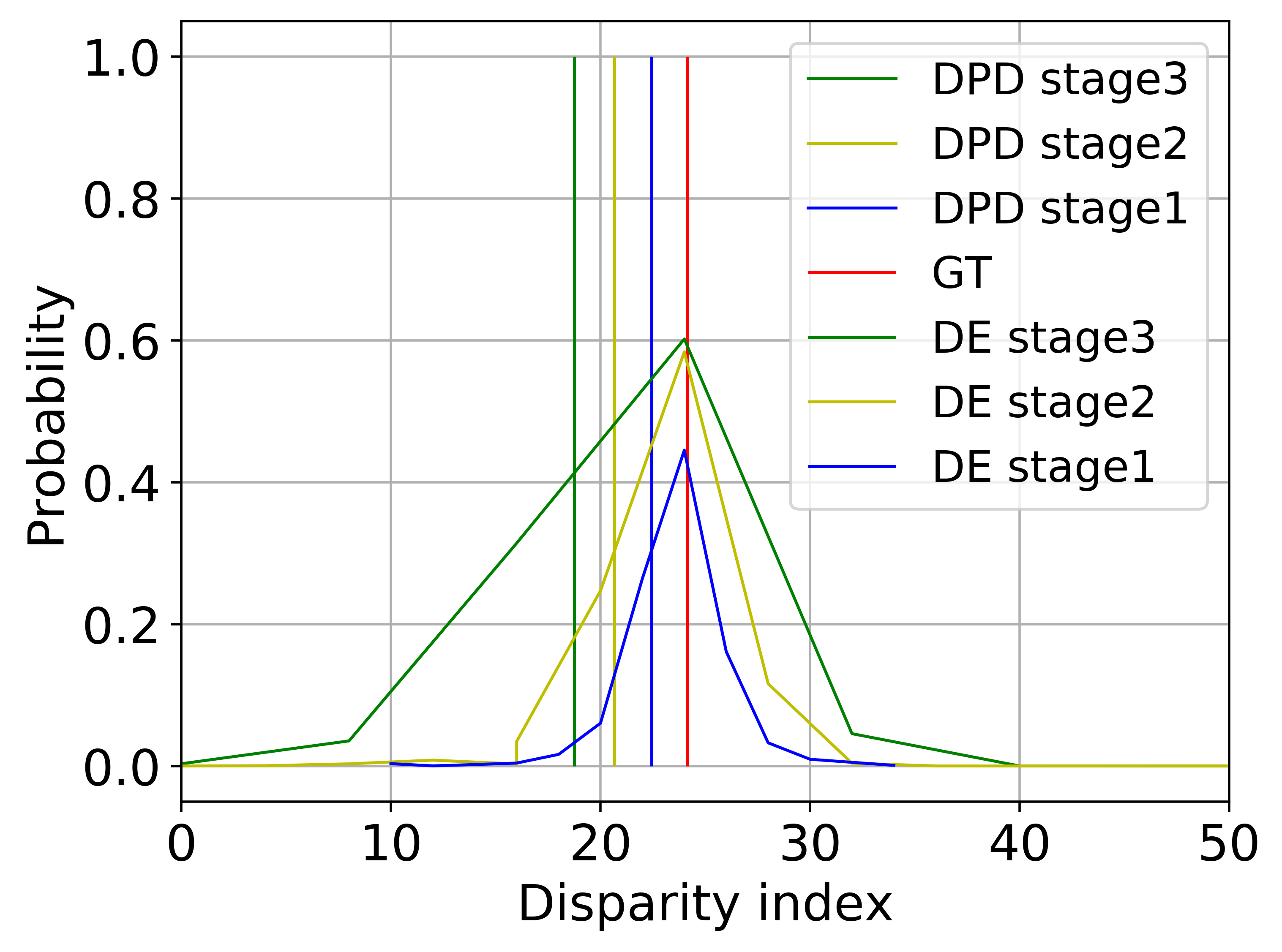}\\

    {(a)} &  {(b)}

	\end{tabular}

\vspace{-0.1in}
	\caption{\footnotesize Some examples of the disparity probability distribution of cascade stereo (first row) and UCFNet (second row). DPD: disparity probability distribution, DE: disparity estimation, GT: ground truth. Final disparity estimation is the estimation of stage 1.
}

	\label{fig: cost volume distribution}
  	\vspace{-0.1in}
\end{figure}

To tackle this problem, we propose a pixel-level uncertainty estimation to adaptively adjust the disparity searching range. As mentioned in Eq. \ref{softmax}, the final predicted disparity can be obtained by softly weighting indices according to their probability. Thus, the discrete disparity probability distribution indeed reflects the similarities between candidate matching pixel pairs and the ideal disparity probability distribution should be unimodal peaked at true disparities. However, the actual probability distribution is predominantly unimodal or even multi-modal at some pixels, e.g., ill-posed and occluded areas. Moreover, existing methods \cite{acfnet,gcnet} have observed that the degree of multimodal distribution is highly correlated with the probability of prediction error. Hence, we propose to define a pixel-level uncertainty estimation to quantify the degree to which the cost volume tends to be multi-modal distribution and employ it to evaluate the confidence of the current estimation. The pixel-level uncertainty is defined as:
\begin{eqnarray}
\begin{array}{c}
{{\rm{U}}^i}{\rm{ = }}\sum\limits_{\forall {d^i}} {{{(d - {{\hat d}^i})}^2} \times \sigma ( - c_d^i)} \\
\\
\widehat {{d^i}}{\rm{ = }}\sum\limits_{\forall {d^i}} {d \times \sigma ( - c_d^i)} 
\end{array}
\end{eqnarray}
where $\sigma$ denotes the softmax operation and $c$ represents the predicted 1-channel 4D volume. Fig. \ref{fig: uncertainty estimation sample} gives a toy sample to show the effectiveness of pixel-level uncertainty estimation. As shown, the uncertainty of unimodal distribution equals to 0 and the more the distribution tends to be multimodal, the higher the error and uncertainty.
Thus, we can employ pixel-level uncertainty to evaluate the confidence of disparity estimation, higher uncertainty implies a higher probability of prediction error and a wider disparity searching space to correct the wrong estimation. Then, the next stage’s disparity searching range can be defined as:
\begin{eqnarray}
\begin{array}{l}
d_{\max }^{i - 1} = \delta (\widehat {{d^i}} + \left( {{\alpha ^i} + 1} \right)\sqrt {{U^i}}  + {\beta ^i})\\
\\
d_{\min }^{i - 1} = \delta (\widehat {{d^i}} - \left( {{\alpha ^i} + 1} \right)\sqrt {{U^i}}  - {\beta ^i})
\end{array}
\end{eqnarray}
where $\delta$ denotes bilinear interpolation. $\alpha$ and $\beta$ are normalization factors, which are initialized as 0 and gradually learn a weight. Then, uniform sampling can be employed to get the next stage discrete hypothesis disparity indexes ${d^{i - 1}}$:
\begin{eqnarray}
\begin{array}{c}
{d^{i - 1}} = d_{\min }^{i - 1} + n(d_{\max }^{i - 1} - d_{\min }^{i - 1})/\left( {{N^{i - 1}} - 1} \right)\\
\\
n \in \{ 0,1,2 \ldots {N^{i-1}} - 1\}
\end{array}
\end{eqnarray}
where ${N^{i - 1}}$ is the number of hypothesis planes at stage $i-1$. Then, a sparse while fine-grained cost volume at stage $i-1$ ($\frac{H}{{{2^{i - 1}}}} \times \frac{W}{{{2^{i - 1}}}} \times {N^{i - 1}} \times F$) can be constructed based on Eq.\ref{eq:volume construction}.  After
getting the next stage cost volume, a similar cost aggregation network (omitting the solid line in Fig. \ref{fig: fusion}) can be employed to predict the corresponding stage disparity map. By iteratively narrowing down the disparity range and higher the cost volume resolution, we can refine the disparity in a coarser to fine manner. Note that the final output of cascade cost volume $\hat{d^{1}}$ is half resolution of the original image. Thus, an up-sampling operation is necessary to up-sample $\hat{d^{1}}$ to the same size of original images, i.e., $\hat{d^{0}} = up(\hat{d^{1}})$, where the up-sampling operation \emph{up} is implemented by bilinear interpolation.

In summary, the proposed UCFNet outperforms previous cascade-based approaches, i.e., casstereo \cite{cascade} in the following three aspects: First, we propose to fuse multiple dense low-resolution cost volumes to generate a more accurate initial disparity estimation at lower resolution (see the comparison between the estimation of casstereo at stage 2 and our UCFNet at stage 3 in Fig. \ref{fig: cost volume distribution} (a) and (b)). Second, pixel-level uncertainty estimation is proposed to adaptively adjust the next stage disparity searching range which can push disparity distribution to be more predominantly unimodal (Fig. \ref{fig: cost volume distribution}(b)). Third, our method can better cover the corresponding ground truth value in the final stage disparity searching range by the proposed Uncertainty-based Cascade and Fused cost volume representation and corrects some biased results in casstereo (Fig. \ref{fig: cost volume distribution}(a)).


\begin{figure}[!t]
    \centering
    \includegraphics[width=\linewidth]{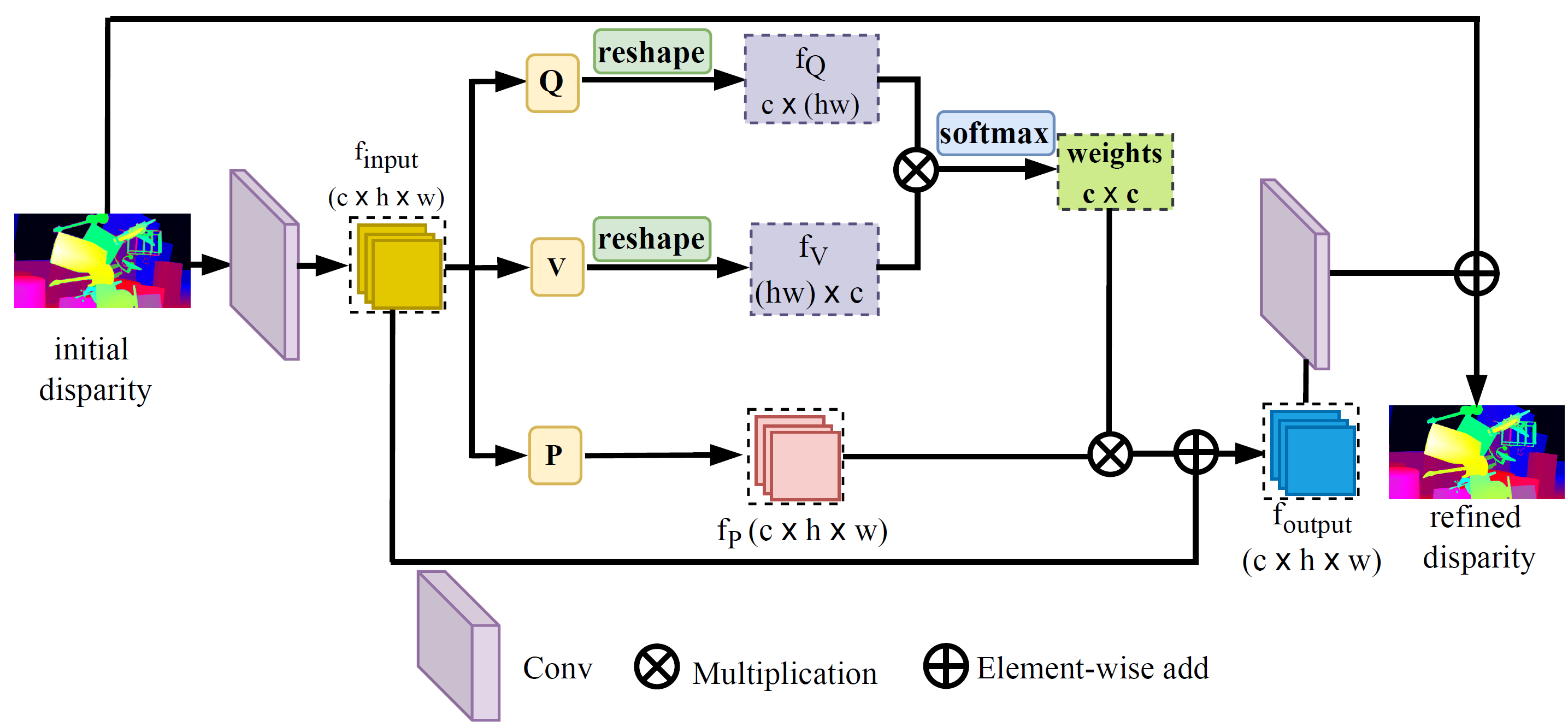}
    \vspace{-0.3in}
    \caption{\footnotesize Detailed architecture of the proposed attention-based disparity refinement module.}
    \label{fig:channel_attention}
\vspace{-0.1in}
\end{figure}

\subsubsection{Attention-based disparity refinement}

As mentioned in Section \ref{cascade_cost}, the output of our cascade cost volume $\hat{d^{0}}$ is up-sampled from the half-resolution disparity map $\hat{d^{1}}$ by bilinear interpolation. However, such direct upsampling operations will lead to the degradation of texture information, which indeed hinders both the finetuning performance and generalization of the proposed method. Hence, we propose a lightweight attention-based disparity refinement module to make up for the missing details. Experiments in Tab.1\&2\&3 demonstrate the proposed refinement network can achieve consistent improvement in both generalization and finetuning performance across multiple datasets. 

The pipeline of the attention mechanism is shown in Fig.~\ref{fig:channel_attention}. Taking up-sampled disparity $\hat{d^{0}}$ as input, we first employ multiple stacked convolution layers to extract the deep feature representation $f_{input}$. Then, three encoding operations \emph{P(.)}, \emph{Q(.)} and \emph{V(.)} are used to convert $f_{input}$ to three components ${f_p}$, ${f_Q}$ and ${f_V}$, in which reshape operation is utilized to convert the shape of ${f_Q}$, ${f_V}$ to ${f_Q} \in [C \times HW]$ and ${f_V} \in [HW \times C]$: 
\begin{eqnarray}
\begin{array}{l}
{f_{input}} = Conv({\hat{d^{0}}}),\\
{f_p} = P({f_{input}}),\\
{f_Q} = reshape(Q({f_{input}})),\\
{f_V} = reshape(V({f_{input}})),
\end{array}
\end{eqnarray}
where \emph{Conv} means convolution operation. Then, a matrix multiplication $\otimes$ and a softmax operation are introduced to generate the attention weight, which reflects the similarity between each channel position of input feature map $f_{input}$. Next, we employ a matrix multiplication operation between weights and $f_p$ with a 2D convolution layer to generate the residual disparity:
\begin{eqnarray}
\hat {d_{residual}^0} = Conv(Weight \otimes {f_p} + {f_{input}}),
\end{eqnarray}
where \emph{Conv} means convolution operation. Finally, an element-wise addition operation is employed to generate the refined disparity:
\begin{eqnarray}
\hat {d_{refine}^0} = \hat {d_{residual}^0} + \hat {{d^0}}
\end{eqnarray}

\begin{figure}[!htb]
	\centering
	\tabcolsep=0.05cm
	\begin{tabular}{c c}
    \includegraphics[width=0.49\linewidth]{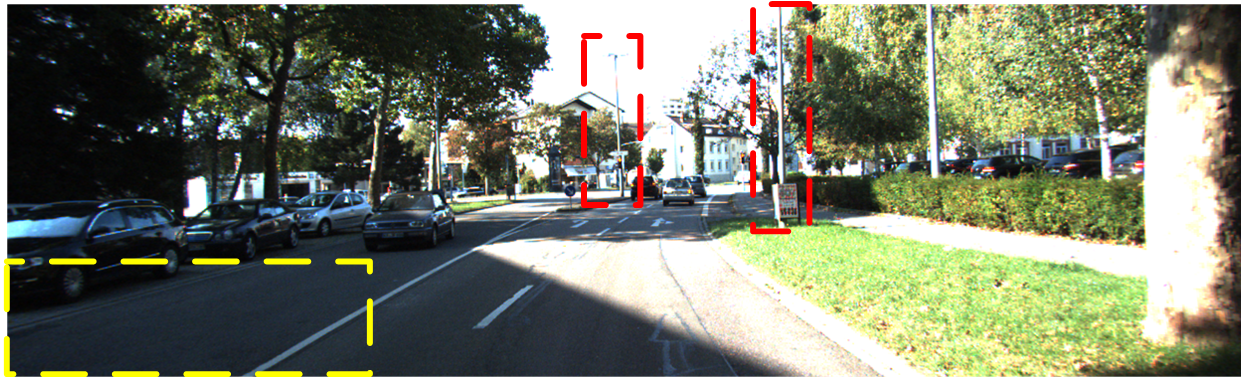}&
	\includegraphics[width=0.49\linewidth]{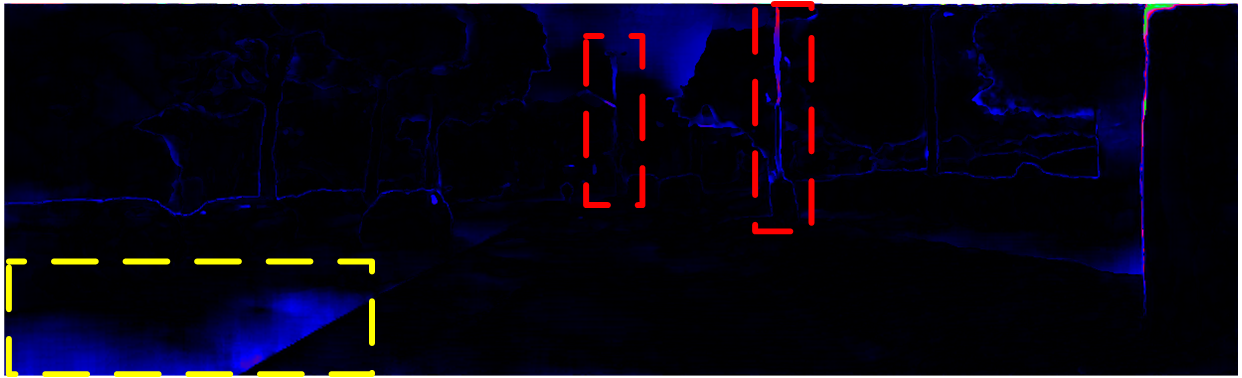} \\
	
	{\small (a) left image} &  {\small (b) residual disparity map }		 \\

    \includegraphics[width=0.49\linewidth]{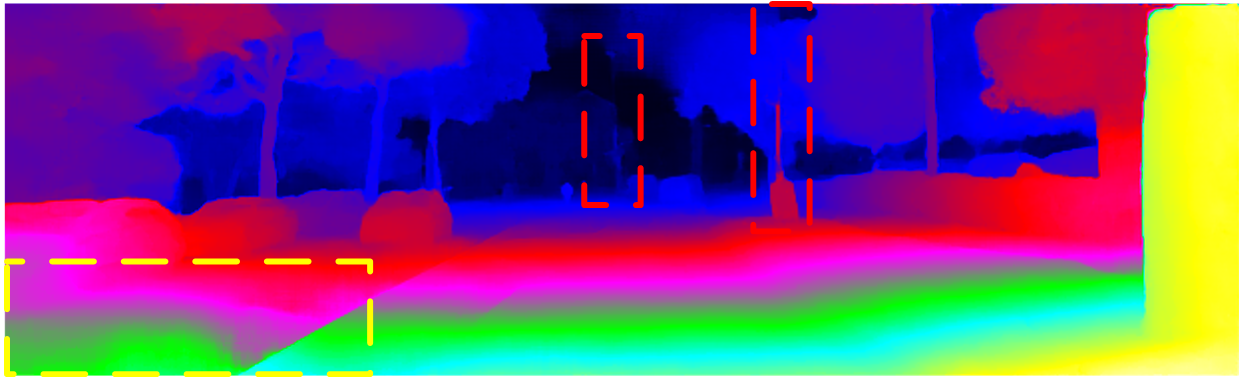}&
	\includegraphics[width=0.49\linewidth]{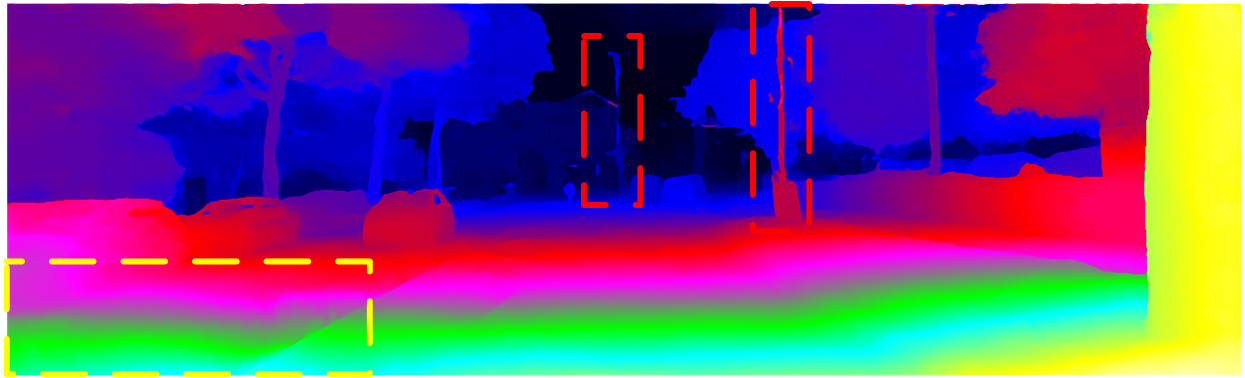} \\
	
	{\small (c) original disparity map} &  {\small (d) refined disparity map }		 \\

    \includegraphics[width=0.49\linewidth]{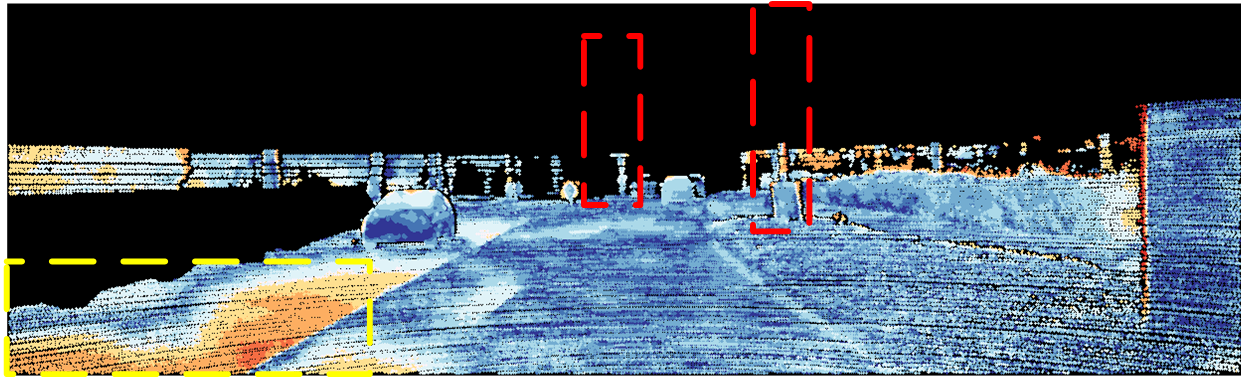}&
	\includegraphics[width=0.49\linewidth]{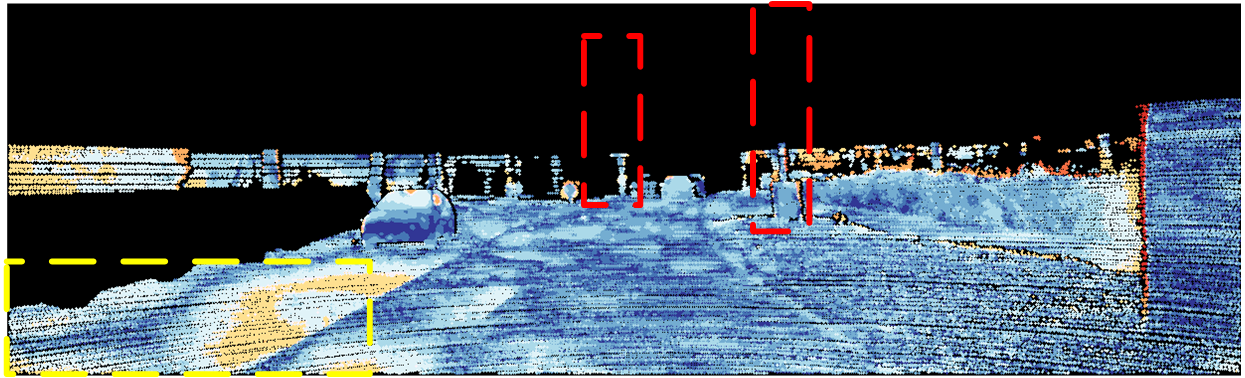} \\
	
	{\small (e) original error map} &  {\small (f) refined error map }		 \\
	
	\end{tabular}
	\vspace{-0.1in}
	\caption{\footnotesize Visualization comparison between the original disparity map and refined disparity map. All methods are only trained on the Scene Flow datatest and tested on unseen KITTI datasets.}
	\label{fig: residual disparity}
 	\vspace{-0.1in}
\end{figure}

Visualization results are shown in Fig. \ref{fig: residual disparity}. All methods are only trained on the Scene Flow datatest and tested on unseen KITTI datasets. As shown, the original disparity estimation results lack texture information, e.g., the missing thin structures (see red dash boxes in the picture) and unsmooth regions due to the segmentation of lane lines (see yellow dash boxes in the picture). Instead, the proposed attention-based disparity refinement module can learn such missing details by the residual disparity (sub-figure (b)) and generate a better refined disparity map.

\subsubsection{Loss function}
Inspired by previous work \cite{psmnet}, we employ smooth $L_1$ loss function \cite{smoothl1} to train the proposed stereo matching network. Specifically, the loss function is described as:
\begin{eqnarray}
L\left( {\hat D,{D_{gt}}} \right) = \frac{1}{{{N_{gt}}}}\sum\limits_{i = 1}^{N_{gt}} {smoot{h_{{L_1}}}({D_{gt}} - \hat D)}
\end{eqnarray}
in which
\begin{eqnarray}
Smoot{h_{{L_1}}}(x) = \left\{ \begin{array}{l}
0.5{x^2},{\rm{    if }}\left| x \right| < 1\\
\left| x \right| - 0.5,{\rm{ otherwise}}
\end{array} \right.
\end{eqnarray}
where $N_{gt}$ denotes the number of available pixels in the provided ground truth disparity of source domain and $\hat D$ represents the predicted disparity.

\subsection{Uncertainty estimation for pseudo-label generation}
\label{sec:uncertainty}
We propose an uncertainty based pseudo-label generation method to generate low-noise disparity maps and leverage them as supervision to adapt the pre-trained model to the target domain. A key observation behind our method is that deep stereo matching methods can be successfully adapted to a new domain by only deploying sparse ground-truth labels or even sparse noisy predictions \cite{unsuperviseddomainadaptation, self-supervisedeccv}. Based on the above observations, we propose to employ the target domain image pairs $({I_L},{I_R})$ and the pre-trained model ${M_{{\rm{pre}}}}$ to generate dense disparity maps ${D_{{\rm{pre}}}}$. Then we can leverage the uncertainty estimation to filter out unreliable points of ${D_{{\rm{pre}}}}$ and generate reliable and sparse disparity maps ${D_u}$. Specifically, two terms of uncertainty estimation, i.e., pixel-level and area-level are employed to generate low-noise disparity maps. Below we will introduce each term of uncertainty estimation for more details.

\textbf{Pixel-level Uncertainty Estimation:}  
As mentioned in Sec. \ref{cascade_cost}, we propose a pixel-level uncertainty estimation to adaptively adjust the next stage disparity searching range. As mentioned before, the proposed pixel-level uncertainty estimation is according to the sharpness of cost volume distribution to evaluate the confidence of current estimations. That is the proposed pixel-level uncertainty estimation has no learnable parameters and is totally decided by the input cost volume distribution. As the geometry of cost volume distribution is domain-invariant, we can ensure the generalization ability of pixel-level uncertainty estimation. Hence, intuitively, we can directly introduce the pixel-level uncertainty estimation to generate the corresponding pseudo-label ${D_{pixel}}$ as follows:
\begin{eqnarray}
\begin{array}{c}
{D_{{\rm{pre}}}} = {M_{{\rm{pre}}}}({I_L},{I_R}),\\
\\
{D_{pixel}} = \{ d \in {D_{pre}}:\sqrt U  < t\}, 
\end{array}
\end{eqnarray}
where $t$ is the threshold that controls the density and reliability of the filtered disparity map ${D_{pixel}}$. A lower value of t will filter out more mistakes in ${D_{{\rm{pre}}}}$ and generating more sparse disparity maps ${D_{pixel}}$. Thus, by setting a reasonable threshold, we can utilize filtered disparity maps ${D_u}$  as if they were ground truth to supervise the fine-tuning of the pre-trained model ${M_{{\rm{pre}}}}$. As shown in Fig. \ref{fig: uncertainty estimation}, the proposed pixel-level uncertainty estimation can filet out most error of the cross-domain disparity estimation ${D_{{\rm{pre}}}}$ and generate low-noise disparity maps ${D_{pixel}}$. However, the proposed pixel-level uncertainty estimation still has the following two shortcomings: 1) it only evaluates pixel-level confidence and doesn’t consider the influence of neighboring pixels or global information. 2) it only employs the cost volume as input curs while ignoring the usage of multi-input, i.e., reference image and disparity maps. Thus, area-level uncertainty estimation is essential for the further refinement of pixel-level uncertainty maps. 

\begin{figure}[!t]
 \centering
 \includegraphics[width=0.98 \linewidth]{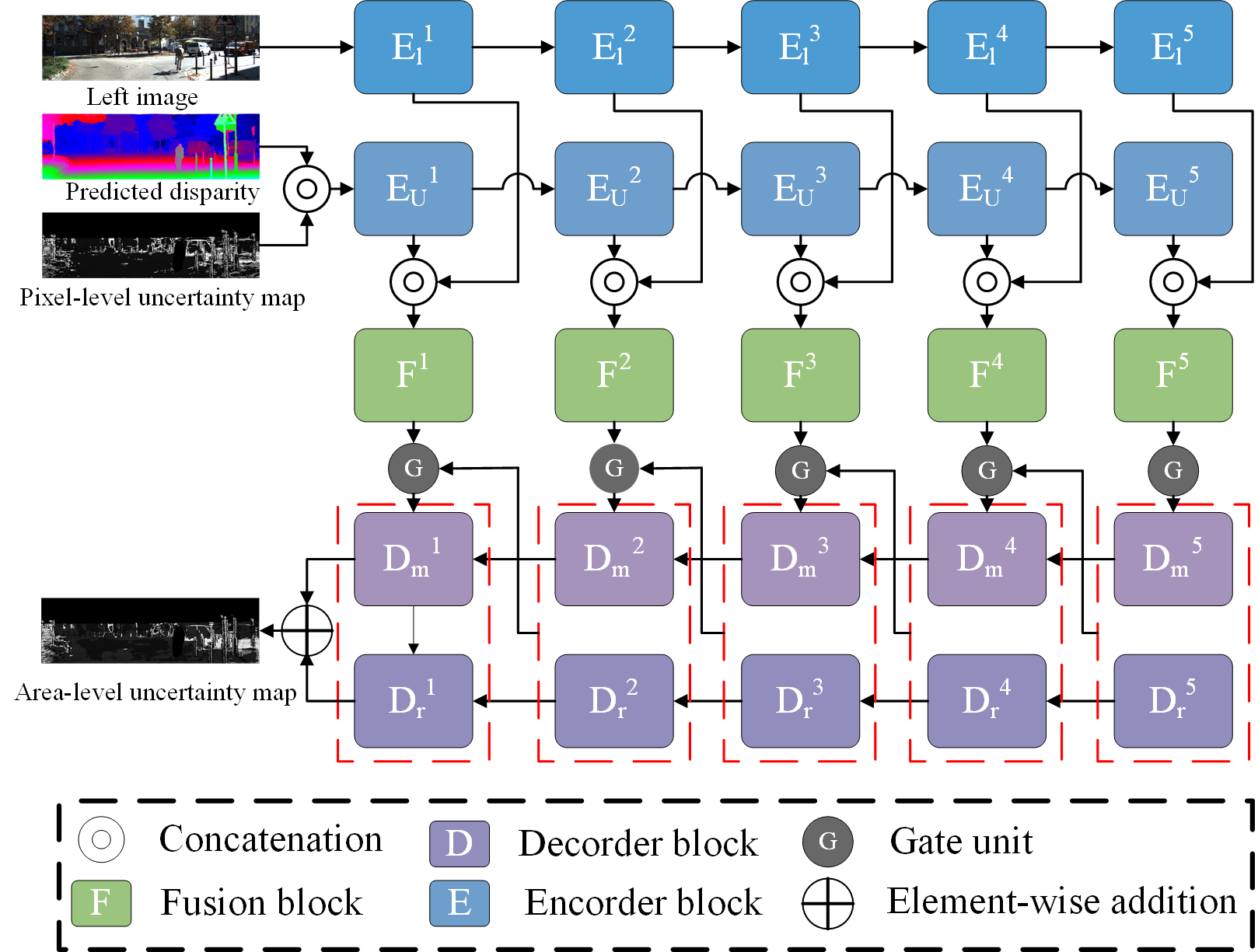} 
 \vspace{-0.1in}
 \caption{\footnotesize The architecture of area-level uncertainty estimation network. Left image, predicted disparity, and pixel-level uncertainty map are employed as input.}
 \label{fig: area-level uncertainty}
 	\vspace{-0.15in}
\end{figure}

\textbf{Area-level Uncertainty Estimation:} We propose an area-level uncertainty estimation to leverage the information of multi-modal input and neighboring pixels. Intuitively, both neighborhood and multi-modal information can better guide the pixel-level uncertainty map to identify the prediction correct region. For example, assuming that a pixel has high pixel-level uncertainty while the pixel-level uncertainty of surrounding similar pixels is relatively low, then we can judge it is the correct prediction to preserve more valid labels. In this case, the neighborhood information can guide the network to discover surrounding pixels and the multi-modal information can help the network to distinguish whether these pixels are similar and can be considered to have close uncertainty. Specifically, let us denote the input initial uncertainty map as ${U_{pixel}}$. Our goal is to recover from ${U_{pixel}}$ an improved uncertainty map ${U_{area}}$ that can more accurately identify the region where the disparity estimation is wrong. Such a task can be seen as a binary classification mission, where the output of the area-level uncertainty estimation will be constrained in the range of $(0,1)$. A higher value denotes a higher possibility of prediction error, i.e., higher uncertainty. Actually, such a task is very similar to the setting of salient object detection, which also employs a binary classification network to identify the visually distinctive regions or objects in a scene. Moreover, the usage of multi-modal input and neighboring information is an essential topic in salient object detection and has drawn great attention from the community. Hence, we propose to employ some network design ideas in salient object detection \cite{salient1,salient2,salient3, gatanet} to construct our area-level uncertainty estimation network. The overall architecture of our area-level uncertainty estimation network is shown in Fig. \ref{fig: area-level uncertainty}, which consists of six parts: left image encoder blocks $E_l^i$, uncertainty map encoder blocks $E_U^i$, fusion blocks $F^i$, gate unit $G^i$, main decoder block $D_m^i$ and residual decoder block $D_r^i$  ($i$ denotes different scales and $i = 0$ represents the original input image resolution). Below we will introduce each part in more detail.

\textbf{Encoder blocks:} Our encoder blocks can be divided into two parts: the reference image encoder block $E_l^i$ and uncertainty map encoder block $E_U^i$, which characterizes the information of the reference image and the concatenation of pixel-level uncertainty map and predicted disparity map, respectively. Specifically, we propose to employ the commonly used pre-trained backbone network ResNet-34 to construct our encoder block. Similar to previous work, we remove the last fully-connected and pooling layers of the employed backbone network.

\textbf{Fusion blocks:} Our fusion block has two main inputs: 1) left image encoder blocks, which represent the multi-scale information of reference images. 2) uncertainty map encoder block, which stores the multi-scale information of pixel-level uncertainty map and predicted disparity map. By employing the fusion blocks, we can integrate scale-matching reference image encoder blocks and uncertainty map encoder blocks to extract robust multi-modal input representation. Specifically, the fusion process can be formulated as:
\begin{equation}
{F^i} = \delta({ E_l^i }||{ E_U^i }),
    \label{eq:decoder block}
\end{equation}
where $||$ denotes the concatenation operation and $\delta$ refers to the convolution layer.

\textbf{Gate unit based decoder block: } Inspired by previous work \cite{salient3,gated2}, we propose the gate unit based decoder block to control the message passing between scale-matching fusion and decoder blocks. In specific, our gate unit is formulated as:
\begin{scriptsize}
\begin{equation}
{G^i} = [G_m^i,G_r^i] = \left\{ \begin{array}{l}
AvgPool(S(\delta ({F^i}\parallel {D_m^{i + 1}}))){\rm{               if }}\;i = 1,2,3,4\\
AvgPool(S(\delta ({F^i}||FASPP({F^i})))){\rm{   if }}\;i  =  5
\end{array} \right.
\end{equation}
\end{scriptsize}
where $AvgPool$ denotes the global average pooling, $S$ is the sigmoid function, $||$ represents the concatenation operation, $\delta$ refers to the convolution layer and $FASPP$ is the Fold-ASPP operation \cite{gatanet}. Note that the output channel of $\delta$ is 2. Hence, the proposed gate unit has two outputs, i.e., $G_m^i$ and $G_r^i$ , which will be employed to control the message passing in main decoder block $D_m^i$ and residual decoder block $D_r^i$, respectively. Specifically,  $G_m^i$  is employed to balance the contribution between upsampled main decoder blocks $D_m^i$ and corresponding fusion blocks $F^i$. The whole process can be written as:
\begin{equation}
D_m^i = \left\{ \begin{array}{l}
\delta (G_m^i \times \delta ({F^i}) + Up(D_m^{i + 1})){\rm{  if }}\;i = 1,2,3,4\\
\delta (G_m^i \times FASPP({F^i})){\rm{          if}}\; i  =  5
\end{array} \right.
\end{equation}
where $\times$ is the element-wise multiplication operation and $Up$ refers to the up-sampling operation which is implemented by bilinear interpolation. Then we further introduce the residual decoder block $D_r^i$ to recover the missed details of the main decoder blocks and also employ the gate unit $G_r^i$ to balance the information flow. The whole process can be formulated as:
\begin{equation}
D_r^i = \left\{ \begin{array}{l}
G_r^i \times \delta ({F^i})||Up(D_r^{i - 1}{\rm{)  if }}\;i = 1,2,3,4\\
G_r^i \times \delta ({F^i}){\rm{           if}}\; i  =  5
\end{array} \right.
\end{equation}
Finally, we can fuse the output of two terms of decoder blocks to generate the area-level uncertainty map ${U_{area}}$:
\begin{equation}
{U_{area}} = S(\delta (D_r^1||D_m^1) + D_m^1)
\end{equation}
where $S$ is the sigmoid function. Thus, the output of the area-level uncertainty estimation ${U_{area}}$ is in the range of $(0,1)$ and a higher value denotes a higher possibility of prediction error. Then, we can use the same operation introduced in pixel-level uncertainty estimation to generate the corresponding pseudo-label ${D_{area}}$ as follows:
\begin{eqnarray}
\begin{array}{c}
{D_{area}} = \{ d \in {D_{pre}}:{U_{area}}  < t\} 
\end{array}
\end{eqnarray}
where $t$ is the threshold that controls the density and reliability of the filtered disparity map ${D_{area}}$.

\subsubsection{Loss function}
We employ the cross-entropy loss to train the proposed area-level uncertainty estimation network. Specifically, the cross-entropy loss can be defined as:
\begin{eqnarray}
l = U_{gt}\log {U_{area}} + (1 - U_{gt})\log (1 - {U_{area}})
\end{eqnarray}
where $U_{area}$ and $U_{gt}$ denote the predicted area-level uncertainty map and ground truth uncertainty mask, respectively. As the ground truth uncertainty mask is not provided in the source domain(synthetic dataset) and indeed the value of it will change as the convergence of the stereo matching network. Here, we specify how to obtain $U_{gt}$ on source domain according to the provided ground truth disparity $D_{gt}$ and predicted disparity $\hat D$. Specifically, our ground truth uncertainty mask is defined as:
\begin{eqnarray}
{U_{gt}} = \left\{ \begin{array}{l}
1,{\rm{  if }}\left| {{D_{gt}} - \hat D} \right|{\rm{ > }}\delta {\rm{ }}\\
0,{\rm{  otherwise}}
\end{array} \right.
\end{eqnarray}
where $\delta$ is the threshold that controls the strictness of uncertainty estimation, e.g., a higher $\delta$ denotes a larger gap between $D_{gt}$ and $\hat D$ can be seen as a correct prediction.

\subsection{Domain Adaptation with supervision of pseudo-label  }
\label{sec:domain_adaptation}

After getting the generated pseudo-label ${D_{area}}$, we can employ it to adapt the pre-trained binocular depth estimation network to the new domain. The loss function is defined as: 
\begin{eqnarray}
L\left( {D,{D_{area}}} \right) = \frac{1}{{{N_{area}}}}\sum\limits_{i = 1}^{N_{area}} {smoot{h_{{L_1}}}({D_{area}} - \hat D)}
\end{eqnarray}
in which
\begin{eqnarray}
Smoot{h_{{L_1}}}(x) = \left\{ \begin{array}{l}
0.5{x^2},{\rm{    if }}\left| x \right| < 1\\
\left| x \right| - 0.5,{\rm{ otherwise}}
\end{array} \right.
\end{eqnarray}
where $N_{area}$ denotes the number of available pixels in the filtered disparity map ${D_{area}}$ and $\hat D$ represents the predicted disparity. Besides, we can also employ the generated pseudo-label as supervision to train the monocular depth estimation network in an unsupervised way. The loss function is defined as:
\begin{eqnarray}
L({\hat D_{mono}},{D_{area}}) = \sqrt {\frac{1}{{{N_{area}}}}\sum\limits_{i = 1}^{{N_{area}}} {d_i^2 - \frac{\lambda }{{{n^2}}}} {{(\sum\limits_{i = 1}^{{N_{area}}} {{d_i}} )}^2}}
\end{eqnarray}
in which
\begin{eqnarray}
{d_i} = \log {{\hat D}_{mono}} - \log {D_{area}}
\end{eqnarray}
where $N_{area}$ denotes the number of available pixels in the filtered disparity map ${D_{area}}$ and $\hat D_{mono}$ represents the predicted disparity map by monocular depth estimation networks. The balancing factor $\lambda$ is set to 0.85. Then the disparity can be converted to depth by triangulation:
\begin{eqnarray}
Depth = \frac{{fB}}{d}
\end{eqnarray}
where f denotes the camera's focal length and B is the baseline, i.e., the distance between two camera centers.

\section{Stereo matching Experiments}
\label{sec:stereo_experinment}
\subsection{Dataset}

\noindent \textbf{SceneFlow:} This is a large synthetic dataset including 35,454 training and 4,370 test images with a resolution of $960\times 540$ for optical flow and stereo matching. We use it to pre-train our network. 

\noindent \textbf{Middlebury:} Middlebury \cite{mid} is an indoor dataset with 28 training image pairs (13 of them are additional training images) and 15 testing image pairs with full, half, and quarter resolutions. It has the highest resolution among the three datasets and the disparity range of half-resolution image pairs is 0-400. 

\noindent \textbf{KITTI 2012\&2015:} They are both real-world datasets collected from a driving car. KITTI 2015 \cite{kitti2} contains 200 training and another 200 testing image pairs while KITTI 2012 \cite{kitti1} contains 194 training and another 195 testing image pairs. Both training image pairs provide sparse ground-truth disparity and the disparity range of them is 0-230.

\noindent \textbf{ETH3D:} ETH3D \cite{eth3d} is the only grayscale image dataset with both indoor and outdoor scenes. It contains 27 training and 20 testing image pairs with sparsely labeled ground truth. It has the smallest disparity range among the three datasets, which is just in the range of 0-64.

\begin{table*}[!htb]
\caption{\footnotesize Joint Generalization comparison on ETH3D, Middlebury, and KITTI2015 datasets. 	\textbf{Top:} Joint Generalization comparison with methods who participated in the Robust Vision Challenge 2020, our method achieves the best overall performance. \textbf{Bottom:} Joint Generalization comparison with the top 3 methods in the past three years, our method surpasses previous work on all three datasets by a noteworthy margin. All methods are tested on three datasets without adaptation. We highlight the best result in \textbf{bold} and the second-best result in \B{blue} for each column. The overall rank is obtained by Schulze Proportional Ranking \cite{rank} to join multiple rankings into one.
}
\vspace{-0.1in}
\centering
\resizebox{\textwidth}{!}{
\begin{tabular}{c|c|c|c|c|c|c|c|c|c|c|c|c|c}
\hline
\multirow{2}{*}{Method}   & \multicolumn{4}{c|}{KITTI}                                 & \multicolumn{4}{c|}{Middlebury}                            & \multicolumn{4}{c|}{ETH3D}                                 & \multirow{2}{*}{\begin{tabular}[c]{@{}c@{}}Overall\\ Rank\end{tabular}} \\ \cline{2-13}
                          & D1\_bg        & D1\_fg        & D1\_all       & Rank & bad 1.0       & bad 2.0       & avg error     & Rank & bad 1.0       & bad 2.0       & avg error     & Rank &                                  \\ \hline
NLCANet\_V2\_RVC \cite{nlcanet}         & \textbf{1.51} & \textbf{\B{3.97}}          & \textbf{1.92} & \textbf{1} & \textbf{\B{29.4}}          & \textbf{\B{16.4}}          & 5.60          & 3          & \textbf{\B{4.11}}          & \textbf{\B{1.2}}           & \textbf{\B{0.29}}          & \textbf{\B{2}}          & \textbf{\B{2}}                                \\ 
HSMNet\_RVC \cite{hsm}               & 2.74          & 8.73          & 3.74          & 6          & 31.2          & 16.5          & \textbf{3.44} & \textbf{1} & 4.40          & 1.51          & 0.28          & 3          & 3                                \\ 
CVANet\_RVC               & 1.74          & 4.98          & 2.28          & 3          & 58.5          & 38.5          & 8.64          & 5          & 4.68          & 1.37          & 0.34          & 4          & 4                                \\ 
AANet\_RVC \cite{aanet}                & 2.23          & 4.89          & 2.67          & 5          & 42.9          & 31.8          & 12.8          & 6          & 5.41          & 1.95          & 0.33          & 5          & 5                                \\ 
GANet\_RVC \cite{ganet}                & 1.88          & 4.58          & 2.33          & 4          & 43.1          & 24.9          & 15.8          & 7          & 6.97          & 1.25          & 0.45          & 6          & 6                                \\ 
\textbf{CFNet\_RVC(ours)} & \textbf{\B{1.65}}          & \textbf{3.53} & \textbf{\B{1.96}}          & \textbf{\B{2}}          & \textbf{26.2} & \textbf{16.1} & \textbf{\B{5.07}}          & \textbf{\B{2}}          & \textbf{3.7}  & \textbf{0.97} & \textbf{0.26} & \textbf{1} & \textbf{1}                       \\ \hline
iResNet\_ROB \cite{iresnet, mcvmfc}                   & 2.27          & 4.89          & 2.71          & 4          & 45.9          & 31.7          & 6.56          & 3          & 4.67          & 1.22          & 0.27          & 4          & 4                                \\ 
Deeppruner\_ROB \cite{deeppruner}                & -             & -             & 2.23          & 3          & 57.1          & 36.4          & 6.56          & 4          & 3.82          & 1.04          & 0.28          & 3          & 3                                \\ 
\textbf{CFNet\_RVC(ours)} & \textbf{\B{1.65}} & \textbf{\B{3.53}} & \textbf{\B{1.96}} & \textbf{\B{2}}          & \textbf{26.2} & \textbf{16.1} & \textbf{5.07} & \textbf{1}          &\textbf{\B{3.7}}           & \textbf{\B{0.97}}          & \textbf{\B{0.26}}          & \textbf{\B{2}}          & \textbf{\B{2}}                               \\ 
\textbf{UCFNet\_RVC(ours)} & \textbf{1.57} & \textbf{3.33} & \textbf{1.86} & \textbf{1}          & \textbf{\B{31.6}} & \textbf{\B{16.7}} & \textbf{\B{5.96}} & \textbf{\B{2}}          &\textbf{3.37}           & \textbf{0.78}          & \textbf{0.25}          & \textbf{1}          & \textbf{1}                               \\ \hline
\end{tabular}
}
\label{tab: robust challenge}
\vspace{-0.15in}
\end{table*}

\subsection{Implementation Details}

We use PyTorch to implement our network and employ Adam (${\beta _{\rm{1}}} = 0.9,{\beta _2} = 0.999$) to train the whole network in an end-to-end way. The batch size is set to 16 for training on 4 Tesla V100 GPUs and the whole disparity search range is fixed to 256 during the training and testing process. ${N^1}$ and ${N^2}$ are set as 12 and 16, respectively. Threshold $\delta$ of the ground truth uncertainty mask is 1. Asymmetric chromatic augmentation and asymmetric occlusion \cite{hsm} are employed for data augmentation. Below we will introduce our training process for each term of generalization in detail.  

\noindent \textbf{Cross-domain generalization training:} Our pre-training process on the synthetic dataset (source domain) can be broken down into three steps. Firstly, we use switch training strategy to pre-train our UCFNet in the SceneFlow dataset. Specifically, we first use ReLU to train our network from scratch for 20 epochs, then we switch the activation function to Mish and prolong the pre-training process in the SceneFlow dataset for another 15 epochs. Secondly, we fix the weights of the UCFNet (except the refinement module) and train the attention-based refinement network individually for 20 epochs with a 0.0001 learning rate. Thirdly, we fix the weights of the UCFNet and train the area-level uncertainty estimation network alone for 15 epochs. The initial learning rate is 0.001 and is down-scaled by 2 after epochs 10,12,14.

\noindent \textbf{Adapt generalization training:} After obtaining a strong pre-training model, we can further adapt our pre-trained model to the new domain with the generated pseudo-labels. Specifically, the training process of domain adaptation can be broken down into two steps. First, we feed the synchronized stereo images of the target domain into the pre-trained model and employ the proposed uncertainty-based pseudo-label generation method to generate corresponding pseudo-labels. Secondly, we employ the generated pseudo-labels as supervision to adapt the pre-trained model to the new domain. Specifically, we first fix the weights of the refinement network and train UCFNet on the target domain for 50 epochs with a 0.001 learning rate. Then, we fix the weights of the UCFNet (except the refinement module) and train the attention-based refinement network individually for 50 epochs with a 0.0001 learning rate.

\noindent \textbf{Joint generalization training:} We propose a three-stage finetuning strategy for joint generalization training. First, as mentioned in the cross-domain generalization training, we employ the switch training strategy to pre-train our model in the SceneFlow dataset. Second, we jointly finetune our pre-train model on four datasets, i.e., KITTI 2015, KITTI2012, ETH3D, and Middlebury for 400 epochs. The initial learning rate is 0.001 and is down-scaled by 10 after epoch 300. Third, we augment Middlebury and ETH3D to the same size as KITTI 2015 and finetune our model for 50 epochs with a learning rate of 0.0001. The core idea of our three-stage finetune strategy is to prevent the small datasets from being overwhelmed by large datasets. By augmenting small datasets at stage three and training our model with a small learning rate, our strategy makes a better trade-off between generalization capability and fitting capability on three datasets.

\begin{table}[!t]
\caption{\footnotesize \textbf{Top}: Cross-domain generalization evaluation on ETH3D, Middlebury, and KITTI training sets. All methods are only trained on the Scene Flow datatest and tested on full-resolution training images of three real datasets. \textbf{Bottom}: Adaptation generalization evaluation on ETH3D, Middlebury, and KITTI training sets. TDD: target domain data. All methods are finetuned on the unlabeled target domain data. We highlight the best result in \textbf{bold} and the second-best result in \B{blue} for each column.}
\vspace{-0.1in}
\centering
\resizebox{0.49\textwidth}{!}{
\begin{tabular}{c|c|c|c|c|c}
\hline
Method       & Training Set         & \begin{tabular}[c]{@{}c@{}}KITTI2012\\ D1\_all(\%)\end{tabular} & \begin{tabular}[c]{@{}c@{}}KITTI2015\\ D1\_all(\%)\end{tabular} & \begin{tabular}[c]{@{}c@{}}Middlebury\\ bad 2.0(\%)\end{tabular} & \begin{tabular}[c]{@{}c@{}}ETH3D\\ bad 1.0(\%)\end{tabular} \\ \hline
\multicolumn{6}{c}{Cross-domain Generalization Evaluation}                                                                                                                                                                                                                                                        \\ \hline
PSMNet \cite{psmnet}       & synthetic            & 15.1                                                            & 16.3                                                            & 39.5                                                             & 23.8                                                        \\ 
GWCNet \cite{gwcnet}      & synthetic            & 12.0                                                            & 12.2                                                            & 37.4                                                             & 11.0                                                        \\ 
CasStereo \cite{cascade}    & synthetic            & 11.8                                                            & 11.9                                                            & 40.6                                                             & 7.8                                                         \\ 
GANet \cite{ganet}       & synthetic            & 10.1                                                            & 11.7                                                            & 32.2                                                             & 14.1                                                        \\ 
DSMNet \cite{dsmnet}       & synthetic            & 6.2                                                             & 6.5                                                             & \B{\textbf{21.8}}                                                    & 6.2                                                         \\
ITSA-CFNet \cite{itsa}       & synthetic            & \textbf{4.2}                                                             & \textbf{4.7}                                                             & \textbf{20.7}                                                    & \B{\textbf{5.1}}                                                         \\ 
\textbf{CFNet(ours)\cite{cfnet}}       & synthetic            & 4.7                                                    & 5.8                                                    & 28.2                                                             & 5.8                                                \\ 
\textbf{UCFNet\_pretrain}       & synthetic            & \B{\textbf{4.5}}                                                    & \B{\textbf{5.2}}                                                    & 26.0                                                             & \textbf{\textbf{4.8}}                                                \\ \hline
\multicolumn{6}{c}{Adaptation Generalization Evaluation}                                                                                                                                                                                                                                                          \\ \hline
AOHNet \cite{aohnet}       & synthetic+TDD(no gt) & 8.6                                                             & 7.8                                                             & -                                                                & -                                                           \\ 
ZOLE \cite{zoom}     & synthetic+TDD(no gt) & -                                                               & 6.8                                                             & -                                                                & -                                                           \\ 
MADNet \cite{madnet}      & synthetic+TDD(no gt) & 9.3                                                             & 8.5                                                             & -                                                                & -                                                           \\
AdaStereo \cite{adastereo,adastereo_ijcv}      & synthetic+TDD(no gt) & \B{\textbf{3.6}}                                                             & \B{\textbf{3.5}}                                                             & \textbf{\textbf{18.5}}                                                                & \B{\textbf{4.1}}                                                           \\ 
\textbf{UCFNet\_adapt} & synthetic+TDD(no gt) & \textbf{2.8}                                                    & \textbf{3.1}                                                    & \B{\textbf{20.8}}                                                             & \textbf{3.0} \\\hline
\end{tabular}
}
\label{tab: cross-domain generalization}
	\vspace{-0.1in}
\end{table}

\subsection{Robustness Evaluation}
\label{Sec. Robustness Evaluation}

In this section, we evaluate our method on three terms of generalization and compare it with state-of-the-art methods in each category.

\noindent \textbf{Cross-domain Generalization:} As the target domain data cannot be easily obtained in many real scenarios, the network’s ability to perform well on unseen scenes is indispensable for robust stereo matching. Towards this end, we evaluate methods' cross-domain generalization by training on synthetic images and testing on real images. As shown in Tab. \ref{tab: cross-domain generalization}, our method far outperforms domain-specific methods \cite{psmnet,gwcnet,cascade,ganet} and our conference version CFNet on all four datasets with a large margin. Specifically, the error rate on KITTI 2012, KITTI 2015, Middlebury, and ETH3D has been decreased by 4.26\%, 10.34\%, 7.80\%, and 17.24\%, respectively compared to CFNet. Moreover, ITAS-CFNet \cite{itsa}, a specially designed stereo matching method developed from our conference version for cross-domain generalization, is the current best-published method. Our method can achieve comparable performance with it on most datasets, which further verifies our uncertainty-based cascade and fused cost volume representation is an efficient approach for robust stereo matching. 

\noindent \textbf{Adaptation Generalization:}  
Comparing with collecting accurate ground-truth disparities, unlabeled target data is much easier to obtain. Thus, how to employ the knowledge of unlabeled target data adapting pre-trained models to the new domain is also essential. We evaluate such adapt generalization by training on synthetic images and finetuning on unlabeled real images. As shown in Tab. \ref{tab: cross-domain generalization}, although the generalization of our pre-trained model has outperformed most domain generalization methods, using the knowledge of unlabeled target data can still achieve a tremendous gain and surpass all domain generalization methods. Specifically, compared to our pre-trained model UCFNet\_pretrain, UCFNet\_adapt achieves 37.78\%,  40.38\%,  20\%,  37.5\% error reduction on KITTI2012, KITTI2015, Middlebury, and ETH3D, respectively. Moreover, compared to the current best-published domain adaptation method AdaStereo \cite{adastereo, adastereo_ijcv}, our method can still outperform it on three of four datasets, which further proves the effectiveness of the proposed method. Note that our method doesn’t employ the non-adversarial progressive color transfer and cost normalization proposed in AdaStereo, thus, the performance of our method has the potential for further improvement. 

\begin{figure}[!t]
 \centering
 \tabcolsep=0.05cm
 \begin{tabular}{c c c}
    \includegraphics[width=0.33\linewidth]{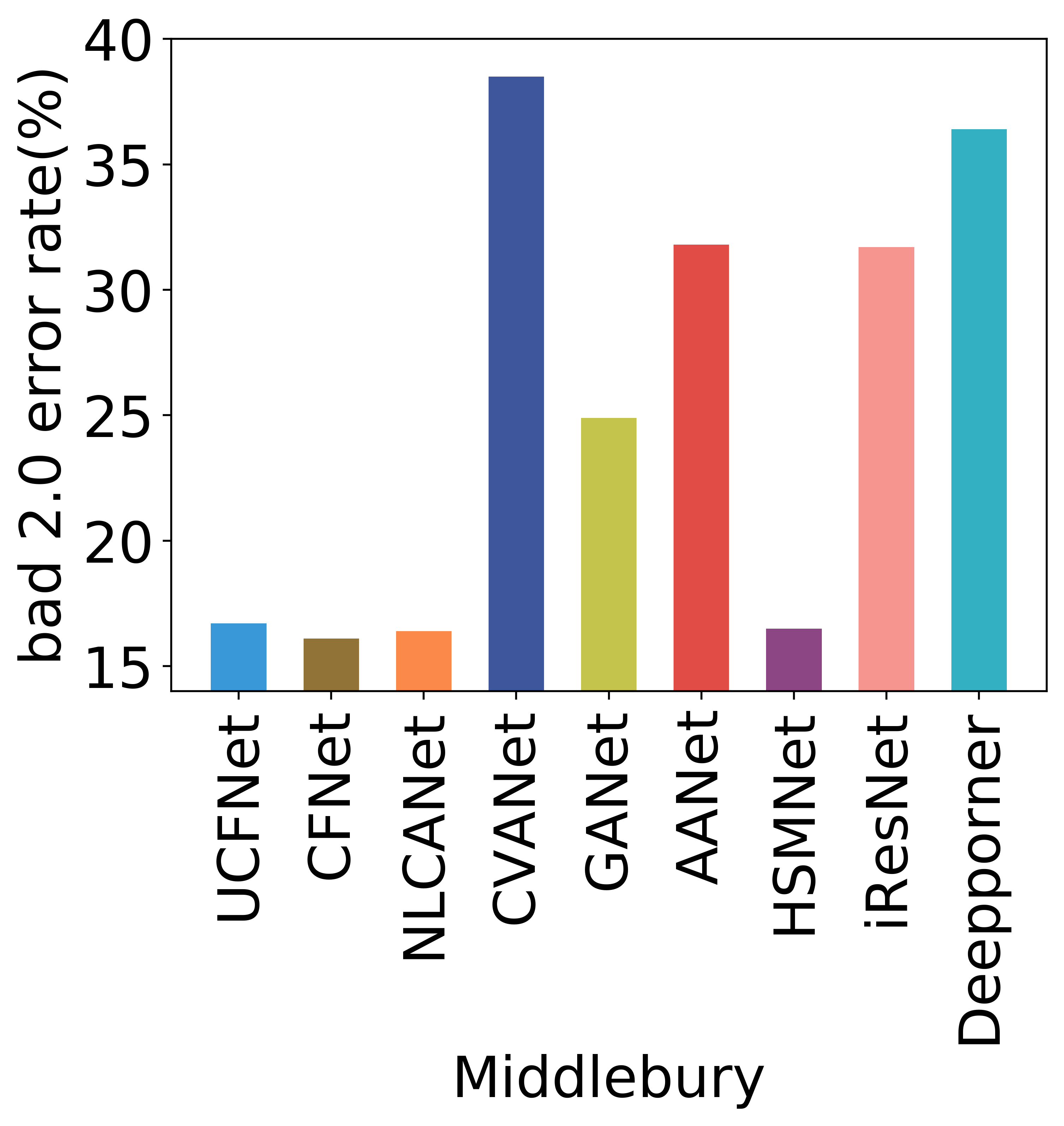}&
	\includegraphics[width=0.33\linewidth]{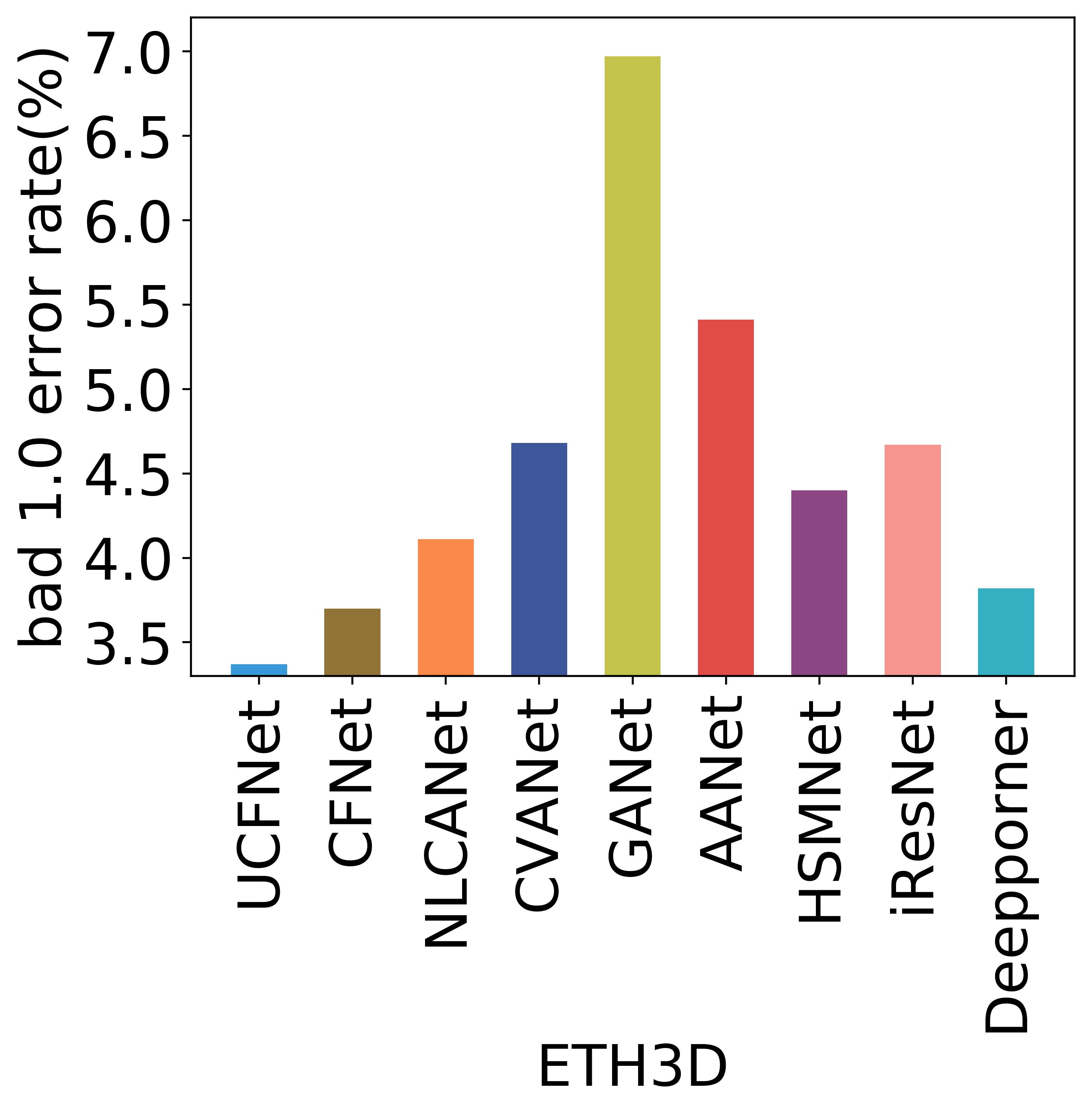}&
	\includegraphics[width=0.33\linewidth]{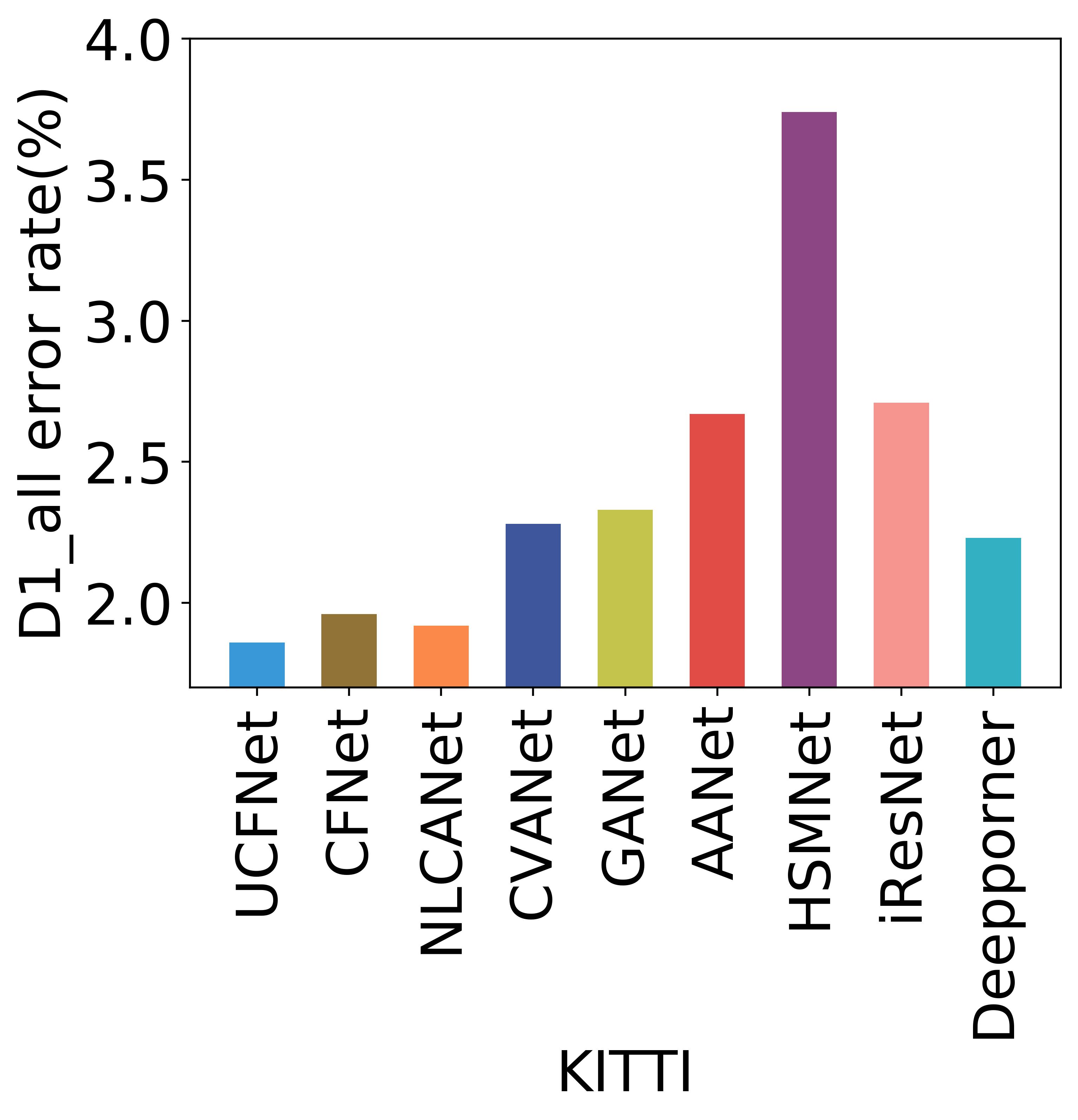}\\
	\end{tabular}
 \vspace{-0.2in}
 \caption{\footnotesize Performance comparison in terms of joint generalization on Middlebury, ETH3D, and KITTI 2015 datasets. Bad 2.0, bad1.0, and D1\_all (the lower the better) are used for evaluation, respectively. All methods are trained on the same training images and tested on three datasets with single model parameters and hyper-parameters. The proposed UCFNet\&CFNet achieves state-of-the-art generalization and performs well on all three real-world datasets.}
 \label{fig: robust comparison}
 	\vspace{-0.15in}
\end{figure}

\noindent \textbf{Joint Generalization:} Learning-based methods are usually limited to specific domains and cannot get comparable results on other datasets. Thus, the network’s ability to perform well on a variety of datasets with the same model parameters is essential for current methods. This is also the goal of Robust Vision Challenge 2020. Towards this end, we evaluate methods' joint generalization by their performance on three real datasets (KITTI, ETH3D, and Middlebury) without finetuning. We list the result of Robust Vision Challenge 2020 in the upper section of Tab. \ref{tab: robust challenge}. It can be seen from this table that HSMNet\_RVC \cite{hsm} ranks first on the Middlebury dataset. But it can’t get comparable results on the other two datasets (3rd on ETH3D 2017 and 6th on KITTI 2015). In particular, its performance on KITTI 2015 dataset is far worse than the other five. This is because this method is specially designed for high-resolution datasets and can’t generalize well to other datasets. The similar situation also appeared on other methods (GANet\_RVC, CVANet\_RVC, and AANet\_RVC). In contrast, our conference version CFNet\_RVC shows great generalization ability and performs well on all three datasets (2nd on KITTI 2015, 1st on ETH3D 2017, and 2nd on Middlebury 2014) and achieves the best overall performance. Additionally, we further compare the proposed UCFNet\_RVC with the top three methods in the previous Robust Vision Challenge in the lower part of the tab. \ref{tab: robust challenge}. As shown, our approach outperforms Deeppruner\_ROB and iResNet\_ROB on all three datasets with a remarkable margin. Compared with our conference version CFNet\_RVC, the proposed UCFNet\_RVC can achieve similar performance on Middlebury and surpass it on the other two datasets, which further verifies the effectiveness of the proposed refinement module. See corresponding visualization results in Fig. \ref{fig: robust comparison}.

\subsection{Results on KITTI Benchmark}
Although our focus is not on domain-specific performance, we still fine-tune our model on KITTI 2012 and KITTI 2015 benchmarks to show the efficiency of our method. Note that the
training strategy is same with our conference paper and the only difference is the proposed attention-based disparity refinement network (please see our conference paper for more details). Specifically, some state-of-the-art real-time methods and best-performing approaches are listed in Tab. \ref{tab: KITTI benchmark}. We find that our method achieves a 1.49\% three-pixel error rate on KITTI2012, a 6\% error reduction from our conference version\cite{cfnet} with a similar running time. Moreover, Lac-GaNet is the best published method on KITTI2012 and the proposed method can achieve comparable performance with 9 times faster speed, i.e., 1.42\% (1.8s) vs 1.49\% (0.21s), which implies the effectiveness and efficiency of the proposed method. A similar situation can also be observed in the KITTI2015 benchmark. 

\begin{table}[!htb]
\caption{\footnotesize Results on KITTI benchmark. \textbf{Top:} Comparison with best-performing methods. \textbf{Bottom:} Comparison with real-time methods. All methods are finetuned on specific datasets.}
\vspace{-0.1in}
\centering
\resizebox{0.4\textwidth}{!}{
\begin{tabular}{c|c|c|c|c|c}
\hline
\multirow{2}{*}{Method} & \multicolumn{2}{c|}{\begin{tabular}[c]{@{}c@{}}KITTI2012\\ 3px(\%)\end{tabular}} & \multicolumn{2}{c|}{\begin{tabular}[c]{@{}c@{}}KITTI2015\\ D1\_all(\%)\end{tabular}} & \multirow{2}{*}{\begin{tabular}[c]{@{}c@{}}time\\  (s)\end{tabular}} \\ \cline{2-5}
                        & Noc                                     & All                                    & Noc                                       & All                                      &                                                                      \\ \hline
LaCGANet \cite{lacnet} & \textbf{1.05}  & \textbf{1.42}    & \textbf{1.49}  & \textbf{1.67}  & 1.8 \\
LEAStereo   \cite{lea}            & 1.13                           & 1.45                          & 1.51                            & 1.65                            & 0.3                                                                  \\
CREStereo \cite{cre}    & 1.14  & 1.46  & 1.54 & 1.69 & 0.41 \\
GANet-deep \cite{ganet}             & 1.19                                    & 1.60                                   & 1.63                                      & 1.81                                     & 1.8                                                                  \\ 
AcfNet \cite{acfnet}                 & 1.17                                    & 1.54                                   & 1.72                                      & 1.89                                     & 0.48                                                                 \\ 
Casstereo \cite{cascade}                & -                                       & -                                      & 1.78                                      & 2.0                                      & 0.6                                                                  \\ \hline
HITNet \cite{hitnet}                  & 1.41                                    & 1.89                                   & 1.74                                      & 1.98                                     & \textbf{0.015}                                                       \\ 
HD\textasciicircum{}3 \cite{hd3}   & 1.40                                    & 1.80                                   & 1.87                                      & 2.02                                     & 0.14                                                                 \\ 
AANet+ \cite{aanet}                  & 1.55                                    & 2.04                                   & 1.85                                      & 2.03                                     & 0.06                                                                 \\ 
HSMNet \cite{hsm}                  & 1.53                                    & 1.99                                   & 1.92                                      & 2.14                                     & 0.14                                                                 \\ 
Deeppruner \cite{deeppruner}              & -                                       & -                                      & 1.95                                      & 2.15                                     & 0.18                                                                 \\ 
\textbf{CFNet(conference version)}    & 1.23                           & 1.58                          &  1.73                                        & 1.88                            & 0.18  \\ \textbf{UCFNet(ours)}    & \textbf{1.12}                           & \textbf{1.49}                          &  \textbf{1.61}                                        & \textbf{1.77}                            & 0.21                                                                 \\ \hline                        
\end{tabular}
}
\label{tab: KITTI benchmark}
	\vspace{0.1in}

\caption{\footnotesize Ablation study of attention-based disparity refinement module.  All methods are trained on the Sceneflow dataset with ground truth and tested on the KITTI2015 dataset. $U_{pixel}$ denotes the pixel-level uncertainty estimation result. The approach which is used in our final model is underlined.}
\vspace{-0.1in}
\centering
\resizebox{0.4\textwidth}{!}{
\begin{tabular}{c|c}
\hline
Method                                            & \textbf{D1\_all} \\ \hline
No refinement                                     & 5.8\%            \\ 
{ \ul Refinement with disparity input}                   & \textbf{5.2\%}            \\ 
Refinement with disparity + left image input      & 5.4\%            \\ 
Refinement with disparity + $U_{pixel}$ input  & 5.4\%            \\ 
refinement with disparity + left feature input    & \textbf{5.2\%}            \\ \hline
\end{tabular}
}
\label{tab: abstudy_refinement}
\vspace{0.1in}

\caption{ \footnotesize Ablation study results of the proposed network on KITTI 2015 training set. All methods are trained on the Sceneflow dataset with ground truth and fintuned on the unlabeled target domain data. SF and K denotes Sceneflow and kitti dataset, respectively. $D_{pixel}$ and $D_{area}$ are the pixel-level proxy label and area-level proxy label. The iteration number is set to one.}
\vspace{-0.1in}
\centering
{
\begin{tabular}{c|c|c|c}
\hline
Backbone      & Supervision & Proxy label & D1\_all \\ \hline
CFNet         & SF                   & -           & 5.8     \\ 
CFNet         & SF+K(no gt)                 & $D_{pixel}$          & 4.5     \\ 
CFNet         & SF+K(no gt)                 & $D_{area}$   & \textbf{3.88}    \\ \hline
UCFNet & SF                   & -           & 5.2     \\ 
UCFNet & SF+K(no gt)                  & $D_{pixel}$          & 3.89    \\ 
UCFNet & SF+K(no gt)                  & $D_{area}$   & \textbf{3.64}    \\ \hline
\end{tabular}
}
\label{tab: abstudy_pixel_area}
\vspace{-0.1in}
\end{table}

\begin{table*}[!t]
\caption{ \footnotesize Ablation study of different proxy label generation methods. We will filter out all the pixels whose uncertainty is larger than the threshold. All methods are trained on the Sceneflow dataset with ground truth and fintuned on the unlabeled KITTI2015 dataset. $U_{pixel}$ and ${U_{area}}$ are pixel-level uncertainty estimation and area-level uncertainty estimation, respectively. Overlap denotes the overlap percentage between the generated proxy label and the ground truth. The iteration number is set to one.}
\vspace{-0.1in}
\centering
{
\begin{tabular}{ccc|ccccc|cl}
\cline{1-9}
\multicolumn{3}{c|}{Pretraining model}                                                                          & \multicolumn{5}{c|}{Proxy label generation}                                                                                                                      & Domain adaptation &  \\ \cline{1-9}
\multicolumn{1}{c|}{Backbone}                & \multicolumn{1}{c|}{D1\_all(\%)}          & density(\%)          & \multicolumn{1}{c|}{Filter}                 & \multicolumn{1}{c|}{threshold} & \multicolumn{1}{c|}{D1\_all(\%)} & \multicolumn{1}{c|}{density(\%)} & overlap(\%) & D1\_all(\%)       &  \\ \cline{1-9}
\multicolumn{1}{c|}{\multirow{9}{*}{UCFNet}} & \multicolumn{1}{c|}{\multirow{9}{*}{5.2}} & \multirow{9}{*}{100} & \multicolumn{1}{c|}{\multirow{4}{*}{$U_{pixel}$}} & \multicolumn{1}{c|}{0.7}       & \multicolumn{1}{c|}{0.74}        & \multicolumn{1}{c|}{47.62}       & 47.22       & 4.01              &  \\ 
\multicolumn{1}{c|}{}                        & \multicolumn{1}{c|}{}                     &                      & \multicolumn{1}{c|}{}                       & \multicolumn{1}{c|}{0.8}       & \multicolumn{1}{c|}{1.8}         & \multicolumn{1}{c|}{73.54}       & 79.17       & 3.96              &  \\ 
\multicolumn{1}{c|}{}                        & \multicolumn{1}{c|}{}                     &                      & \multicolumn{1}{c|}{}                       & \multicolumn{1}{c|}{0.9}       & \multicolumn{1}{c|}{2.5}         & \multicolumn{1}{c|}{84.01}       & 89.04       & \textbf{3.89}     &  \\ 
\multicolumn{1}{c|}{}                        & \multicolumn{1}{c|}{}                     &                      & \multicolumn{1}{c|}{}                       & \multicolumn{1}{c|}{1.0}       & \multicolumn{1}{c|}{3.08}        & \multicolumn{1}{c|}{88.88}       & 93.95       & 4.17              &  \\ \cline{4-9}
\multicolumn{1}{c|}{}                        & \multicolumn{1}{c|}{}                     &                      & \multicolumn{1}{c|}{\multirow{5}{*}{$U_{areal}$}}  & \multicolumn{1}{c|}{0.1}       & \multicolumn{1}{c|}{0.30}        & \multicolumn{1}{c|}{24.97}       & 26.55       & 3.76              &  \\ 
\multicolumn{1}{c|}{}                        & \multicolumn{1}{c|}{}                     &                      & \multicolumn{1}{c|}{}                       & \multicolumn{1}{c|}{0.2}       & \multicolumn{1}{c|}{0.46}        & \multicolumn{1}{c|}{45.07}       & 51.29       & \textbf{3.64}     &  \\ 
\multicolumn{1}{c|}{}                        & \multicolumn{1}{c|}{}                     &                      & \multicolumn{1}{c|}{}                       & \multicolumn{1}{c|}{0.3}       & \multicolumn{1}{c|}{0.69}        & \multicolumn{1}{c|}{59.49}       & 68.19       & 3.78              &  \\ 
\multicolumn{1}{c|}{}                        & \multicolumn{1}{c|}{}                     &                      & \multicolumn{1}{c|}{}                       & \multicolumn{1}{c|}{0.4}       & \multicolumn{1}{c|}{1.00}        & \multicolumn{1}{c|}{70.13}       & 79.66       & 3.82              &  \\ 
\multicolumn{1}{c|}{}                        & \multicolumn{1}{c|}{}                     &                      & \multicolumn{1}{c|}{}                       & \multicolumn{1}{c|}{0.5}       & \multicolumn{1}{c|}{1.45}        & \multicolumn{1}{c|}{78.09}       & 87.42       & 3.91              &  \\ \cline{1-9}
\end{tabular}
}
\label{tab: abstudy_threshold}
\vspace{-0.15in}
\end{table*}



	

\begin{table}[!htb]
\caption{ \footnotesize Ablation study of different iteration numbers. All methods are trained on the Sceneflow dataset with ground truth and fintuned on the unlabeled KITTI2015 dataset. $U_{area}$ denotes the area-level uncertainty estimation.}
\vspace{-0.1in}
\centering
{
\begin{tabular}{c|c|c|c}
\hline
Backbone                         & Filter                       & iteration & D1\_all \\ \hline
\multirow{4}{*}{UCFnet}                               & \multirow{4}{*}{$U_{area}$}  & 0         & 5.2     \\ 
                               &                              & 1         & 3.64    \\ 
                               &                              & 2         & 3.14    \\  
                               &                              & 3         & 3.16    \\ \hline
\end{tabular}
}
\label{tab: iteration}
\vspace{0.1in}

\caption{ \footnotesize Ablation study of area-level uncertainty estimation network module.  All methods are trained on the Sceneflow dataset with ground truth and tested on the KITTI2015 dataset. Density denotes the percentage of valid pixels in the generated proxy label. The approach which is used in our final model is underlined.}
\vspace{-0.1in}
\centering
\resizebox{0.44\textwidth}{!}{
\begin{tabular}{c|ccc}
\hline
\multirow{2}{*}{Method}                                           & \multicolumn{3}{c}{Kitti}                                               \\ \cline{2-4} 
& \multicolumn{1}{c|}{D1\_all} & \multicolumn{1}{c|}{density} & overlap \\  \hline
{\ul Multi-modal input}                         & \multicolumn{1}{c|}{\textbf{0.46\%}}  & \multicolumn{1}{c|}{45.07\%}    & 51.29\% \\ 
Multi-modal input without uncertainty map  & \multicolumn{1}{c|}{0.81\%}  & \multicolumn{1}{c|}{48.16\%}    & 55.56\% \\ 
Multi-modal input without disp            & \multicolumn{1}{c|}{0.73\%}  & \multicolumn{1}{c|}{49.02\%}    & 56.83\% \\ 
Multi-modal input without left image     & \multicolumn{1}{c|}{0.55\%}  & \multicolumn{1}{c|}{45.30\%}    & 49.54\% \\ \hline
\end{tabular}
}
\label{tab: abstudy_area-level}
\vspace{-0.15in}
\end{table}

\subsection{Ablation Study}
To verify the effectiveness of different modules, we set a series of experiments in this section. Note that all methods are trained on the synthetic dataset first and then finetuned on the unlabeled target dataset with the proposed pseudo-labels. Generally, seven types of experiments have been executed here.

\noindent \textbf{Attention-based refinement module:} In the attention-based disparity refinement module, we only employ initial disparity estimation as input. Here, we test the impact of adding multi-modal input, i.e., left image, left image feature, and pixel-level uncertainty map. As shown in Tab. \ref{tab: abstudy_refinement}, adding multi-modal input doesn’t bring noticeable gain and we can obtain the best performance by only using initial disparity estimation as input.

\noindent \textbf{Domain adaptation with self-generated pseudo-label:} Two terms of proxy labels, i.e., ${D_{pixel}}$ and ${D_{area}}$ can be generated by pixel-level uncertainty estimation and area-level uncertainty estimation, respectively. Here, we test the impact of each proxy label for domain adaptation individually. As shown in the Tab. \ref{tab: abstudy_pixel_area}, the two terms of proxy labels can both promote the performance of the pre-training model on the target dataset and the improvement is consistent on both CFNet and UCFNet, which verifies the effectiveness of the proposed uncertainty-based pseudo-labels generation method. Moreover, the proposed area-level proxy label can achieve a larger gain due to the leveraging of multi-modal input and neighboring pixel information, e.g., the d1\_all error rate of CFNet can further decrease from 4.5\% to 3.88\% after employing ${D_{area}}$ as supervision.

\noindent \textbf{Threshold of proxy label generation:} Threshold $t$ is an essential hyperparameter in proxy label generation, which controls the density and reliability of the filtered proxy labels. Hence, we made a detailed ablation study to analyze the impact of different threshold settings for domain adaptation. As shown in Tab. \ref{tab: abstudy_threshold}, we show the tradeoff between accuracy and density of generated proxy labels in different threshold settings and the corresponding domain adaptation results. It can be seen from the table that the proposed uncertainty estimation can effectively evaluate the confidence of current disparity estimations, e.g., by removing 21.91\% of uncertain pixels ($U_{area}>0.5$), we decrease the D1\_all error rate by 72.12\% (from 5.2\% to 1.45\% in KITTI 2015 training set). Moreover, we can find that a more accurate while sparser pseudo-label is not always good for final domain adaptation performance and we should seek a suitable balance between accuracy and density, i.e., threshold 0.9 for pixel-level uncertainty estimation and threshold 0.2 for area-level uncertainty estimation in our experiment setting.

\begin{figure*}[!t]
	\centering
	\tabcolsep=0.05cm
	\begin{tabular}{c c c}
    \includegraphics[width=0.28\linewidth]{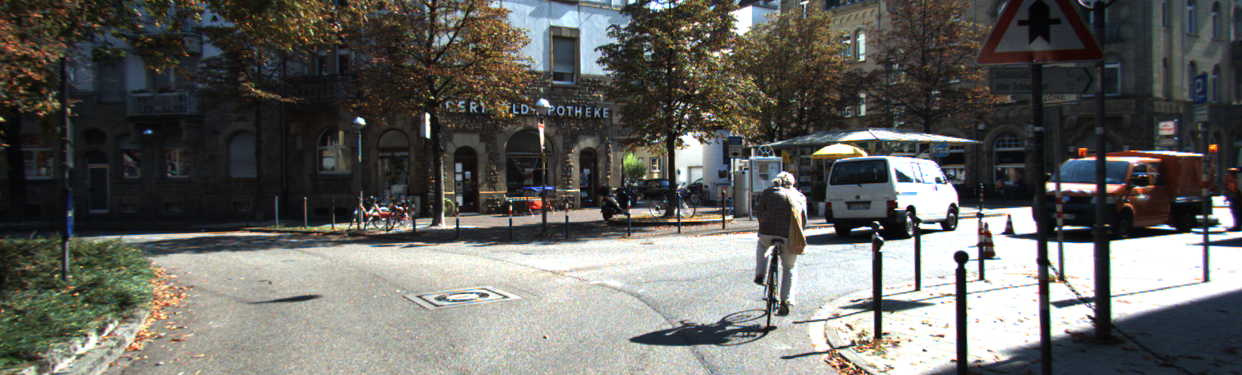}&
	\includegraphics[width=0.28\linewidth]{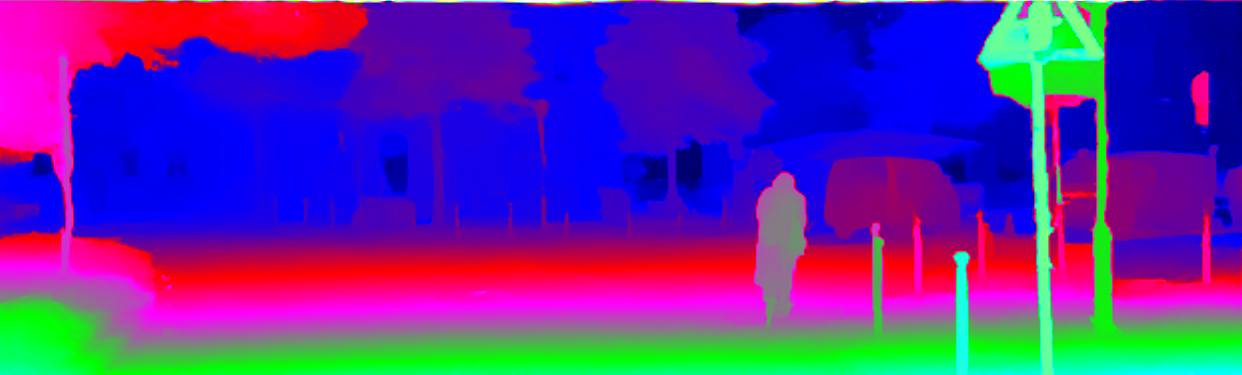}&
	\includegraphics[width=0.28\linewidth]{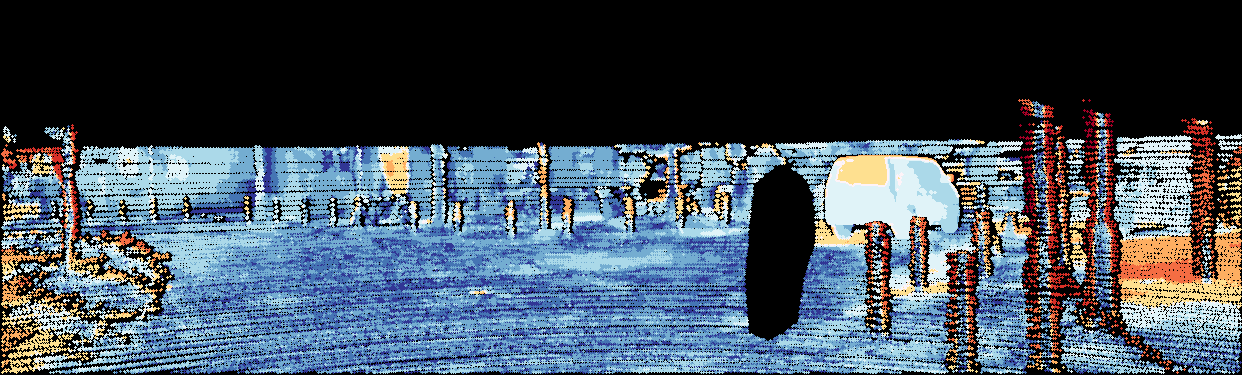} \\
	
	{\small (a) left image} &  {\small (b) original disparity map } &	{\small (c)  original error map}		 \\

    \includegraphics[width=0.28\linewidth]{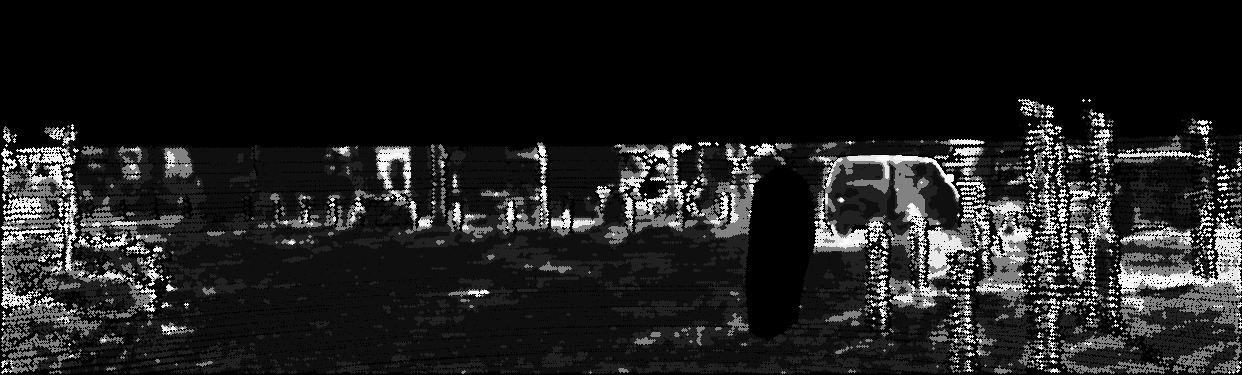}&
    \includegraphics[width=0.28\linewidth]{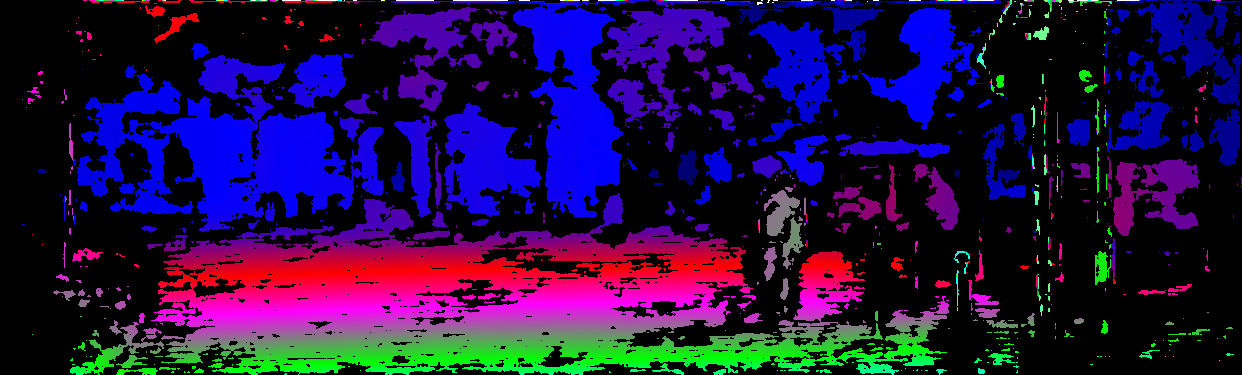}&
	\includegraphics[width=0.28\linewidth]{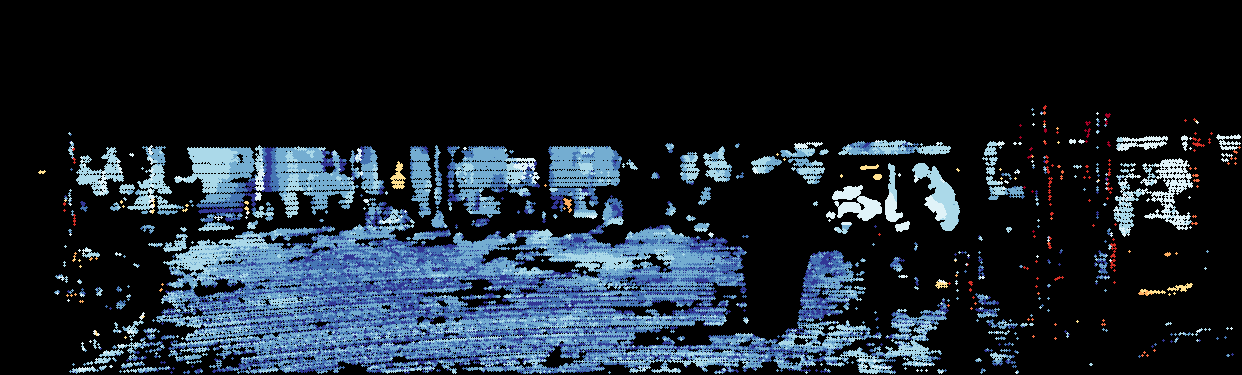} \\
	
	{\small (d) pixel-level uncertainty map $U_{pixel}$} &  {\small (e) filtered pseudo label $D_{pixel}$} &	{\small (f)  filtered error map $E_{pixel}$}		 \\

    \includegraphics[width=0.28\linewidth]{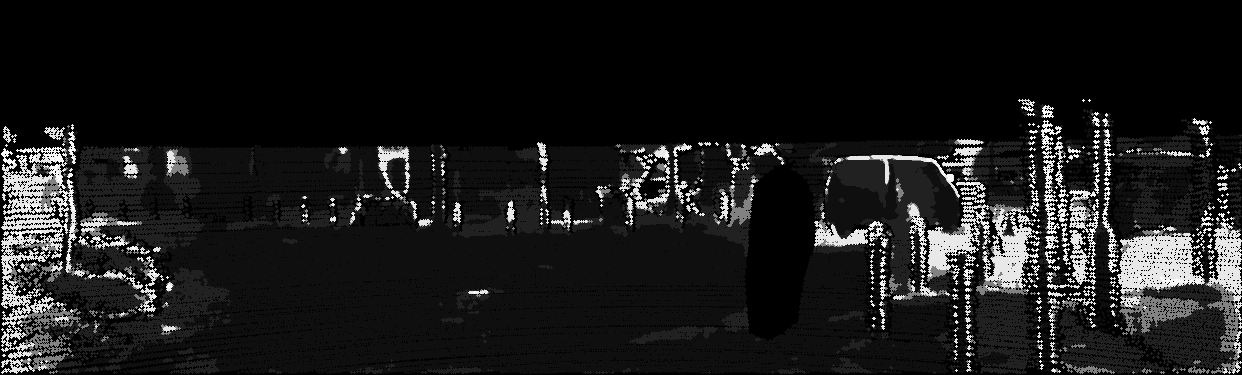}&
    \includegraphics[width=0.28\linewidth]{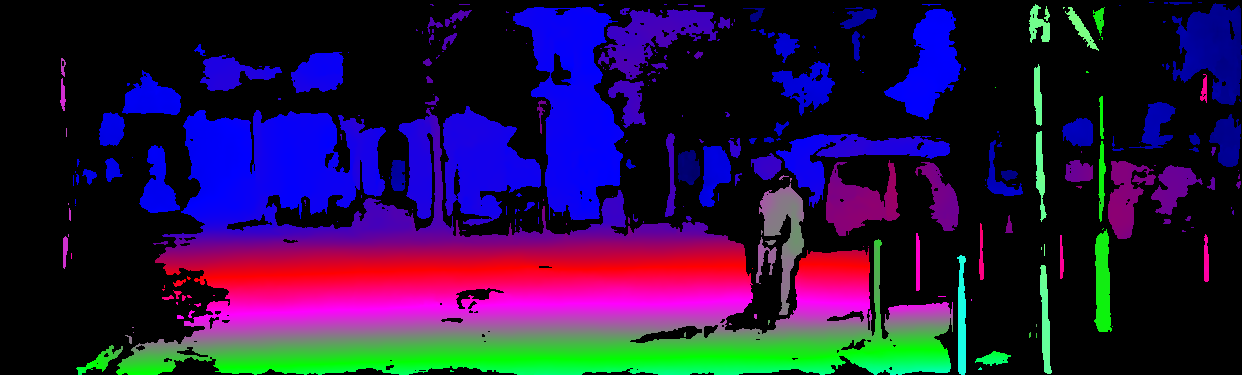}&
	\includegraphics[width=0.28\linewidth]{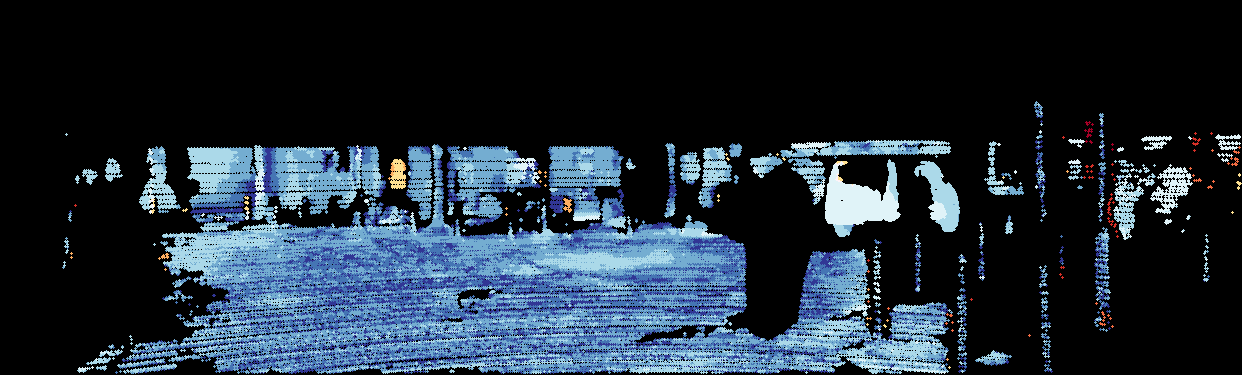} \\
	
	{\small (g) pixel-level uncertainty map $U_{area}$} &  {\small (h) filtered pseudo label $D_{area}$} &	{\small (i)  filtered error map $E_{area}$}		 \\




	


	\end{tabular}
	\vspace{-0.05in}
	\caption{\footnotesize Visualization of the generated pseudo label by two terms of uncertainty estimation. For each example, the first row shows the original predicted disparity, the second row shows the generated pseudo label by pixel-level uncertainty estimation $U_{pixel}$ and the third row shows generated pseudo label by area-level uncertainty estimation $U_{area}$.}
	\label{fig: uncertainty estimation}
\vspace{-0.15in}
\end{figure*}

\begin{figure}[!t]
    \centering
    \includegraphics[width=0.85\linewidth]{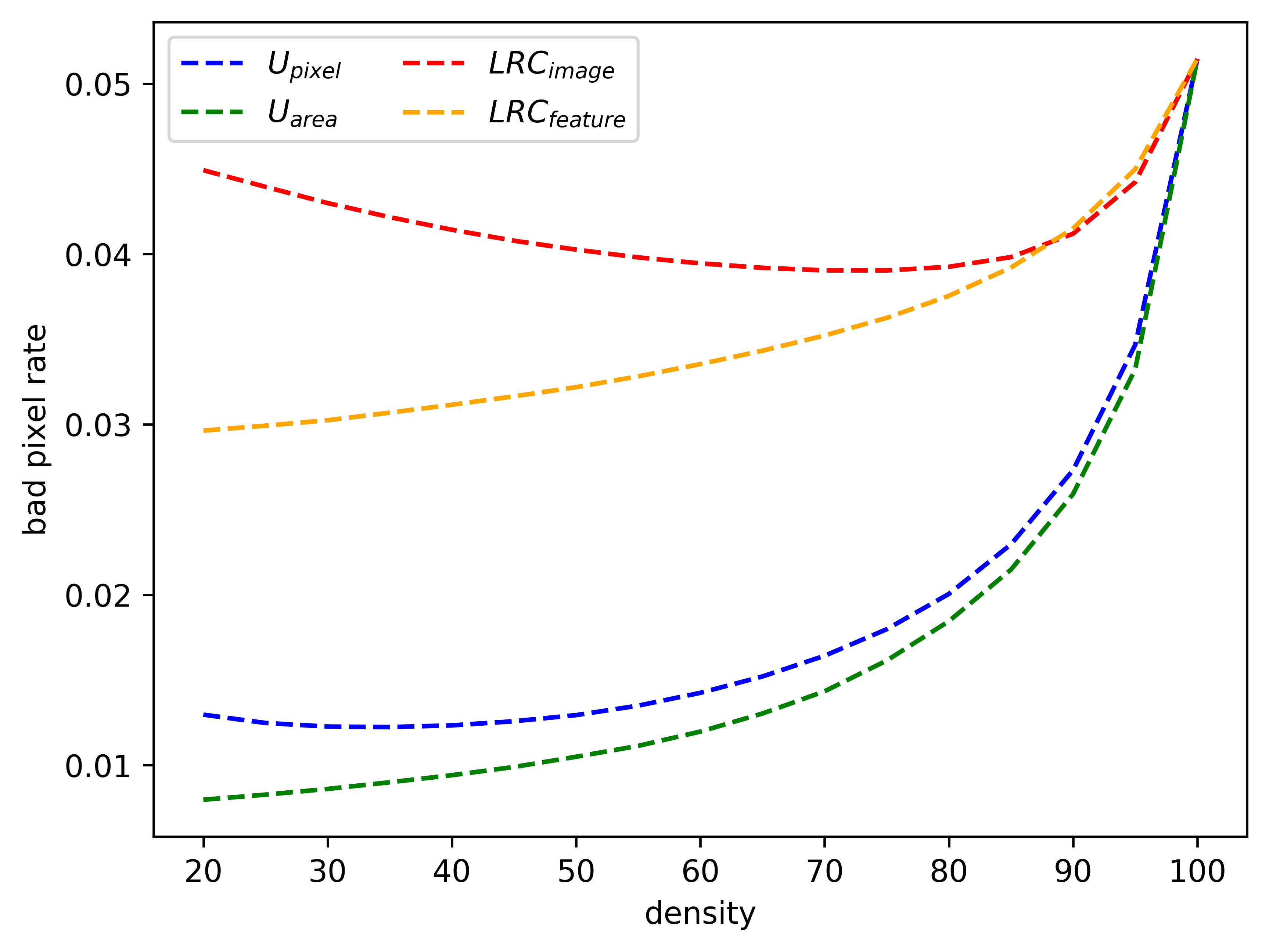} 
    \vspace{-0.2in}
    \caption{ \footnotesize The ROC curves of uncertainty estimation network on KITTI2015 dataset. D1\_all is used for evaluation (the lower the better). $U_{pixel}$ and $U_{area}$ denotes pixel-level and area-level uncertainty estimation result, respectively. $LRC_{image}$ and $LRC_{feature}$ denotes image-level and feature-level left-right consistency check, respectively. UCFNet\_pretrain is used as the main network.}
    \label{fig: roc_compare}
    \vspace{-0.15in}
\end{figure}

\noindent \textbf{Pixel-level uncertainty estimation vs area-level uncertainty estimation:}
Inspired by the standard evaluation metric in confidence estimation \cite{confidencesurvey, confidencesurvey2}, we propose to use the ROC curve and its area under curve (AUC) to quantitatively evaluate the performance of the proposed pixel-level and area-level uncertainty estimation. Specifically, we firstly sort pixels in the predicted disparity map following decreasing order of uncertainty. Then, we compute the D1\_all error rate on sparse maps obtained by iterative filtering (e.g., 5\% of pixels with higher uncertainty each time) from the dense map and plot the ROC curve, whose AUC quantitatively assesses the confidence effectiveness(the lower, the better). The ROC curve of the proposed method is shown in Fig. \ref{fig: roc_compare}. Note that we also plot the roc curve of the traditional uncertainty estimation method, i.e., image-level and feature-level left-right consistency check \cite{mcvmfc, crl} to further show the effectiveness of the proposed method. As shown in Fig. \ref{fig: roc_compare}, the proposed pixel-level and area-level uncertainty estimation can both generate a more accurate disparity map than the traditional left-right consistency check at any density and the area-level uncertainty estimation can achieve the best performance. Visualization of generated proxy labels is shown in Fig. \ref{fig: uncertainty estimation}. As shown, the generated two terms of uncertainty map (sub-figs (d) and (g)) are highly correlated with the error map. Hence, we can employ the proposed uncertainty estimation to filter out the high-uncertainty pixels of the original estimation and generate reliable pseudo labels. Moreover, the comparison between sub-figs (e) and (h) further shows the superiority of the proposed area-level uncertainty estimation. As shown, our area-level uncertainty estimation ${D_{area}}$ can employ the neighboring pixel information to better preserve the instance-level correct disparity estimation result, e.g., the pedestrian and pole of the input image.

\noindent \textbf{Iteration number of domain adaptation:} Our result can further be improved by iterative domain adaptation. Take two times of iteration as an example. At iteration 1, we employ the proxy label generated by the pre-training model to finetune the pre-training model. Then, at iteration 2, we can further finetune the pre-training model by employing the finetuned model to generate better proxy labels. More specifically, the generated pseudo-labels are employed to adapt the main network, refinement module, and area-level uncertainty estimation network one by one in each iteration. It can be seen from Tab. \ref{tab: iteration} that iterative domain adaptation can significantly improve the performance of disparity estimation on the target dataset, e.g., employing two iterations can decrease the D1\_all error rate from 3.64\% to 3.14\%. Note that, using more iterations cannot further improve estimation accuracy due to the optimization of pseudo-label has an upper limit by iterative domain adaptation.       

\begin{table}[!t]
\caption{\footnotesize \textbf{Top}: Cross-domain generalization evaluation on ETH3D, and KITTI training sets. All methods are only trained on the Scene Flow datatest and tested on full-resolution training images of two real datasets. \textbf{Bottom}: Adaptation generalization evaluation on ETH3D, and KITTI training sets. TDD: target domain data. All methods are finetuned on the unlabeled target domain data. 
* denotes not using additional data augmentation. We highlight the best result in \textbf{bold} and the second-best result in \B{blue} for each column.}
\vspace{-0.1in}
\centering
\resizebox{0.4\textwidth}{!}{
\begin{tabular}{c|c|c|c}
\hline
Method       & Training Set          & \begin{tabular}[c]{@{}c@{}}KITTI2015\\ D1\_all(\%)\end{tabular}  & \begin{tabular}[c]{@{}c@{}}ETH3D\\ bad 1.0(\%)\end{tabular} \\ \hline
\multicolumn{4}{c}{Cross-domain Generalization Evaluation}                                                                                                                                                                                                                                                        \\ \hline
CasStereo* \cite{cascade}    & synthetic                                                                        & 11.9                                                                                                                         & 7.8                                                         \\ 
DSMNet* \cite{dsmnet}       & synthetic                                                                      & 6.5                                                                                                                 & 6.2                                                         \\
ITSA-CFNet \cite{itsa}       & synthetic                                                                         & \textbf{4.7}                                                                                                               & \B{\textbf{5.1}}                                                         \\ 
\textbf{CFNet(ours)\cite{cfnet}}       & synthetic                                                              & 5.8                                                                                                               & 5.8                                                \\
\textbf{UCFNet\_pretrain*}       & synthetic                                                               & 7.8                                                                                                                & 5.9   \\
\textbf{UCFNet\_pretrain}       & synthetic                                                               & \B{\textbf{5.2}}                                                                                                                & \textbf{\textbf{4.8}}                                                \\ \hline
\multicolumn{4}{c}{Adaptation Generalization Evaluation}                                                                                                                                                                                                                                                          \\ \hline
AdaStereo \cite{adastereo,adastereo_ijcv}      & synthetic+TDD(no gt)                                                             & \B{\textbf{3.5}}                                                                                                                         & 4.1                                                           \\
\textbf{UCFNet\_adapt*} & synthetic+TDD(no gt)                                                     & 3.6                                                                                                                & \B{\textbf{4.0}} \\
\textbf{UCFNet\_adapt} & synthetic+TDD(no gt)                                                     & \textbf{3.1}                                                                                                                & \textbf{3.0} \\ \hline
\end{tabular}
}
\label{tab: data_augmentation}
	\vspace{-0.1in}
\end{table}
\noindent \textbf{Area-level uncertainty estimation:} In the area-level uncertainty estimation module, we employ multi-model input, i.e., pixel-level uncertainty map, initial disparity map, and left image to drive our network better evaluate the uncertainty of current disparity estimation. Here, we test the impact of each input individually. All methods are trained on the scene flow training dataset and then tested on the unseen kitti2015 dataset. As shown in Tab. \ref{tab: abstudy_area-level}, the result verifies all the multi-modal inputs work positively to filter out matching-error points and compared with other inputs, the pixel-level uncertainty map achieves the largest gain.

\begin{figure*}[!t]
	\centering
	\tabcolsep=0.05cm
	\begin{tabular}{c c c}
	

	

	
	
	\includegraphics[width=0.26\linewidth]{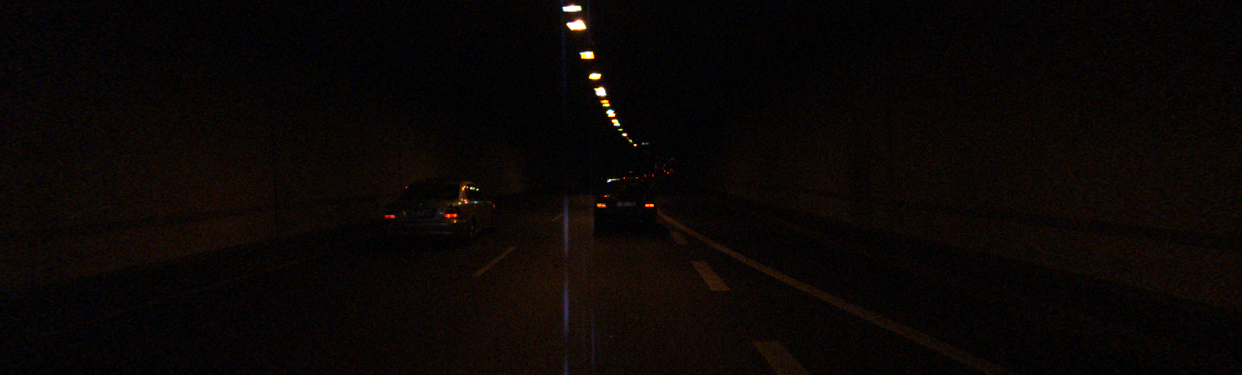}&
	\includegraphics[width=0.26\linewidth]{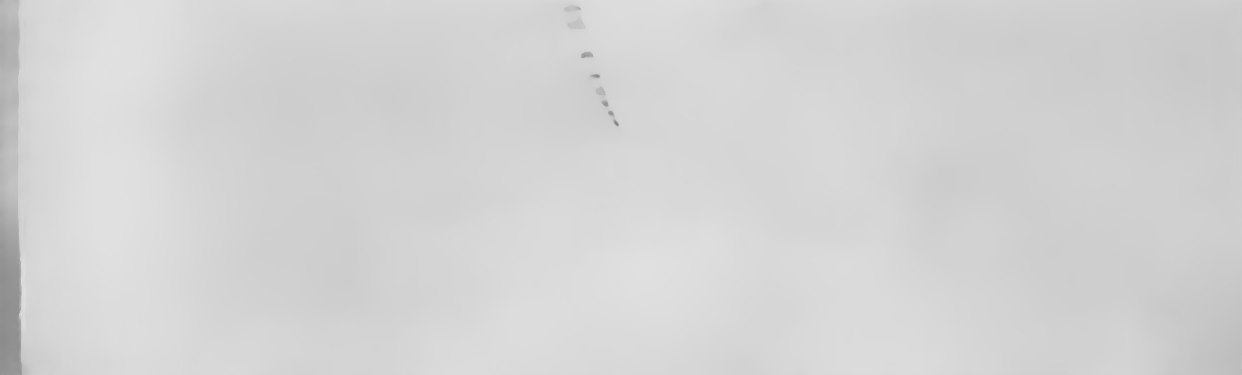}&
	\includegraphics[width=0.26\linewidth]{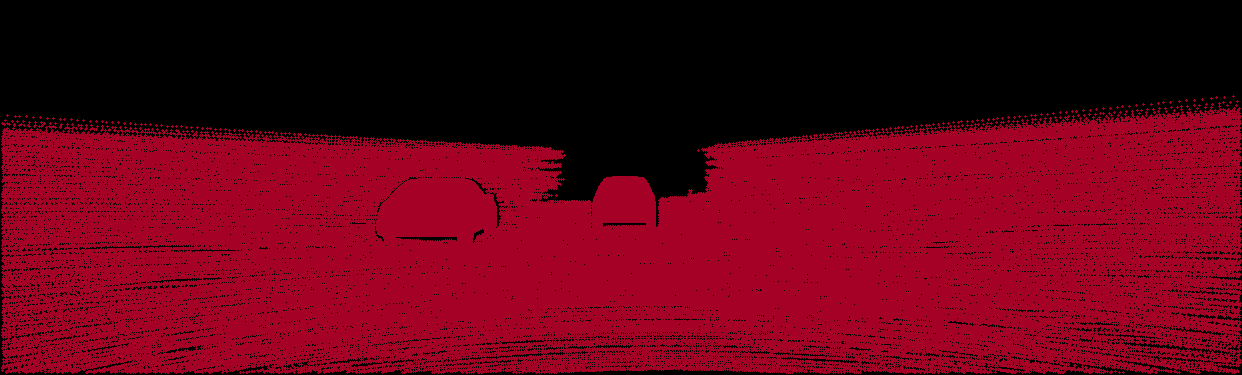} \\
		{\footnotesize (a) left image} &  {\footnotesize (b) original disparity map } &	{\footnotesize (c)  original error map}		 \\

    \includegraphics[width=0.26\linewidth]{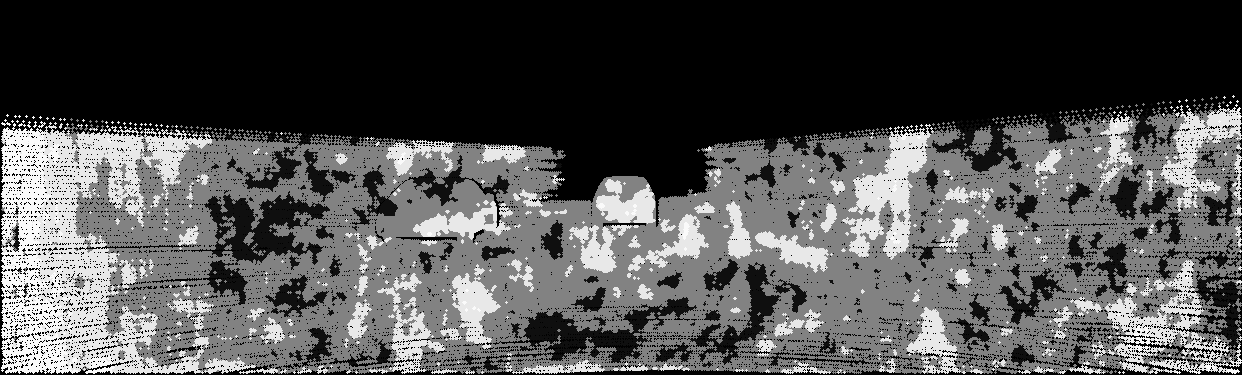}&
    \includegraphics[width=0.26\linewidth]{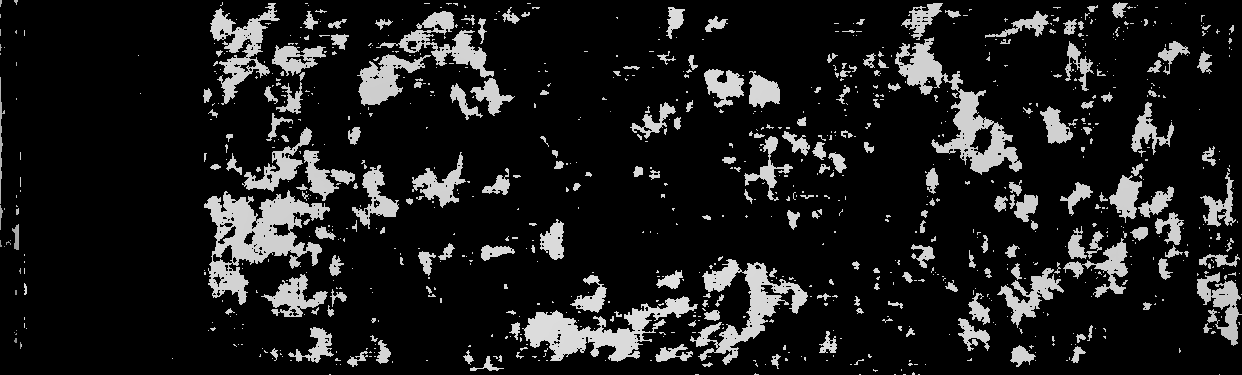}&
	\includegraphics[width=0.26\linewidth]{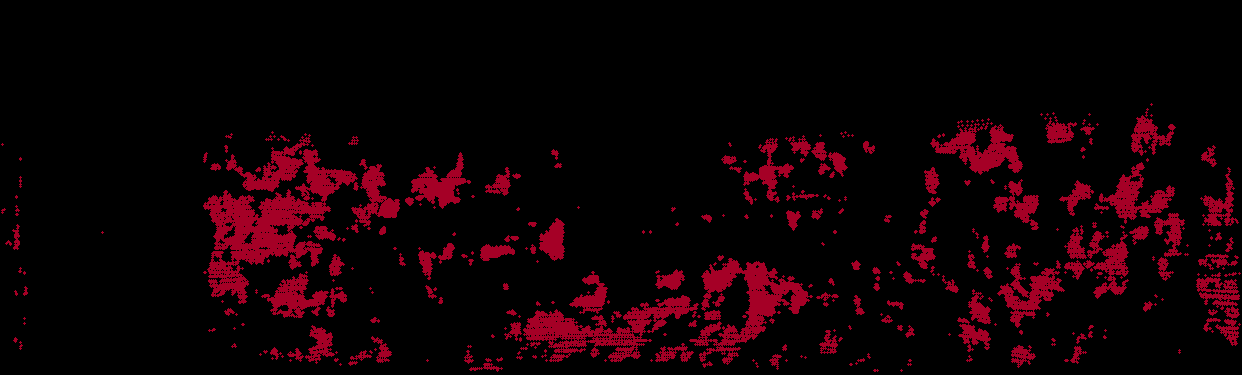} \\
		{\footnotesize (d) pixel-level uncertainty map $U_{pixel}$} &  {\footnotesize (e) filtered pseudo label $D_{pixel}$} &	{\footnotesize (f)  filtered error map $E_{pixel}$}		 \\

    \includegraphics[width=0.26\linewidth]{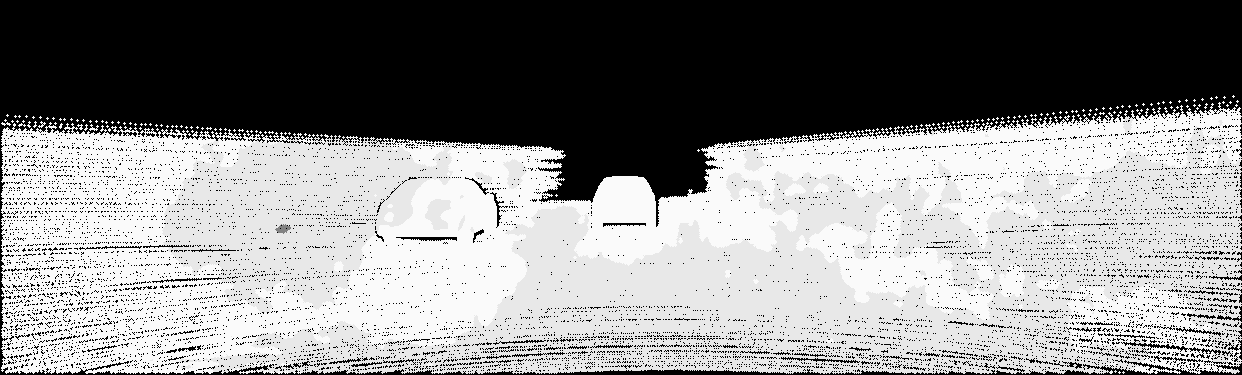}&
    \includegraphics[width=0.26\linewidth]{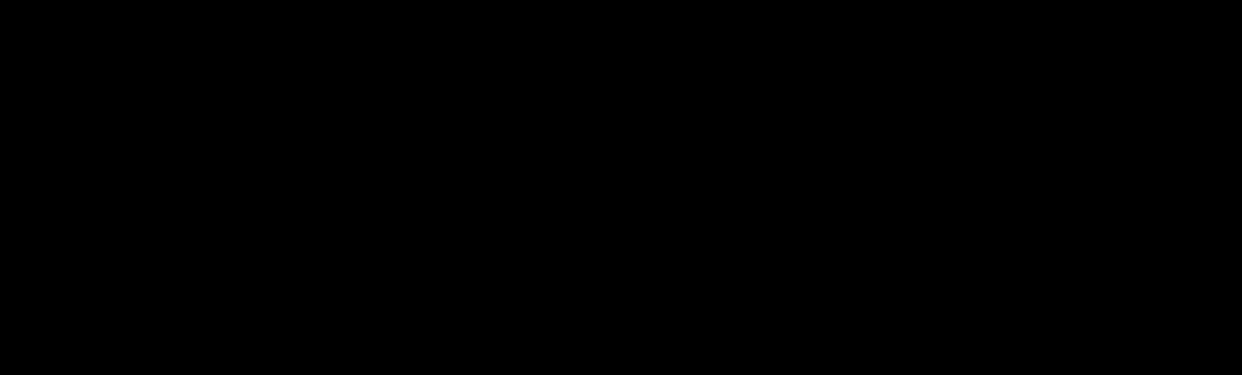}&
	\includegraphics[width=0.26\linewidth]{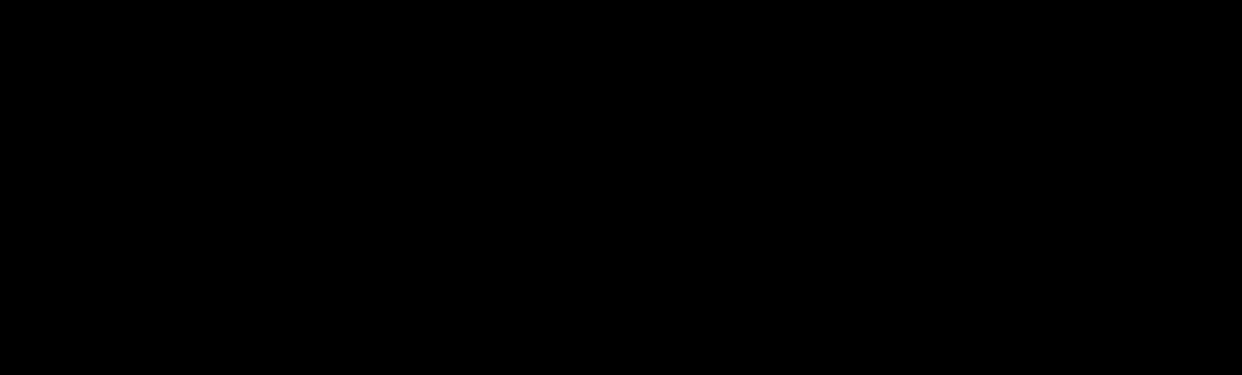} \\
		{\footnotesize (g) area-level uncertainty map $U_{area}$} &  {\footnotesize (h) filtered pseudo label $D_{area}$} &	{\footnotesize (i)  filtered error map $E_{area}$}		 \\

     \includegraphics[width=0.26\linewidth]{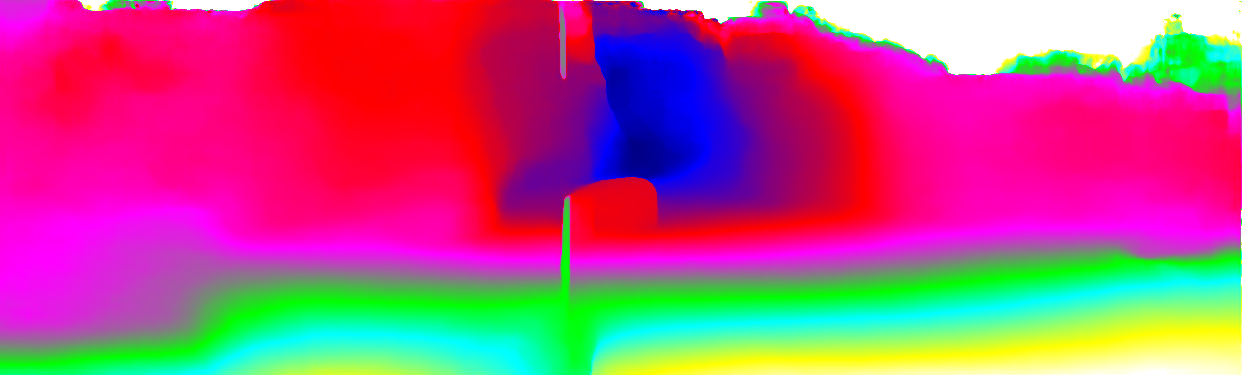}&
    \includegraphics[width=0.26\linewidth]{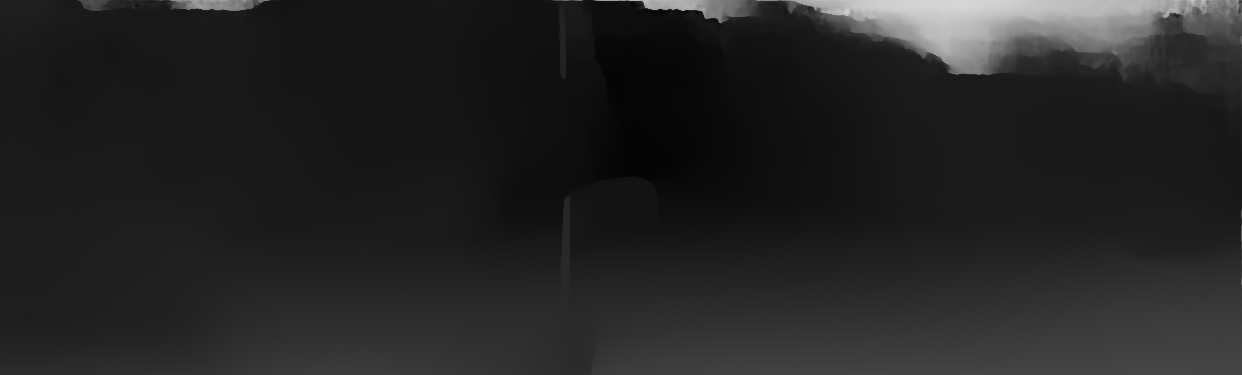}&
	\includegraphics[width=0.26\linewidth]{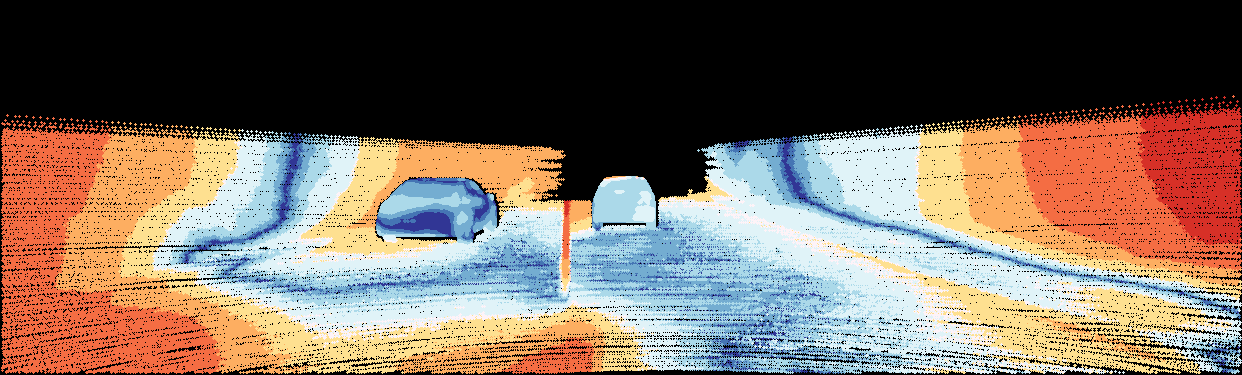} \\
		{\footnotesize (j) coloured disparity map after adaptation} &  {\footnotesize (k) disparity map after adaptation} &	{\footnotesize (m)  error map after adaptation}		 \\

	


	\end{tabular}
	\vspace{-0.1in}
	\caption{Extreme situation visualization of the generated pseudo label by two terms of uncertainty estimation. For each example, the first row shows the original predicted disparity, the second row shows the generated pseudo label by pixel-level uncertainty estimation $U_{pixel}$, the third row shows generated pseudo label by area-level uncertainty estimation $U_{area}$, and the fourth row shows the predicted disparity after domain adaptation.}
	\label{fig: extreme situation}
	\vspace{-0.15in}
\end{figure*}

\noindent \textbf{Data argumentation:} We also test the influence of removing additional data augmentation, e.g., asymmetric chromatic augmentation and occlusion on two real datasets. As shown in Tab. \ref{tab: data_augmentation}, the proposed method can still achieve a comparable result with the state-of-the-art method on both two terms of generalization even without data augmentation. For example, AdaStereo is the best-published domain adaptation method and indeed uses additional data augmentation. However, the proposed no data augmentation version UCFNet\_adapt* can still obtain comparable results on both ETH3D and KITTI datasets. Moreover, data augmentation works positively to improve the performance on all evaluation metrics and we recommend deploying it in the final model.

Additionally, as our original cascade and fused cost volume representation (CFNet) is mainly designed for joint generalization, we also perform various ablation studies to show the effectiveness of each network design in joint generalization evaluation. We divide 5 images from each real dataset (KITTI 2015, Middlebury, and ETH3D) as a validation set and use the rest of them as a training set to finetune our pretrain model. Results are shown in Tab. \ref{tab: ablation study}. Below we describe each component in more detail.

\noindent \textbf{Feature extraction:} We compare our pyramid feature extraction with the most widely used Resnet-like-network \cite{cascade,gwcnet}. As shown, our pyramid feature extraction can achieve similar performance with a faster speed, likely because the employing of small scale features is also helpful in feature extraction.

\noindent \textbf{Cost volume fusion:} We fuse three small-resolution cost volumes to generate the initial disparity map. Here, we test the impact when only a single volume is used. Cost volume fusion can achieve better performance with a slight additional computational cost.

\noindent \textbf{Cost volume cascade:} We test three ways of generating the next stage’s disparity searching space in cascade cost volume representation. As shown, learned parameters based pixel-level uncertainty estimation achieves the best performance with tiny additional computation complexity.



\noindent \textbf{Finetuning strategy:} We test three terms of finetuning strategy. As shown,  neither directly augmenting small datasets at the beginning (two-stages) nor only extending the number of iterations (three stages\_no augment) can improve the accuracy of predictions on small datasets. Instead, our strategy can greatly alleviate the problem of small datasets being overwhelmed by large ones.

\renewcommand\arraystretch{1.3}
\begin{table}[!t]
\centering
\caption{\footnotesize Ablation study results of the proposed network on KITTI 2015, Middlebury, and ETH3D validation set. PUE: pixel-level uncertainty estimation. We test a component of our method individually in each section of the table and the approach which is used in our final model (UCFNet without the refinement module) is underlined. Time is measured on the KITTI dataset by a single Tesla V100 GPU.}
\vspace{-0.1in}
\resizebox{0.45\textwidth}{!}{
\begin{tabular}{c|c|c|c|c|c}
\hline
Experiment                           & Method                           & \begin{tabular}[c]{@{}c@{}}KITTI\\ D1\_all\end{tabular} & \begin{tabular}[c]{@{}c@{}}Middlebury\\ bad 2.0\end{tabular} & \begin{tabular}[c]{@{}c@{}}ETH3D\\ bad 1.0\end{tabular} & \begin{tabular}[c]{@{}c@{}}time\\ (s)\end{tabular} \\ \hline
\multirow{2}{*}{Feature Extraction}  & Resnet-like-network              & 1.76                                                    & 22.81                                                         & \textbf{3.49}                                           & 0.270                                              \\  
                                     & \underline{Pyramid Feature Extraction} & \textbf{1.71}                                           & \textbf{22.27}                                               & 3.57                                                    & \textbf{0.225}                                     \\ \hline
\multirow{2}{*}{Cost Volume Fusion}  & Not Fuse                         & 1.79                                                    & 22.65                                                        & 3.67                                                    & \textbf{0.220}                                     \\  
                                     & \underline{Fuse}                       & \textbf{1.71}                                           & \textbf{22.27}                                               & \textbf{3.57}                                           & 0.225                                              \\ \hline
\multirow{3}{*}{Cost Volume Cascade} & Uniform Sample                   & 1.92                                                    & 23.8                                                         & 3.97                                                    & 0.225                                              \\  
                                     & PUE + Hyperparameters             & 1.78                                                    & 23.13                                                        & 3.83                                                    & 0.225                                              \\  
                                     & \underline{PUE + Learned Parameters}    & \textbf{1.71}                                           & \textbf{22.27}                                               & \textbf{3.57}                                           & \textbf{0.225}                                     \\ \hline
  \multirow{3}{*}{Fine-tuning strategy} & two stages               & \textbf{1.70}                                                    & 22.77                                                        & 3.99                                           & 0.225                                              \\ 
                                      & three stages\_no augment & \textbf{1.70}                                                    & 22.57                                                        & 3.92                                                    & 0.225                                              \\  
                                      & \underline {three stages}       & 1.71                                           & \textbf{22.27}                                               & \textbf{3.57}                                                    & \textbf{0.225}                                     \\ \hline
\end{tabular}
}
\label{tab: ablation study}
\vspace{-0.1in}
\end{table}

\renewcommand\arraystretch{1}

\subsection{Extreme Situation}
As mentioned in Sec. \ref{Sec. Robustness Evaluation}, our pre-training model has strong cross-domain generalization and can generate relatively reasonable results for the subsequent pseudo-label generation. However, it still cannot work in some extreme situations, e.g., pictures taken inside a tunnel. As shown in Fig. \ref{fig: extreme situation}, we give some extreme cases, in which our pre-training method predicts a totally wrong result. As for such cases, the proposed pixel-level uncertainty estimation (sub-figs (f)) method is difficult to filter out all errors in the disparity estimation while area-level uncertainty estimation (sub-figs (i)) can achieve this goal and generate a disparity map without valid pixels, making the noisy labels don’t affect subsequent domain adaptation. Moreover, the proposed UCF\_adapt can indeed predict a reasonable result in such extreme situations (sub-figs (m)) even if we don’t have corresponding valid proxy labels for these extreme cases, which further verifies the effectiveness of the proposed domain adaptation method.

\section{Monocular depth estimation Experiments}
\label{sec:monocular}
Recall our goal is to push methods to be robust and perform well across different datasets without using the ground truth of the target domain. This is same in the monocular depth estimation setting. Indeed, as the monocular depth estimation is an ill-posed problem, it even needs more annotated data. However, acquiring labeled real-world data is cumbersome and costly in most practical settings, e.g., expensive LiDAR with careful calibration is required to obtain depth ground truth in outdoor scenes. Instead, stereo matching is a cheaper option. Hence, we propose to explore the feasibility of eliminating the need for lidar and using stereo matching methods to collect ground truth data. That is the proposed stereo matching method is served as the offline ground truth collection system and the monocular depth estimation network is the deployed online depth estimation module. Following this motivation, we select to use the stereo matching model trained on the synthetic dataset and unlabeled target domain data (UCFNet\_adapt) to generate the pseudo-label for the training of monocular depth estimation. Note that both state-of-the-art supervised and self-supervised monocular depth estimation approaches are compared here.

\subsection{Dataset}

We use the KITTI dataset~\cite{kitti1} as the training dataset which consists of calibrated videos registered to LiDAR measurements of city scenarios. The depth evaluation is done on the LiDAR pointcloud. Following~\cite{eigen2014depth}\cite{monodepth2_iccv2019}\cite{depth_hint_iccv2019}, seven standard metrics, named ''Abs Rel'', ''Sq Rel'', ''RMSE'', ''RMSE log'', ''$\delta < 1.25 $'', ''$\delta < 1.25^2 $'' and ''$\delta < 1.25^3 $'' are used to evaluate the performance of the predicted depth information. Tab.~\ref{tab:metrics} demonstrates the definition of each evaluation metric, and please see~\cite{eigen2014depth} for evaluation details. 

\textbf{Training mode:} According to different training modes, three kinds of results are provided here. (1). Results of self-supervised monocular depth estimation approaches, which do not need the supervision of ground truth; (2). Results of supervised monocular depth estimation approaches with ground truth as supervision; (3). Results of supervised monocular depth estimation approaches with generated pseudo-labels as supervision. Moreover, as these methods eliminate the need for ground truth, they can be regarded as \textbf{unsupervised approaches}.

\begin{table}[!t]
    \centering
    \caption{\footnotesize Detailed evaluation metrics of monocular depth estimation, where $n$ is the number of pixels, $y_{pred}$ and $y_{gt}$ are the estimated depth and ground truth depth, respectively. $y_{{pred}_i}$ and $y_{{gt}_i}$ denote the $i_{th}$ pixel in the estimated and ground truth depth map. $T$ is the threshold.}
    \vspace{-0.1in}
    \begin{tabular}{|c|c|} \hline
        Abs Rel & $\frac{1}{n} \sum \frac{y_{pred} - y_{gt}}{y_{gt}}$ \\ \hline
        Sq Rel & $\frac{1}{n} \sum (\frac{y_{pred} - y_{gt}}{y_{gt}})^2$ \\ \hline
        RMSE & $\sqrt{\frac{1}{n}\sum{(y_{pred} - y_{gt})^2}}$  \\ \hline 
        RMSE log & $\sqrt{\frac{1}{n}\sum{(log(y_{pred}) - log(y_{gt}))^2}}$ \\ \hline
        $\delta$ & $max(\frac{y_{{pred}_i}}{y_{{gt}_i}}, \frac{y_{{gt}_i}}{y_{{pred}_i}}) < T$\\ \hline
    \end{tabular}
    
    \label{tab:metrics}
    \vspace{-0.1in}
\end{table}

\begin{table*}[!t]
\caption{\footnotesize Quantitative evaluations on the kitti dataset using the test split of Eigen et al. * denotes the performance is evaluated using the official annotated ground truth (default: using raw velodyne data). Note that we use the cropping strategy introduced by Garg et al.}
\vspace{-0.1in}
\centering
\footnotesize
{
\begin{tabular}{c|c|ccc|cccc}
\hline
\multirow{2}{*}{Method} & \multirow{2}{*}{GT} & \multicolumn{1}{c|}{$\delta  < {1.25}$} & \multicolumn{1}{c|}{$\delta  < {1.25^2}$} & $\delta  < {1.25^3}$ & \multicolumn{1}{c|}{Abs Rel} & \multicolumn{1}{c|}{Sq Rel} & \multicolumn{1}{c|}{RMSE}  & RMSE log \\ \cline{3-9} 
                  &     & \multicolumn{3}{c|}{Higher value is better}                                                        & \multicolumn{4}{c}{Lower value is better}                                                         \\ \hline
                        
\rowcolor{gbypink}{Godard \textit{ et al.}\cite{godard2017unsupervised}}        & N & \multicolumn{1}{c|}{0.861}            & \multicolumn{1}{c|}{0.949}             & 0.976             & \multicolumn{1}{c|}{0.114}   & \multicolumn{1}{c|}{0.898}  & \multicolumn{1}{c|}{4.935} & 0.206    \\
\rowcolor{gbypink}{Kuznietsov \textit{ et al.}  \cite{Kuznietsov}}                & N & \multicolumn{1}{c|}{0.862}            & \multicolumn{1}{c|}{0.960}             & 0.986             & \multicolumn{1}{c|}{0.113}   & \multicolumn{1}{c|}{0.741}  & \multicolumn{1}{c|}{4.621} & 0.189    \\
\rowcolor{gbypink}{Monodepth2~\cite{monodepth2_iccv2019}} & N &\multicolumn{1}{c|}{0.876} & \multicolumn{1}{c|}{0.957} & 0.980 & \multicolumn{1}{c|}{0.106} & \multicolumn{1}{c|}{0.806} & \multicolumn{1}{c|}{4.630} & 0.193  \\
\rowcolor{gbypink}{Monodepth2 dh~\cite{depth_hint_iccv2019}} & N &\multicolumn{1}{c|}{0.888} & \multicolumn{1}{c|}{0.962} & 0.982 & \multicolumn{1}{c|}{0.100} & \multicolumn{1}{c|}{0.757} & \multicolumn{1}{c|}{4.490} & 0.185 \\
\rowcolor{gbypink}{MLDA-Net~\cite{song2021mlda}} & N &\multicolumn{1}{c|}{0.887} & \multicolumn{1}{c|}{0.963} & 0.983 & \multicolumn{1}{c|}{0.099} & \multicolumn{1}{c|}{0.724} & \multicolumn{1}{c|}{4.415} & 0.183 \\
\rowcolor{gbypink}{monoResMatch~\cite{monoresmatch}} & N &\multicolumn{1}{c|}{0.890} & \multicolumn{1}{c|}{0.961} & 0.981 & \multicolumn{1}{c|}{0.096} & \multicolumn{1}{c|}{0.673} & \multicolumn{1}{c|}{4.351} & 0.184 \\ 
\rowcolor{gbypink}{SD-SSMDE~\cite{SD-SSMDE}} & N &\multicolumn{1}{c|}{0.902} & \multicolumn{1}{c|}{0.968} & 0.985 & \multicolumn{1}{c|}{0.098} & \multicolumn{1}{c|}{0.674} & \multicolumn{1}{c|}{4.187} & 0.170 \\ \hline  

\rowcolor{gbygray}{Eigen\textit{ et al.} \cite{eigen}}                & Y & \multicolumn{1}{c|}{0.692}            & \multicolumn{1}{c|}{0.899}             & 0.967             & \multicolumn{1}{c|}{0.190}   & \multicolumn{1}{c|}{1.515}  & \multicolumn{1}{c|}{7.156} & 0.270    \\
\rowcolor{gbygray}{Liu \textit{ et al.}\cite{liu}}             &   Y  & \multicolumn{1}{c|}{0.647}            & \multicolumn{1}{c|}{0.882}             & 0.961             & \multicolumn{1}{c|}{0.217}   & \multicolumn{1}{c|}{1.841}  & \multicolumn{1}{c|}{6.986} & 0.289    \\

\rowcolor{gbygray}{Gan et al \textit{ et al.}   \cite{gan}}              & Y & \multicolumn{1}{c|}{0.890}            & \multicolumn{1}{c|}{0.964}             & 0.985             & \multicolumn{1}{c|}{0.098}   & \multicolumn{1}{c|}{0.666}  & \multicolumn{1}{c|}{3.933} & 0.173    \\
\rowcolor{gbygray}{DORN     \cite{dorn}}           & Y & \multicolumn{1}{c|}{0.897}            & \multicolumn{1}{c|}{0.966}             & 0.986             & \multicolumn{1}{c|}{0.099}   & \multicolumn{1}{c|}{0.593}  & \multicolumn{1}{c|}{3.714} & 0.161    \\ \hline
\rowcolor{gbygreen}{Bts~\cite{bts}}                & Y & \multicolumn{1}{c|}{0.913}            & \multicolumn{1}{c|}{0.970}             & 0.985             & \multicolumn{1}{c|}{0.082}   & \multicolumn{1}{c|}{0.484}  & \multicolumn{1}{c|}{3.694} & 0.166 \\
\rowcolor{gbygreen}{bts\_pseudo\_label}                & N  & \multicolumn{1}{c|}{0.910}            & \multicolumn{1}{c|}{0.970}             & 0.986             & \multicolumn{1}{c|}{0.094}   & \multicolumn{1}{c|}{0.554}  & \multicolumn{1}{c|}{3.730} & 0.167 \\
\rowcolor{gbygreen}{bts\_pseudo\_label\_full}              &   N  & \multicolumn{1}{c|}{\textbf{0.939}}            & \multicolumn{1}{c|}{\textbf{0.975}}             & \textbf{0.987}             & \multicolumn{1}{c|}{\textbf{0.072}}   & \multicolumn{1}{c|}{\textbf{0.454}}  & \multicolumn{1}{c|}{\textbf{3.222}} &\textbf{ 0.147} \\ \hline
\rowcolor{gbygreen}{lapdepth~\cite{lapdepth}}              &   Y   & \multicolumn{1}{c|}{0.915}            & \multicolumn{1}{c|}{0.970}             & 0.985             & \multicolumn{1}{c|}{0.083}   & \multicolumn{1}{c|}{0.481}  & \multicolumn{1}{c|}{3.658} & 0.165 \\
\rowcolor{gbygreen}{lapdepth\_pseudo\_label}               &  N  & \multicolumn{1}{c|}{0.909}            & \multicolumn{1}{c|}{0.972}             & \textbf{0.989}             & \multicolumn{1}{c|}{0.089}   & \multicolumn{1}{c|}{0.520}  & \multicolumn{1}{c|}{3.623} & 0.162    \\ 
\rowcolor{gbygreen}{lapdepth\_pseudo\_label\_full}               &  N  & \multicolumn{1}{c|}{\textbf{0.936}}            & \multicolumn{1}{c|}{\textbf{0.975}}             & 0.987             & \multicolumn{1}{c|}{\textbf{0.075}}   & \multicolumn{1}{c|}{\textbf{0.443}}  & \multicolumn{1}{c|}{\textbf{3.160}} & \textbf{0.147}    \\ 
\hline
\rowcolor{gbypink}{Godard* \textit{ et al.}} \cite{godard2017unsupervised}              &  N & \multicolumn{1}{c|}{0.916}            & \multicolumn{1}{c|}{0.980}             & 0.994             & \multicolumn{1}{c|}{0.085}   & \multicolumn{1}{c|}{0.584}  & \multicolumn{1}{c|}{3.938} & 0.135    \\ 
\rowcolor{gbypink}{Kuznietsov* \textit{ et al.}} \cite{Kuznietsov}           & N & \multicolumn{1}{c|}{0.906}            & \multicolumn{1}{c|}{0.980}             & 0.995             & \multicolumn{1}{c|}{0.138}   & \multicolumn{1}{c|}{0.478}  & \multicolumn{1}{c|}{3.60}  & 0.138    \\ \hline
\rowcolor{gbygray}{Amiri*  \textit{ et al.}}  \cite{amiri}             &  Y & \multicolumn{1}{c|}{0.923}            & \multicolumn{1}{c|}{0.984}             & 0.995             & \multicolumn{1}{c|}{0.078}   & \multicolumn{1}{c|}{0.417}  & \multicolumn{1}{c|}{3.464} & 0.126    \\ 
\rowcolor{gbygray}{DORN*  \textit{ et al.}}   \cite{dorn}             & Y  & \multicolumn{1}{c|}{0.932}            & \multicolumn{1}{c|}{0.984}             & 0.994             & \multicolumn{1}{c|}{0.072}   & \multicolumn{1}{c|}{0.307}  & \multicolumn{1}{c|}{2.727} & 0.120    \\ \hline
\rowcolor{gbygreen}{Bts* \cite{bts}}          & Y & \multicolumn{1}{c|}{0.960}            & \multicolumn{1}{c|}{0.994}          & 0.999             & \multicolumn{1}{c|}{0.060}   & \multicolumn{1}{c|}{0.210}  & \multicolumn{1}{c|}{2.459} & 0.093    \\ 
\rowcolor{gbygreen}{Bts\_pseudo\_label*}     & N & \multicolumn{1}{c|}{0.956}                & \multicolumn{1}{c|}{0.993}                 & 0.998                 & \multicolumn{1}{c|}{0.064}   & \multicolumn{1}{c|}{0.243}  & \multicolumn{1}{c|}{2.576} & 0.099    \\ 
\rowcolor{gbygreen}{Bts\_pseudo\_label\_full*}     & N & \multicolumn{1}{c|}{\textbf{0.982}}                & \multicolumn{1}{c|}{\textbf{0.997}}                 & \textbf{0.999}                 & \multicolumn{1}{c|}{\textbf{0.043}}   & \multicolumn{1}{c|}{\textbf{0.127}}  & \multicolumn{1}{c|}{\textbf{1.845}} & \textbf{0.070}    \\ \hline
\rowcolor{gbygreen}{lapdepth* \cite{lapdepth}}     & Y & \multicolumn{1}{c|}{0.961}            & \multicolumn{1}{c|}{0.994}             & 0.999             & \multicolumn{1}{c|}{0.061}   & \multicolumn{1}{c|}{0.206}  & \multicolumn{1}{c|}{2.398} & 0.092    \\ 
\rowcolor{gbygreen}{lapdepth\_pseudo\_label*} &  N & \multicolumn{1}{c|}{0.952}           & \multicolumn{1}{c|}{0.992}             & 0.998             & \multicolumn{1}{c|}{0.067}   & \multicolumn{1}{c|}{0.263}  & \multicolumn{1}{c|}{2.598} & 0.100
\\ 
\rowcolor{gbygreen}{lapdepth\_pseudo\_label\_full*} &  N & \multicolumn{1}{c|}{\textbf{0.979}}           & \multicolumn{1}{c|}{\textbf{0.996}}             & \textbf{0.999}             & \multicolumn{1}{c|}{\textbf{0.046}}   & \multicolumn{1}{c|}{\textbf{0.134}}  & \multicolumn{1}{c|}{\textbf{1.847}} & \textbf{0.074}
\\ \hline
\end{tabular}
}
\label{tab: mono_result}
\vspace{-0.1in}
\end{table*}

\subsection{Implementation Details}
We select two representative supervised monocular depth estimation models LapDepth$\footnote{\href{https://github.com/tjqansthd/LapDepth-release/} {LapDepth} uses GPL-3.0 License and we download the code from their official GitHub website.}$ and BTS$\footnote{\href{ https://github.com/cleinc/bts/}{BTS} uses GPL-3.0 License and we download the code from their official GitHub website.}$ to verify the effectiveness of the proposed pseudo label in monocular depth estimation. All settings remain the same as in their original paper except that we employ the generated pseudo-labels rather than ground truth to supervise the network. Specifically, the whole training process can be divided into three steps: Firstly, we employ the synthetic data (source domain) and the unlabeled stereo images of the real dataset (target domain) to train a robust stereo matching network UCFNet\_adapt. Secondly, we feed the synchronized stereo images of the KITTI raw dataset into the UCFNet\_adapt and employ the proposed uncertainty-based pseudo-label generation method to generate corresponding pseudo-labels. Thirdly, we employ the generated pseudo-labels as supervision to train the selected two representative supervised monocular depth estimation methods: LapDepth and BTS.

\subsection{Comparisons among different training modes}
Tab. \ref{tab: mono_result} demonstrates the results among different training modes. In specific, pink areas mean results obtained by self-supervised approaches, purple areas mean results obtained by supervised approaches with ground truth supervision, and green areas mean results obtained by supervised approaches with our generated pseudo-labels. ''*'' means results evaluated by the official annotated ground truth of KITTI, where the ground truth is obtained by combing multi-frame of the point cloud, while default (without ''*'') means results evaluated by raw point cloud data of KITTI.

Note that, the ground truth is not needed in the generation of pseudo-labels, therefore, approaches supervised with our generated pseudo-labels (\emph{Bts\_pseudo\_label}, \emph{Bts\_pseudo\_label\_full}, \emph{lapdepth\_pseudo\_label} and \emph{lapdepth\_pseudo\_label\_full} in Tab.~\ref{tab: mono_result}) can be regarded as unsupervised approaches.

\begin{figure*}[!t]
	\centering
	\tabcolsep=0.05cm
	\begin{tabular}{c c c c}
	
	
    \includegraphics[width=0.22\linewidth]{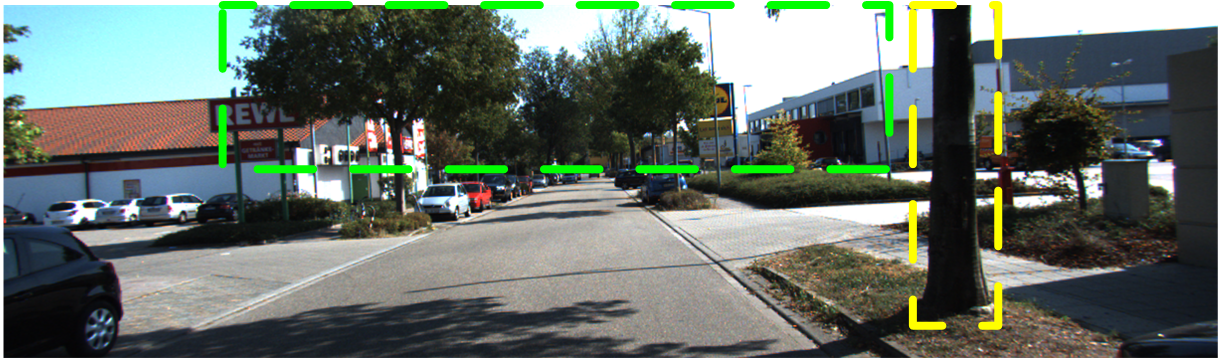}&
	\includegraphics[width=0.22\linewidth]{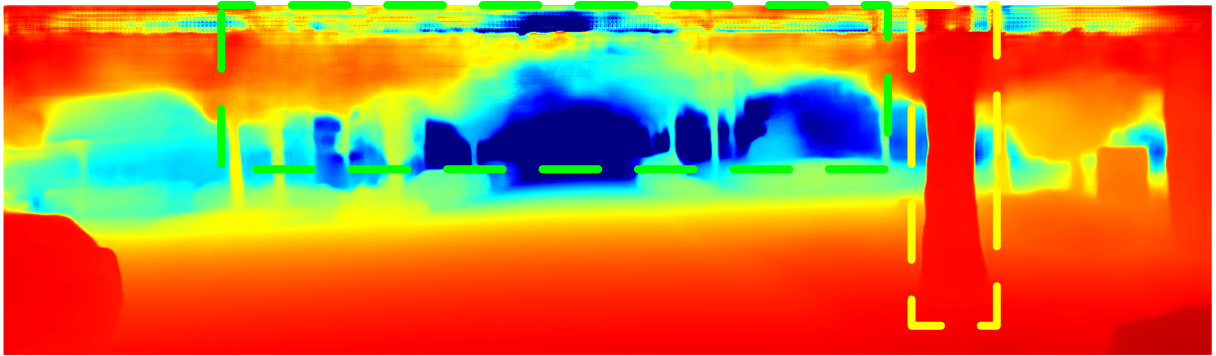}&
	\includegraphics[width=0.22\linewidth]{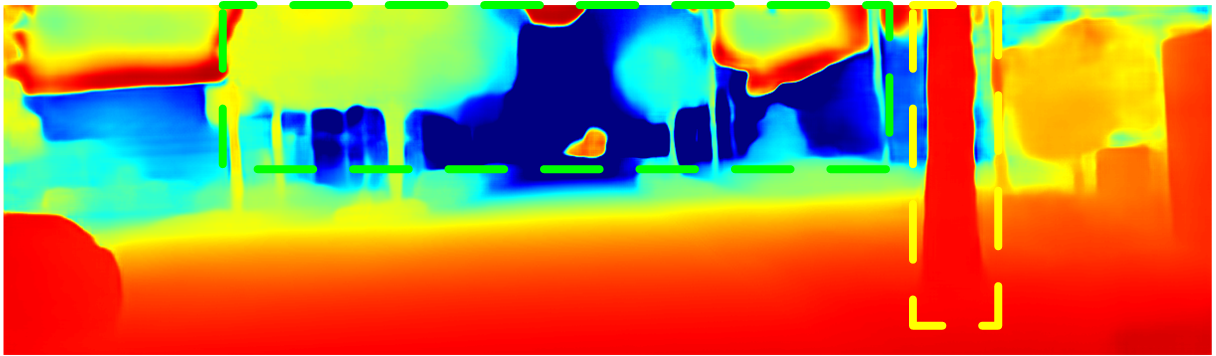}&
	\includegraphics[width=0.22\linewidth]{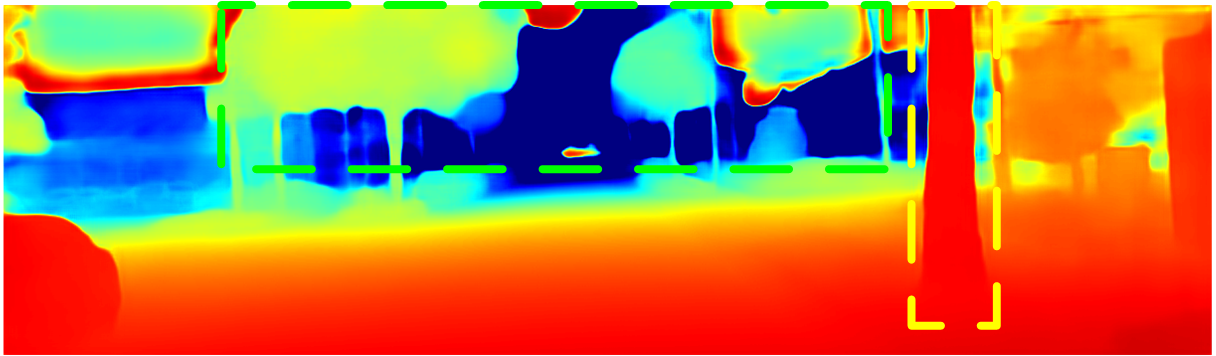}\\
	
	& \includegraphics[width=0.22\linewidth]{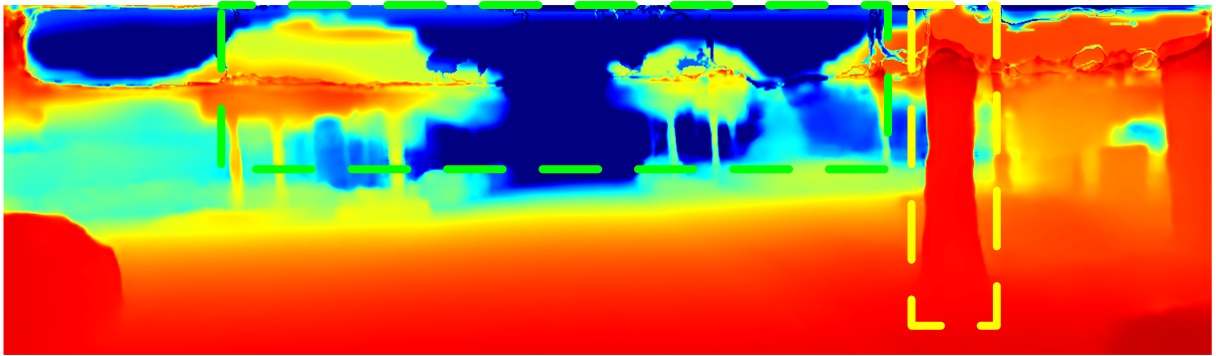}&
	\includegraphics[width=0.22\linewidth]{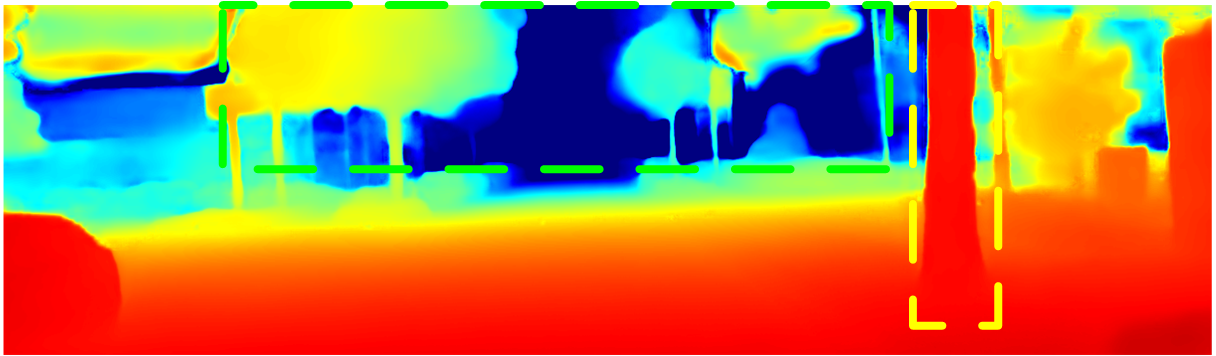}&
	\includegraphics[width=0.22\linewidth]{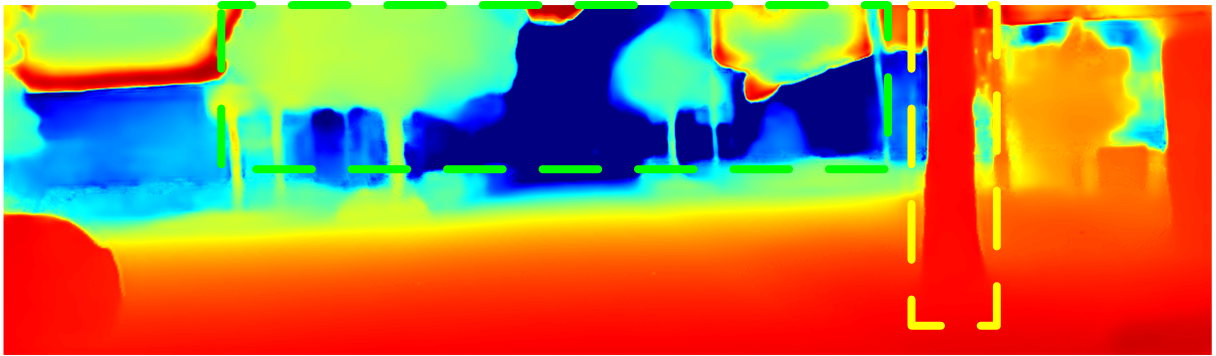}\\

	
	
	\includegraphics[width=0.22\linewidth]{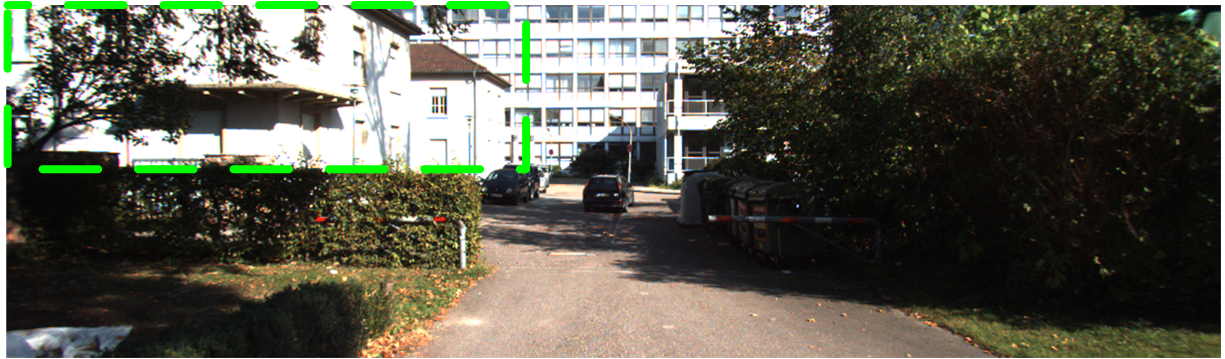}&
	\includegraphics[width=0.22\linewidth]{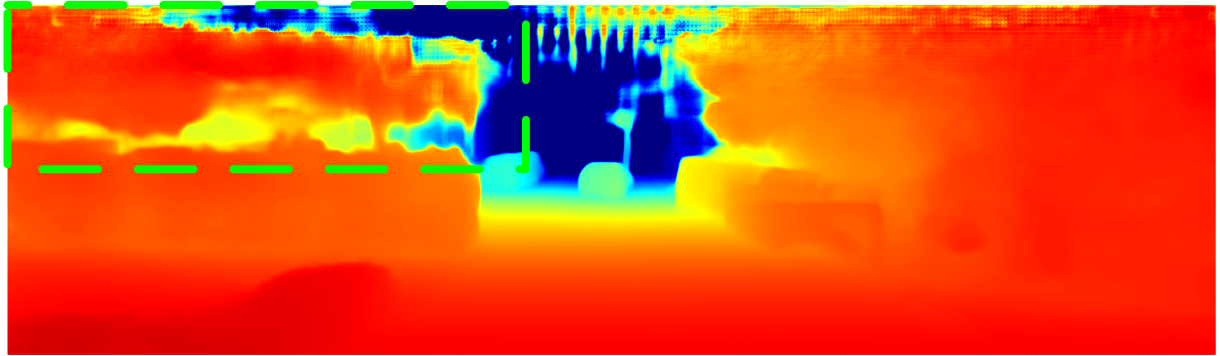}&
	\includegraphics[width=0.22\linewidth]{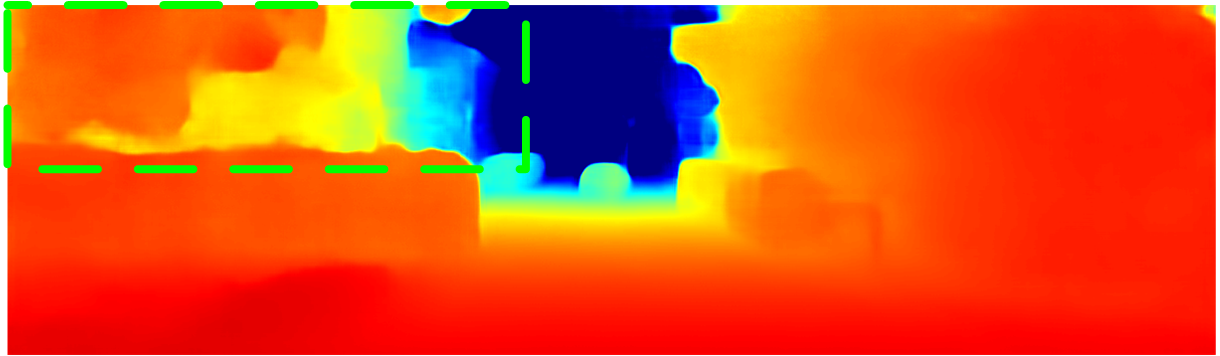}&
	\includegraphics[width=0.22\linewidth]{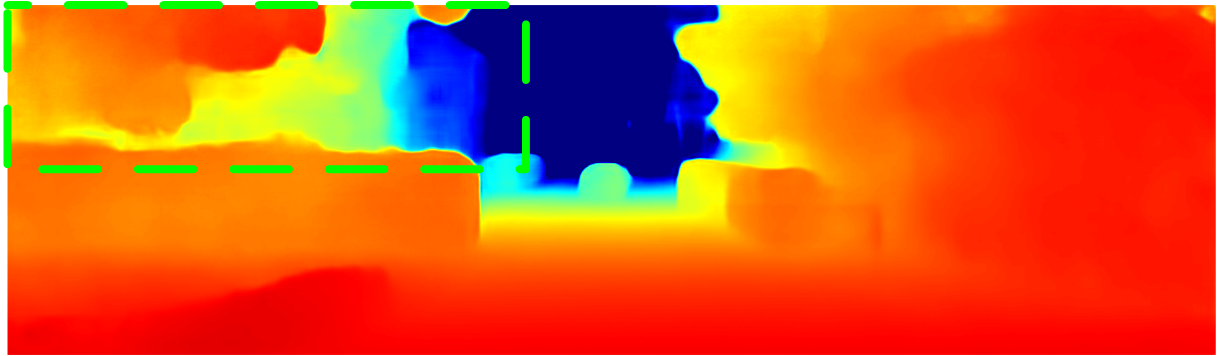}\\
	
	& \includegraphics[width=0.22\linewidth]{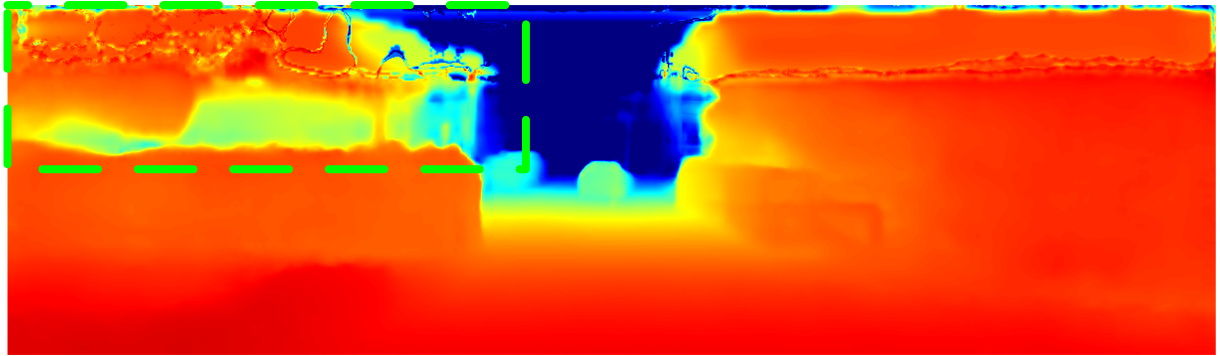}&
	\includegraphics[width=0.22\linewidth]{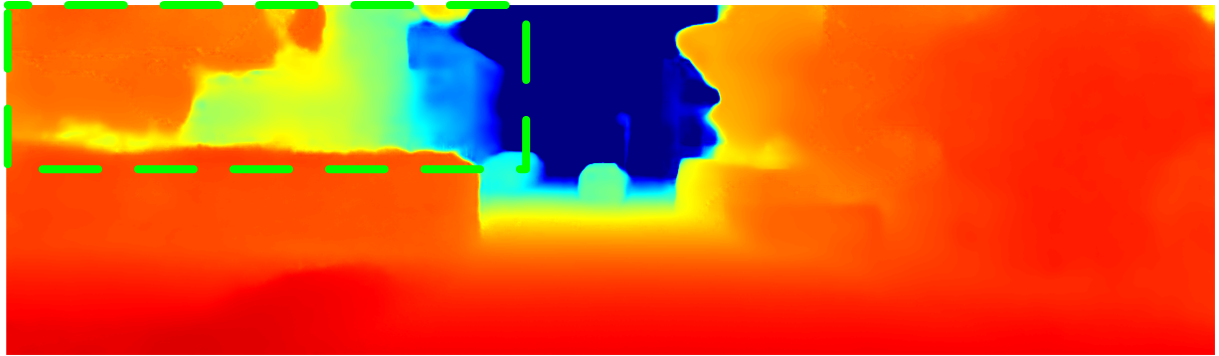}&
	\includegraphics[width=0.22\linewidth]{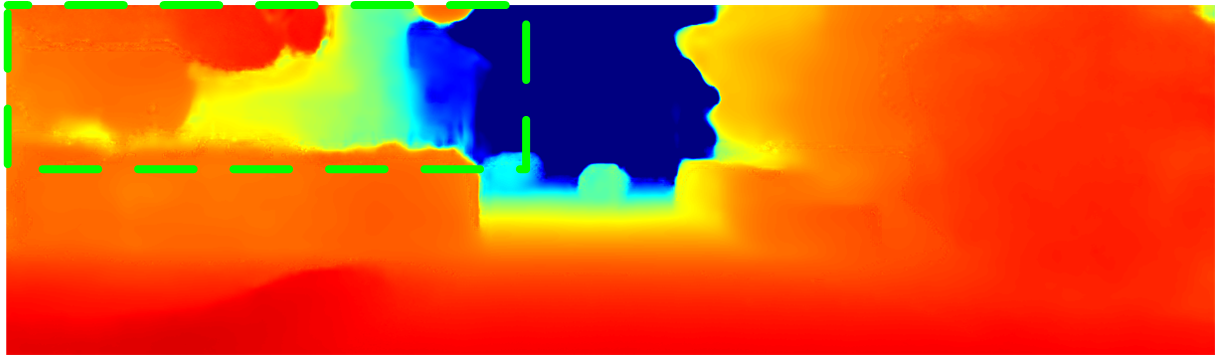}\\
	{\small (a) Left image} &  {(b) \small Supervision} &	{\small (c) Pseudo label }	&  {\small (d) Pseudo label\_full }	  	\\
	\end{tabular}
        \vspace{-0.05in}
		\caption{Visualization of monocular depth estimation results on KITTI dataset. The left panel shows the left input image of stereo image pairs, and for each example, \textbf{the first row} shows the predicted colorized depth map of \textbf{bts}\cite{bts} and \textbf{the second row} shows the predicted colorized depth map of \textbf{lapdepth} \cite{lapdepth}. Supervision denotes the method is trained with the ground truth depth map of training set. Pseudo label denotes the method is trained with the generated pseudo-labels of training set and Pseudo label full denotes the method in trained with the generated pseudo-labels of both training set and testing set.}
	\label{fig: monodepth}
	\vspace{-0.15in}
\end{figure*}

The comparison between state-of-the-art self-supervised and supervised monocular depth estimation methods is shown in Tab.~\ref{tab: mono_result}. It can be seen from the table that the proposed pseudo-label-based method (green areas) outperforms self-supervised monocular depth estimation approaches (pink areas) by a large margin, which proves that our generated pseudo-labels can well supervise monocular depth estimation approaches.  Moreover, due to the usage of ground truth, the performances of supervised depth estimation approaches (purple areas), such as DORN~\cite{dorn}, Bts~\cite{bts}, and lapdepth~\cite{lapdepth}, commonly outperform self-supervised based approaches (pink areas). In this paper, we claim that pseudo-label-based methods can achieve comparable or even better results than supervised approaches without the need for ground truth. Specifically, as shown in the green areas of Tab. \ref{tab: mono_result}, Bts~\cite{bts} and lapdepth~\cite{lapdepth} denote the original result of two representative supervised monocular depth estimation models trained by ourselves with the official implementation.  \emph{Bts\_pseudo\_label} and \emph{lapdepth\_pseudo\_label} denote we employ the pseudo-labels of the training dataset to supervise the two representative supervised monocular depth estimation models. Note that the only difference between the proposed pseudo-label-based methods and the corresponding original implementation is the supervision signals, i.e., pseudo-labels vs ground truth. As shown, \emph{Bts\_pseudo\_label} and \emph{lapdepth\_pseudo\_label} can achieve comparable results with the supervised version, e.g., the rmse of \emph{Bts\_pseudo\_label} in default evaluation is 3.730, which is only 0.97\% higher than  \emph{BTS}. Similar situations can also be observed in \emph{lapdepth\_pseudo\_label}. Note that we also observe that some previous work, such as SD-SSMDE \cite{SD-SSMDE} and monoResMatch \cite{monoresmatch} also explore using pseudo-label generated by traditional stereo matching methods \cite{monoresmatch} or self-distillation \cite{SD-SSMDE} to supervise the monocular depth estimation method. However, these methods still have a large gap between supervised depth estimation approaches and cannot achieve comparable results with the proposed uncertainty-based pseudo-label. Additionally, as the generation of pseudo-labels is not dependent on ground truth, pseudo-labels of the testing dataset can also be generated, which can be combined into the training dataset for better performance.  \emph{Bts\_pseudo\_label\_full} and \emph{lapdepth\_pseudo\_label\_full} in Tab. \ref{tab: mono_result} demonstrate the corresponding results. We can see that \emph{Bts\_pseudo\_label\_full} and \emph{lapdepth\_pseudo\_label\_full} can greatly improve the performances of \emph{Bts\_pseudo\_label} and \emph{lapdepth\_pseudo\_label} on all evaluation metrics. Moreover, they can even outperform the fully supervised method Bts~\cite{bts} and lapdepth~\cite{lapdepth} with large margins, which further verifies our hypothesis that stereo matching can be a viable alternative to reduce the cost of ground truth collection in the monocular depth estimation setting.

Qualitative comparison results on the KITTI Eigen test split are shown in Fig.~\ref{fig: monodepth}. Specifically, scenes with different depths of field are provided in which red denotes a smaller depth. As shown, our method can better distinguish objects in both foreground and background areas (see dash boxes in the pictures). Moreover, supervised methods generally cannot generate reasonable results on unlabeled areas, e.g., the sky region and the upper part of the scenes. This is mainly caused by the limitation of LIDAR, e.g., the ground truth obtained by lidar is very sparse and cannot collect valid data in the upper region of the scene. Instead, the proposed method can provide denser pseudo-labels and cover all regions in the image, thus significantly improving the visualization results on unlabeled areas. (see green dash boxes in the picture).

\section{Conclusion}
We have proposed an Uncertainty based cascade and fused cost volume representation for robust stereo matching. Specifically, a fused cost volume is proposed to alleviate the domain shifts and a cascade cost volume is employed to balance different disparity distributions, where pixel-level uncertainty estimation is at the core. We use it to adaptively narrow down the next stage’s pixel-level disparity searching space. Then, we propose an uncertainty-based pseudo-labels generation method to further narrow down the domain gap. By the cooperation between pixel-level and area-level uncertainty estimation, we can obtain a sparse while reliable pseudo label for domain adaptation without the need for ground truth. Experiment results show that our proposed method can achieve strong cross-domain, adapt, and joint generalization and obtain the 1st place on the stereo task of Robust Vision Challenge 2020. Moreover, our uncertainty-based pseudo-labels can be extended to train monocular depth estimation networks in an unsupervised way and even achieves comparable performance with the supervised methods.

\ifCLASSOPTIONcompsoc
  \section*{Acknowledgments}
  This research was supported in part by the National Natural Science Foundation of China (62271410), Jiangxi Natural Science Foundations (20224BAB212009), and the Fundamental Research Funds for the Central Universities.
\else
  \section*{Acknowledgment}
\fi


%





\ifCLASSOPTIONcaptionsoff
  \newpage
\fi

\bibliographystyle{plain}
\bibliography{ref_short}

\begin{IEEEbiography}
[{\includegraphics[width=1in,height=1.25in,clip,keepaspectratio]{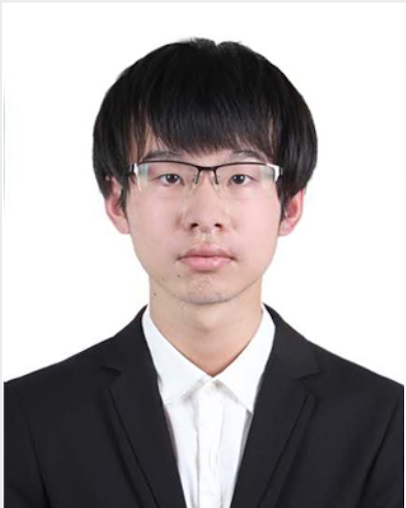}}] 
{Zhelun Shen} is currently a senior researcher at Robotics and Autonomous Driving Laboratory (RAL) of Baidu. He received his Master. degree in signal and information processing from Peking University, Beijing, China, and his B.E. degree in Communication Engineering from Northwestern Polytechnical University, Xi'an, China. He has published some papers in CVPR, ECCV, TNNLS, etc. Moreover, he won 1st place in the stereo matching task of the ECCV Robust Vision Challenge (RVC) and 1st place in the Argoverse Stereo Competition of CVPR2021 workshop on Autonomous Driving. His research interests include 3D scene understanding and its application in Autonomous Driving.
\end{IEEEbiography}

\vspace{-0.5in}

\begin{IEEEbiography}
[{\includegraphics[width=1in,height=1.25in,clip,keepaspectratio]{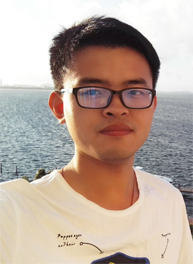}}] 
{Xibin Song} is a senior researcher at Robotics and Autonomous Driving Laboratory (RAL) of Baidu. He received his Ph.D. degree in school of Computer Science and Technology from Shandong University, Jinan, China, 2017. He worked as a joint Ph.D. student in the Research School of Engineering at the Australian National University, Canberra, Australia in 2015-2016. He received his B.E. degree in Digital Media and Technology from Shandong University, Jinan, China, in 2011. He served as reviewer for IEEE TIP, IEEE TPAMI, IEEE T-CSVT, CVPR, ICCV, ECCV, AAAI, etc. His research interests include Computer Vision and Augmented Reality.
\end{IEEEbiography}

\vspace{-0.5in}

\begin{IEEEbiography}
[{\includegraphics[width=1in,height=1.25in,clip,keepaspectratio]{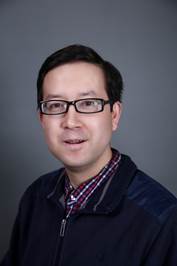}}] 
{Yuchao Dai} is currently a Professor with School of Electronics and Information at the Northwestern Polytechnical University (NPU), Xi'an, China. He received the B.E. degree, M.E degree and Ph.D. degree all in signal and information processing from NPU, in 2005, 2008 and 2012, respectively. He was an ARC DECRA Fellow with the Research School of Engineering at the Australian National University, Canberra, Australia. His research interests include structure from motion, multi-view geometry, low-level computer vision, deep learning, compressive sensing and optimization. He won the Best Paper Award at IEEE CVPR 2012, the Best Paper Award Nominee at IEEE CVPR 2020, the DSTO Best Fundamental Contribution to Image Processing Paper Prize at DICTA 2014, the Best Algorithm Prize in NRSFM Challenge at CVPR 2017, the Best Student Paper Prize at DICTA 2017 and the Best Deep/Machine Learning Paper Prize at APSIPA ASC 2017. He served as Area Chair for IEEE CVPR, ICCV, NeurIPS, ACM MM and etc.
\end{IEEEbiography}

\vspace{-0.5in}

\begin{IEEEbiography}
[{\includegraphics[width=1in,height=1.25in,clip,keepaspectratio]{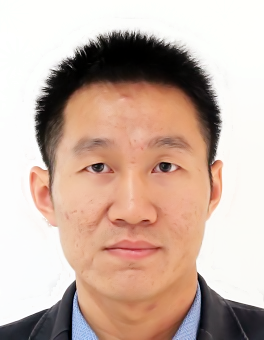}}] 
{Dingfu Zhou} is a senior research scientist at RAL of Baidu Research. Before joining Baidu, he worked as a Researcher Fellow in the Australian National University, Canberra, Australia. He obtained his Ph.D. degree in the University of Technology of Compiègne (member of Sorbonne University Association), Compiègne, France. He received both the B.E. degree and M.E degree in signal and information processing from Northwestern Polytechnical University, Xi'an, China, respectively. His research interests include deep learning-based point cloud analysis, structure from motion, multi-view geometry, and their application in Autonomous Driving. He won the Best Submission Prize in the nuScenes Detection Challenge at ICRA 2021. 
\end{IEEEbiography}

\vspace{-0.5in}

\begin{IEEEbiography}[{\includegraphics[width=1in,height=1.25in,clip,keepaspectratio]{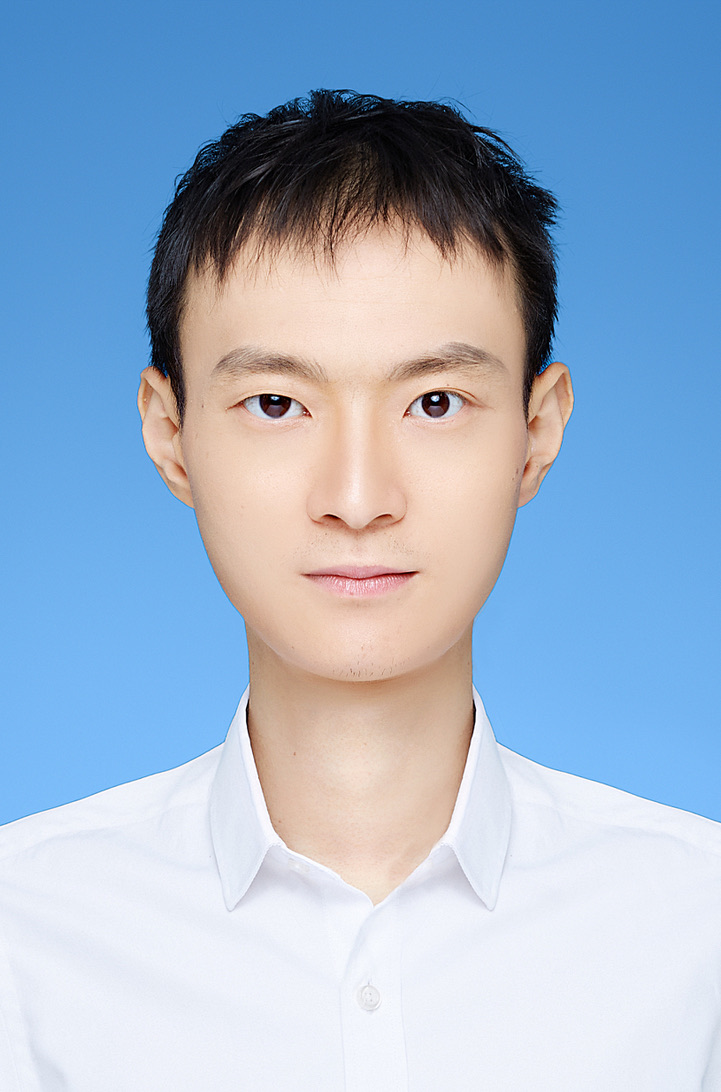}}]{Zhibo Rao}
received his B.E. and M.E. degrees in Electronic Information Engineering from Nanchang Hangkong University in 2017. He received his Ph.D. degree in signal and information processing from Northwestern Polytechnical University in 2022. He is currently a lecturer in the School of Information Engineering, Nanchang Hangkong University, Jiangxi, China. His primary research interests include pattern recognition, image processing, and multi-task learning in artificial intelligence. He has published some papers in TNNLS, TGRS, NCAA, TJVC, CVPR, ECCV, ICIP, etc. Meanwhile, He won 2nd place in the stereo matching task of ECCV Robust Vision Challenge (RVC).
\end{IEEEbiography}

\vspace{-0.5in}

\begin{IEEEbiography}
[{\includegraphics[width=1in,height=1.25in,clip,keepaspectratio]{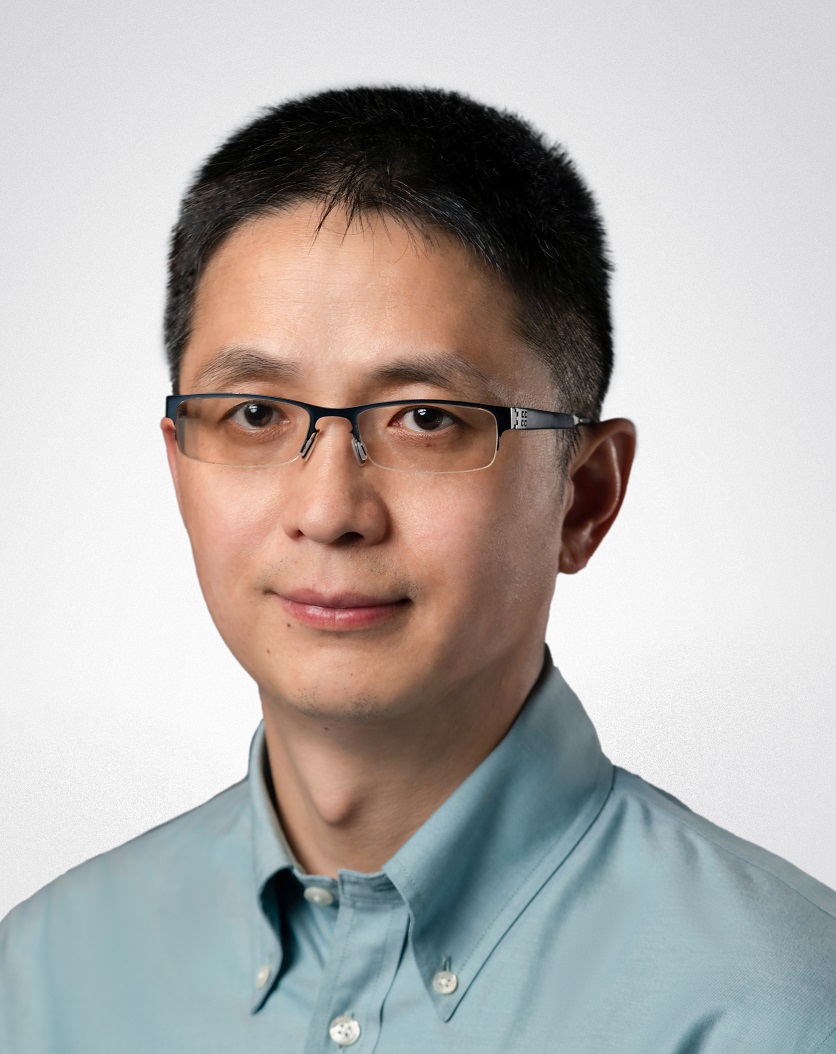}}] 
{Liangjun Zhang} is currently the Director of Robotics and Autonomous Driving Lab (RAL) of
Baidu Research USA and China. He received his PhD in computer science from the University of North Carolina at Chapel Hill in 2009 and MS/BS from Zhejiang University. He was an NSF Computing Innovation Fellow in the computer science department at Stanford University from 2009 to 2011. His research interests span robotics, autonomous driving, computer vision, simulation and geometric computing. He published research papers at Science Robotics, IJRR, TRO, TITS, TMM, ICCV, CVPR, ECCV, RSS, ICRA, IROS, ACC, and AAAI. He has received a number of awards including the First Place of nuScenes Detection Challenge organized in conjunction with ICRA 2021, the Best Paper Award at the International CAD Conference 2008 and the UNC Linda Dykstra Distinguished PhD Dissertation Award.
\end{IEEEbiography}

\end{document}